\documentclass[manuscript,screen]{acmart}

\usepackage{hyperref}
\usepackage{multirow}
\usepackage{graphicx}

\usepackage{caption} 
\usepackage{pgfplots}
\usepackage{pgfplotstable}
\usepackage{colortbl}
\usepackage{float}
\pgfplotsset{compat=1.17}
\usepackage{tikz}
\usetikzlibrary{patterns}
\usepackage{graphicx} 

\usepackage{tikz} 
\usepackage{amsmath}
\usepackage{CJKutf8}
\usepackage{hyperref}
\usepackage{cleveref}
\pdfoutput=1

\definecolor{skyblue}{RGB}{86,180,233}
\definecolor{bluishgreen}{RGB}{0,158,115}
\definecolor{yellow}{RGB}{240,228,66}
\definecolor{blue}{RGB}{0,114,178}
\definecolor{vermilion}{RGB}{213,94,0}
\definecolor{reddishpurple}{RGB}{204,121,167}

\acmConference[Conference acronym 'XX]{Make sure to enter the correct
  conference title from your rights confirmation emai}{June 03--05,
  2018}{Woodstock, NY}

\title{LogEval: A Comprehensive Benchmark Suite for Large Language Models In Log Analysis}

\author{Tianyu Cui}
\email{cuitianyu@mail.nankai.edu.cn}
\affiliation{%
  \institution{Nankai University}
  \city{Tianjin}
  \country{China}
}

\author{Shiyu Ma}
\email{mashiyu@mail.nankai.edu.cn}
\affiliation{%
  \institution{Nankai University}
  \city{Tianjin}
  \country{China}
}

\author{Ziang Chen}
\email{2012217@mail.nankai.edu.cn}
\affiliation{%
  \institution{Nankai University}
  \city{Tianjin}
  \country{China}
}

\author{Tong Xiao}
\email{xiaotong18@hnu.edu.cn}
\affiliation{%
  \institution{Tsinghua University}
  \city{Beijing}
  \country{China}
}

\author{Shimin Tao}
\email{taoshimin@huawei.com}
\affiliation{%
  \institution{Huawei}
  \city{Beijing}
  \country{China}
}

\author{Yilun Liu}
\affiliation{%
  \institution{Huawei}
  \city{Beijing}
  \country{China}
}

\author{Shenglin Zhang}
\authornote{Shenglin Zhang is the corresponding author.}
\email{zhangsl@nankai.edu.cn}
\affiliation{%
  \institution{Nankai University}
  \city{Tianjin}
  \country{China}
}

\author{Duoming Lin}
\email{2114010@mail.nankai.edu.cn}
\affiliation{%
  \institution{Nankai University}
  \city{Tianjin}
  \country{China}
}

\author{Changchang Liu}
\email{2113411@mail.nankai.edu.cn}
\affiliation{%
  \institution{Nankai University}
  \city{Tianjin}
  \country{China}
}

\author{Yuzhe Cai}
\email{2212113@mail.nankai.edu.cn}
\affiliation{%
 \institution{Nankai University}
  \city{Tianjin}
  \country{China}
}

\author{Weibin Meng}
\email{m_weibin@163.com}
\affiliation{%
  \institution{Huawei}
  \city{Beijing}
  \country{China}
}

\author{Yongqian Sun}
\email{sunyongqian@nankai.edu.cn}
\affiliation{%
  \institution{Nankai University}
  \city{Tianjin}
  \country{China}
}

\author{Dan Pei}
\email{peidan@tsinghua.edu.cn}
\affiliation{%
  \institution{Tsinghua University}
  \city{Beijing}
  \country{China}
}

\renewcommand{\shortauthors}{Shenglin Zhang, et al.}

\begin{abstract}
 Log analysis is crucial for ensuring the orderly and stable operation of information systems, particularly in the field of Artificial Intelligence for IT Operations (AIOps). Large Language Models (LLMs) have demonstrated significant potential in natural language processing tasks. In the AIOps domain, they excel in tasks such as anomaly detection, root cause analysis of faults, operations and maintenance script generation, and alert information summarization. However, the performance of current LLMs in log analysis tasks remains inadequately validated. To address this gap, we introduce \textbf{LogEval}, a comprehensive benchmark suite designed to evaluate the capabilities of LLMs in various log analysis tasks for the first time. This benchmark covers tasks such as log parsing, log anomaly detection, log fault diagnosis, and log summarization. LogEval evaluates each task using 4,000 publicly available log data entries and employs 15 different prompts for each task to ensure a thorough and fair assessment. By rigorously evaluating leading LLMs, we demonstrate the impact of various LLM technologies on log analysis performance, focusing on aspects such as self-consistency and few-shot contextual learning. We also discuss findings related to model quantification, Chinese-English question-answering evaluation, and prompt engineering. These findings provide insights into the strengths and weaknesses of LLMs in multilingual environments and the effectiveness of different prompt strategies. Various evaluation methods are employed for different tasks to accurately measure the performance of LLMs in log analysis, ensuring a comprehensive assessment. The insights gained from LogEval’s evaluation reveal the strengths and limitations of LLMs in log analysis tasks, providing valuable guidance for researchers and practitioners. Key findings indicate that while LLMs show promise in certain areas, there are notable challenges in handling complex log data and maintaining high accuracy across diverse tasks. LogEval is poised to significantly advance the application and development of LLMs in log analysis, offering effective solutions for practical log analysis challenges. The data and code are publicly available at  \href{https://github.com/LinDuoming/LogEval}{https URL}  to facilitate further research and development in this domain.
\end{abstract}

\begin{document}
\begin{CJK}{UTF8}{gbsn}
\maketitle

\makeatletter
\gdef\@ACM@checkaffil{}
\makeatother

\section{Introduction}

With the rapid development of information technology, information systems have become the cornerstone of business and organizational operations. Especially in fields such as cloud computing, 5G networks, and financial information systems, efficient IT operations are crucial for ensuring the stability and efficiency of these systems. The increasing scale and complexity of these systems, driven by the rapid development of the internet, have made AI-assisted operations, or AIOps, an emerging trend in the operations field. Gartner\cite{gartner} defines AIOps as a set of methods that use AI technology to handle tasks including anomaly detection, fault analysis, alert summarization, performance optimization, and capacity planning. In this context, log analysis plays a particularly significant role. Logs record the real-time status, key events, and error information of systems. By analyzing these data in-depth, operations personnel can quickly identify and resolve system issues, thereby enhancing system performance, reliability, and security. Logs are also essential in various scenarios: in cloud computing, they help monitor resource utilization and detect anomalies in virtual machines; in 5G networks, they track the performance of network components and identify faults; in financial information systems, they are used to audit transactions and ensure regulatory compliance. The ability to quickly and accurately analyze logs can prevent system outages, enhance performance, and maintain security. However, traditional log analysis methods primarily rely on manual processing and rule-setting, which are inefficient, prone to high false positive rates, and limited in handling large-scale, complex, and evolving log data.

Given the critical role of log analysis in maintaining system health, applying large language models (LLMs) to log analysis can significantly improve the efficiency and accuracy of these tasks. The widespread application of LLMs has revolutionized numerous fields. LLMs have garnered significant attention from both academia\cite{4-DBLP:journals/corr/abs-1711-02173,10-cticang2023survey} and industry\cite{36-hoffmann2022training,46-Li_2022,54-nakano2022webgpt,67-workshop2023bloom,90-zhang2022opt,93-ZHANG2021216} due to their advantages over traditional text generation methods. For instance, due to being capable of capturing rather long dependencies in sentences, LLMs are seeing wide adoption in commercial text generation including OpenAI’s GPT products (e.g., ChatGPT)\cite{6-Brown2020LanguageMA,11-chen2021evaluating,57-NEURIPS2022_b1efde53,61-Radford2019LanguageMA} and Meta’s LLaMA products\cite{73-touvron2023llama, roziere2023code}. Models like GPT-4\cite{GPT4}, LLaMA-2\cite{llama2}, ChatGLM4\cite{chatglm4}, and Qwen1.5\cite{qwen} have demonstrated their capabilities in tasks such as text generation, language translation, sentiment analysis, and more. For instance, GPT-4 is used in customer service chatbots to provide quick and accurate responses, LLaMA-2 enhances virtual assistants by improving natural language understanding, ChatGLM4 aids in content creation by generating coherent and contextually relevant text, and Qwen has been applied in various natural language understanding and generation tasks. These applications highlight the versatility and potential of LLMs in handling various natural language processing (NLP) tasks.

Despite the significant achievements of LLMs in natural language processing tasks and the existence of benchmarks for evaluating general NLP-related capabilities, their performance and applicability in log analysis tasks remain unclear. Therefore, we propose LogEval, a specialized benchmark suite designed to comprehensively assess the capabilities of LLMs across various log analysis tasks, such as log parsing, anomaly detection, fault diagnosis, and log summarization.

Nevertheless, due to the specialty of the log tasks, constructing
 an LogEval benchmark presents the following \textbf{challenges}:
 
\begin{itemize}
    \item \textbf{Data Sensitivity and Availability}: Although companies have vast amounts of operational data, high-quality datasets suitable for model training and evaluation are scarce. The lack of high-quality public datasets limits the effective evaluation and optimization of models.
    \item \textbf{Model Selection and Optimization}: The lack of a comprehensive and authoritative benchmark makes it difficult to evaluate the AIOps capabilities of current large models, especially their performance in log analysis tasks. The AI community sees rapid development with new models and technologies emerging frequently, making it crucial to select the most practical and effective solutions.
    \item \textbf{Evaluation and Verification}: Different log analysis tasks require distinct evaluation frameworks. Each task needs tailored evaluation metrics to ensure accurate and appropriate assessment of model performance.
\end{itemize}

To address these challenges, LogEval makes the following \textbf{contributions}:

\begin{itemize}
    \item \textbf{Characterization}: We are the first to investigate and characterize the application of LLMs in log analysis tasks, addressing the critical challenges and opportunities in this domain. Our comprehensive study involves extensive empirical analyses of LLMs, highlighting their potential to significantly enhance the efficiency and accuracy of log analysis while identifying specific areas that require further optimization.
    \item \textbf{Approach}: We introduce LogEval, a pioneering benchmark suite designed specifically for evaluating LLM capabilities in log analysis. LogEval includes:
        \begin{itemize}
            \item \textbf{Dataset Construction}: We constructed a diverse dataset containing 4,000 publicly available log entries, addressing challenges related to data sensitivity and resource limitations. This dataset encompasses 15 different Chinese and English prompts, rotated to minimize prompt-specific model performance biases.
            \item \textbf{Comprehensive Benchmark Development}: LogEval evaluates 18 mainstream large models across four primary log analysis tasks: log parsing, anomaly detection, fault diagnosis, and log summarization. We employ zero-shot and few-shot evaluation methods, leveraging techniques like self-consistency and prompt engineering to ensure consistent and accurate assessments.
            \item \textbf{Multidimensional Evaluation Metrics}: We designed various evaluation rules for each model to ensure precise assessments. For different tasks, we use metrics such as F1-score and accuracy, introducing new metrics based on semantic matching and average inference time to comprehensively evaluate LLM performance in log analysis tasks.
        \end{itemize}
    \item \textbf{Evaluation}: We conduct a rigorous evaluation of LogEval, assessing the performance of 18 mainstream LLMs on log analysis tasks. The evaluation demonstrates the strengths and limitations of each model, providing valuable insights for researchers and practitioners. The results reveal that while LLMs show promise in enhancing log analysis efficiency and accuracy, there are significant variations in performance across different tasks and models, underscoring the need for targeted optimizations.
\end{itemize}

Through the evaluation and analysis of LogEval, we aim to gain deeper insights into the strengths and limitations of LLMs in log analysis tasks, providing valuable guidance and reference for researchers and practitioners in the field. We believe that LogEval will play a significant role in advancing the application and development of LLMs for log analysis, offering effective solutions for practical challenges in real-world scenarios.

\textbf{Paper organization}：Our paper is organized as follows, \hyperref[sec:related_work]{Section 2} presents background information and discusses related work. \hyperref[sec:framework]{Section 3} details our methodology and evaluation framework. \hyperref[sec: EXPERIMENT DESIGN]{Section 4} describes the experimental setup. \hyperref[sec: 5]{Section 5}  shows the experimental results.  \hyperref[sec: discussion]{Section 6} provides a summary of the findings. \hyperref[sec: Conclusion]{Section 7} delves into a discussion of the results.\hyperref[sec: appendix]{Section 8} offers additional context and insights.

\section{Related Work}
\label{sec:related_work}

With the rapid advancement of LLMs, their diverse and complex capabilities have increasingly garnered significant attention. Traditional NLP metrics often fall short in accurately assessing these capabilities, prompting scholars to propose benchmarks specifically tailored for LLMs. This section discusses the evaluation of LLMs in general NLP domains and their applications in the specific context of log analysis tasks.

\subsection{Evaluation of LLMs in General NLP Tasks}

The evaluation of LLMs in NLP tasks has diversified as these models have become capable of handling increasingly complex and varied tasks. Evaluations now not only measure basic linguistic understanding and generation but also delve into nuanced capabilities such as reasoning, domain-specific knowledge, and adaptability to different tasks. Here, we categorize these evaluations based on the nature of the tasks and the methodologies used.

\textbf{Comprehensive Assessments:}
Comprehensive assessments are designed to evaluate the broad capabilities of LLMs across multiple dimensions. For instance, HELM \cite{helm} utilizes a diverse set of metrics to assess LLMs in 42 unique scenarios, providing insights into their general linguistic abilities and reasoning skills. BIG-bench \cite{big-bench} extends this by including tasks that challenge the models' understanding of common sense, logic, and even creativity.

\textbf{Specialized Knowledge Assessments:}
These assessments focus on evaluating the LLMs' performance in domains requiring specialized knowledge. For example, FinEval \cite{fineval} measures financial acumen, while MultiMedQA \cite{MultiMedQA} tests medical knowledge by using datasets derived from professional exams and consultation records. Similarly, Huatuo-26M \cite{Huatuo-26M} evaluates medical consultation capabilities, reflecting real-world medical inquiry handling.

\textbf{Real-World Application Simulations:}
Several benchmarks simulate real-world applications to see how well LLMs perform in practical scenarios. OpsEVAL \cite{Opseval} assesses the ability of LLMs to manage IT operations through a set of structured tasks in both Chinese and English. NetOps \cite{NetOps} focuses on network operations, testing LLMs with tasks that mimic real-world challenges in network management.

\textbf{Language Generation and Comprehension:}
This category tests the LLMs' ability to generate coherent and contextually appropriate text and to comprehend complex material. CG-Eval \cite{cgeval} assesses generation capabilities through tasks requiring term definitions, short-answer responses, and solving computational problems. MMCU \cite{mmcu} establishes a comprehension baseline with questions from academic and professional exams, pushing the models to demonstrate their understanding and application of learned knowledge.

\subsection{Evaluation of LLMs in Log Analysis Tasks}

Log analysis plays a crucial role in maintaining the health and performance of information systems. It involves several key tasks: log parsing, log anomaly detection, log fault diagnosis, and log summarization. Each task presents unique challenges and requires specific capabilities from LLMs.

As the application of LLMs in log analysis tasks increases, researchers have begun to explore how these models can be leveraged to enhance system monitoring and fault detection. Although some studies have attempted to apply LLMs to tasks such as log parsing \cite{llmparser,xu2023prompting,zhang2024eclipse} and anomaly detection \cite{egersdoerfer2023early,karlsen2023exploring,liu2023scalable,qi2023loggpt}, these applications are largely in the preliminary stages and lack a systematic evaluation framework to comprehensively measure the performance of LLMs in these tasks.

\textbf{Log Parsing Tasks:} Log parsing, the process of transforming raw logs into structured data, is foundational for log analysis. LILAC \cite{llmparser} introduces an adaptive parsing cache to significantly improve template accuracy and query times for large language models. DivLog \cite{xu2023prompting} is an LLM-based log parsing framework that achieves state-of-the-art performance, with an average accuracy of 98.1\%, precision of 92.1\%, and recall of 92.9\% across 16 public datasets. ECLIPSE \cite{zhang2024eclipse} introduces a novel approach, leveraging LLMs and semantic entropy-LCS, to address the challenges of log parsing in industrial settings.

\textbf{Log Anomaly Detection Tasks:} Identifying anomalous patterns within logs, commonly used for fault warning and detection, is another critical area. SeaLog \cite{liu2023scalable} employs a Trie-based Detection Agent for real-time anomaly detection and incorporates feedback from experts, including large language models like ChatGPT, to enhance accuracy. LogGPT \cite{qi2023loggpt} leverages ChatGPT's language interpretation capabilities for log-based anomaly detection, showing promising results and interpretability on BGL and Spirit datasets. This research indicates that the potential of LLMs in log anomaly detection tasks warrants further exploration.

\textbf{Log Fault Diagnosis Tasks:} Log fault diagnosis aims to identify the root causes of system faults through log analysis. Face It Yourselves \cite{shan2024face} introduces an LLM-powered two-stage approach for localizing configuration errors via logs, aiding end-users in identifying root causes without source code access. LogConfigLocalizer demonstrates high accuracy and feasibility in a case study.

\textbf{Log Generation Tasks:} Log generation involves automatically generating appropriate log statements to facilitate system maintenance and problem troubleshooting. UniLog \cite{unilog} leverages the in-context learning paradigm of large language models to generate log statements without the need for model tuning. With only a prompt containing five demonstration examples, UniLog can produce appropriate logging statements and further enhance its logging capabilities after warming up with a few hundred random samples.

\textbf{Other Applications:} Additionally, some studies explore the application of LLMs in specific areas of log analysis, such as LLM4Sec \cite{karlsen2023benchmarking} which evaluates various large language models for their suitability in log file analysis for cybersecurity. Summary Cycles \cite{block2023summary} investigates how Large Language Models can improve the efficiency of information handoff in collaborative intelligence analysis.

Currently, there is a lack of dedicated benchmarks for evaluating LLMs specifically in the context of log analysis tasks, making it challenging to assess and compare the performance of different models on these tasks. Therefore, this work aims to propose an evaluation framework for LLMs in log analysis tasks, addressing this research gap. Our evaluation efforts are not only intended to understand the strengths and limitations of LLMs in log analysis but also aim to provide valuable evaluation resources and guidance for the log analysis domain, promoting the effective application of LLMs in real-world log analysis scenarios.

Compared to previous research, our work provides a comprehensive evaluation framework that covers various log analysis tasks in the intelligent operations domain. By clearly defining task and capability classifications, LogEval offers detailed and extensive assessments, aiding in the selection and optimization of LLMs in log analysis and beyond.

\section{LogEval Benchmark}
\label{sec:framework}

\begin{figure*}
  \centering
  \includegraphics[width=1\linewidth]{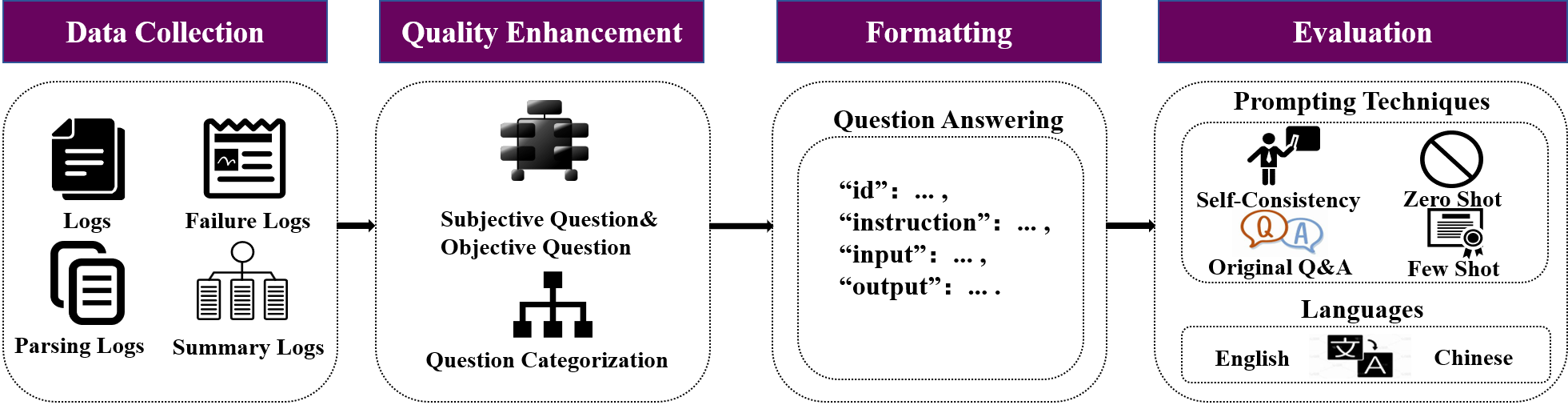}
  \caption{
    The framework of LogEval
  }
  \label{fig:framework}
\vspace{-0.3 cm}
\end{figure*}

This section presents the comprehensive framework of LogEval (\autoref{fig:framework}) from data collection to evaluation. The process involves four main stages: data collection, quality enhancement, formatting, and evaluation. The following subsections provide detailed descriptions and expansions for each step.

\subsection{Data Collection}
The data collection phase is critical for ensuring the breadth and representativeness of LogEval’s evaluation results. We systematically collected open-source and industry datasets for four key log analysis tasks: log parsing, log anomaly detection, log fault diagnosis, and log summarization.

\subsubsection{Log Parsing and Log Anomaly Detection}
We utilized large-scale datasets from LogPub~\cite{logpub1}, LogHub~\cite{logpub2}, and LogPAI~\cite{logpai}. LogPub~\cite{logpub1} includes real templates from 14 log datasets sourced from distributed systems, operating systems, and server-side applications. On average, each dataset in LogPub comprises 3.6 million log messages, all labeled with authentic log templates, totaling approximately 3500 templates. We selected commonly used BGL and ThunderBird datasets from LogPub for these tasks.

Specifically, the BGL (Blue Gene/L) dataset contains logs from large-scale parallel computing systems, while the ThunderBird dataset originates from high-performance computing clusters. By selecting these datasets, we ensure the diversity and representativeness of the data, covering a wide range of scenarios from distributed systems to high-performance computing environments. These datasets provide a solid foundation for evaluating LLM performance in log parsing and log anomaly detection tasks.

\subsubsection{Log fault Diagnosis}
We employed open-source datasets from Alibaba Cloud and China Mobile, both demonstrating strong performance in relevant events. These datasets are crucial for evaluating the diagnostic capabilities of LLMs.

The significance of these datasets lies in their comprehensive log entries generated during system operations, recording various operations and fault information. For example, the Alibaba Cloud dataset includes logs from cloud service operations, capturing diverse fault events, while the China Mobile dataset covers logs from telecommunication networks, providing rich practical data for evaluation.

\subsubsection{Log Summarization}
We used datasets labeled by LogSummary~\cite{logsummary}, including BGL, HDFS, HPC, Spark, Zookeeper, and Proxifier datasets, manually annotated based on data from LogHub. For each task, we collected 4000 logs, ensuring diversity and scale to cover various log types and analysis tasks comprehensively and fairly.

In the log summarization task, we paid particular attention to the diversity of log types. The BGL and HDFS datasets represent logs from high-performance computing and distributed file systems, while the HPC and Spark datasets involve logs from high-performance computing and big data processing environments. The Zookeeper and Proxifier datasets record logs from distributed coordination services and network proxy tools. By encompassing these different log types, we comprehensively evaluate LLM performance in generating concise and accurate log summaries.

By compiling these datasets, we ensured that LogEval can assess LLM performance across a wide range of scenarios, capturing the complexity and variability inherent in real-world log data.

\vspace{-1em}

\subsection{Quality Enhancement}
To enhance the quality of evaluation, we implemented a rigorous data preprocessing and quality enhancement process:

\subsubsection{Classification}
We categorized log analysis tasks into four types: Log Parsing (subjective questions), Log Anomaly Detection, Log fault Diagnosis (objective questions), and Log Summary (subjective questions). This classification ensures a comprehensive and detailed assessment of LLM performance across different log analysis tasks.

During the classification process, we carefully reviewed each task's dataset to ensure its applicability and representativeness. For example, subjective questions in log parsing tasks typically involve understanding and templating log structure, requiring models to identify and extract key elements from logs. In contrast, objective questions in log anomaly detection and log fault diagnosis require models to accurately identify and classify log events, often with clear answers. This classification ensures that each task's evaluation metrics and methods accurately reflect the model's performance in the specific task.

\subsubsection{Standardization}
We standardized the format of manually curated questions to ensure consistency. Each question was structured to include an instruction prompt, input, and output. This standardization is crucial for maintaining uniformity in evaluation and facilitating comparative analysis across different models.

The standardization process involved clearly defining the instruction prompts for each question, ensuring that models understand the task requirements. For example, in log parsing tasks, we provide clear instructions for models to convert logs into template formats; in log anomaly detection tasks, we instruct models to mark logs as "normal" or "abnormal." Additionally, we ensured that each input log and expected output adhered to a uniform format standard, enabling comparative analysis across different models.

\subsubsection{Question Categorization}
To further refine the evaluation process, we designed both subjective and objective questions. Subjective questions require models to generate responses based on understanding and contextual relevance, while objective questions provide clear, definitive answers. This dual approach helps in accurately gauging both the interpretative and factual capabilities of LLMs.

In designing subjective questions, we focus on the quality of model-generated responses, including coherence and contextual relevance. For example, in log summarization tasks, subjective questions require models to generate concise yet comprehensive log summaries, demonstrating the model's language generation capabilities and understanding of log content. For objective questions, we emphasize accuracy and consistency, such as identifying and marking abnormal log events in log anomaly detection tasks. This categorization allows for a comprehensive evaluation of LLM performance across different types of tasks.

\subsection{Formatting}
In the formatting phase, we established a structured approach to ensure clear and effective evaluation prompts:

\subsubsection{Prompt Structure}
Each prompt was designed to include clear instructions, context, and expected output. This structure ensures that the model understands the task requirements and can generate relevant responses. \autoref{fig:2} illustrates three zero-shot examples of formatted questions, demonstrating the clarity and coherence of the prompts used.

The key to prompt design is providing sufficient contextual information to enable the model to understand the task accurately. For example, in log parsing tasks, we provide a sample log and instruct the model to parse it into a standard template format; in log anomaly detection tasks, we provide multiple log samples and instruct the model to mark them as normal or abnormal. By providing clear instructions and context, we ensure that the model can generate high-quality responses, facilitating accurate evaluation.

\subsubsection{Bilingual Prompts}
We developed both Chinese and English prompts for each task, utilizing fifteen different prompts per task to mitigate the impact of prompt variations on evaluation results. This bilingual approach ensures that the evaluation covers linguistic diversity and provides a robust assessment of model capabilities in different languages. \autoref{tab
} presents examples of English prompts for each task.

The bilingual prompt design not only enhances the linguistic coverage of the evaluation but also helps detect performance differences when models handle tasks in different languages. For example, in log parsing tasks, we designed prompts with identical content in both Chinese and English, evaluating the model's performance in processing logs in both languages. By comparing model performance under bilingual prompts, we gain insights into the model's language processing capabilities and adaptability, providing references for model improvement and optimization.

\subsubsection{Diverse Scenarios}
The prompts were designed to cover a wide range of scenarios, reflecting the real-world complexity and variability of log data. This diversity is essential for testing the adaptability and generalization capabilities of LLMs.

In prompt design, we considered various possible log scenarios, including but not limited to system start-up and shutdown logs, error logs, performance logs, and user activity logs. Each scenario has its unique characteristics and challenges, requiring models to possess extensive knowledge and flexible processing capabilities. By covering these diverse scenarios, we comprehensively test the adaptability and generalization capabilities of LLMs, evaluating their performance in handling various real-world log tasks.

\begin{figure}[!ht]
    \begin{center}
    \begin{tikzpicture}
        \definecolor{labelcolor}{RGB}{240,240,240}
        \definecolor{predictcolor}{RGB}{255,228,216}
    
        \newcommand{\boxwidth}{0.31\textwidth} 
        \newcommand{\boxheight}{6.5cm}
        \newcommand{\boxheighttop}{0.5cm}
        \newcommand{\boxheightmiddle}{4cm}
        \newcommand{\boxheightbottom}{1cm}
        \newcommand{\boxgap}{0.005\textwidth} 
    
        \matrix[column sep=\boxgap, row sep=0mm] {
            \node[draw, text width=\boxwidth, minimum height=\boxheight] (box1) {};
            \node[fill=labelcolor, draw, text width=\boxwidth, minimum height=\boxheighttop, anchor=north west] at (box1.north west) {\parbox{\boxwidth} \raggedright \small \textbf{"id":} 0,};
            \node[draw, text width=\boxwidth, minimum height=\boxheightmiddle, anchor=north west] at ([yshift=-\boxheighttop]box1.north west) {\parbox{\boxwidth} \raggedright \small \textbf{"instruction":} "Please review the log entry and explicitly mark it as ‘normal' or ‘abnormal', only output ‘normal' or ‘abnormal'"};
            \node[fill=predictcolor, draw, text width=\boxwidth, minimum height=\boxheightbottom, anchor=north west] at ([yshift=-\boxheighttop-\boxheightmiddle]box1.north west) {\parbox{\boxwidth} \raggedright \small \textbf{"input":} "\textbackslash nlog entry:\textbackslash\textbackslash ninstruction cache parity error corrected"}; 
            \node[draw, text width=\boxwidth, minimum height=1cm, anchor=north west] at ([yshift=-\boxheighttop-\boxheightmiddle-\boxheightbottom]box1.north west) {\parbox{\boxwidth} \raggedright \small \textbf{"output":} "normal"}; &
    
            \node[draw, text width=\boxwidth, minimum height=\boxheight] (box2) {};
            \node[fill=labelcolor, draw, text width=\boxwidth, minimum height=\boxheighttop, anchor=north west] at (box2.north west) {\parbox{\boxwidth} \raggedright \small \textbf{"id"}: 0,};
            \node[draw, text width=\boxwidth, minimum height=\boxheightmiddle, anchor=north west] at ([yshift=-\boxheighttop]box2.north west) {\parbox{\boxwidth} \raggedright \small \textbf{"instruction":} "Parse the following log entry into a template format, replacing variable parts with a wildcard <*>, and focus the answer after the keyword ‘Answer'"};
            \node[fill=predictcolor, draw, text width=\boxwidth, minimum height=\boxheightbottom, anchor=north west] at ([yshift=-\boxheighttop-\boxheightmiddle]box2.north west) {\parbox{\boxwidth} \raggedright \small \textbf{"input":} "\textbackslash nlog entry:\textbackslash\textbackslash nsynchronized to 10.100.28.250, stratum 3"}; 
            \node[draw, text width=\boxwidth, minimum height=1cm, anchor=north west] at ([yshift=-\boxheighttop-\boxheightmiddle-\boxheightbottom]box1.north west) {\parbox{\boxwidth} \raggedright \small \textbf{"output":} "synchronized to <*>, stratum <*>"}; &
    
            \node[draw, text width=\boxwidth, minimum height=\boxheight] (box3) {};
            \node[fill=labelcolor, draw, text width=\boxwidth, minimum height=\boxheighttop, anchor=north west] at (box3.north west) {\parbox{\boxwidth} \raggedright \small \textbf{"id"}: 0,};
            \node[draw, text width=\boxwidth, minimum height=\boxheightmiddle, anchor=north west] at ([yshift=-\boxheighttop]box3.north west) {\parbox{\boxwidth} \raggedright \small \textbf{"instruction":} "In our data scenario, there are three types of faults: Processor CPU Caterr, Memory Throttled Uncorrectable Error Correcting Code, Hard Disk Drive Control Error Computer System Bus Short Circuit Programmable Gate Array Device Unknown. Analyze the log entry and identify the type of fault that occurred. Only output the fault type."};
            \node[fill=predictcolor, draw, text width=\boxwidth, minimum height=\boxheightbottom, anchor=north west] at ([yshift=-\boxheighttop-\boxheightmiddle]box3.north west) {\parbox{\boxwidth} \raggedright \small \textbf{"input":} "\textbackslash nlog entry:\textbackslash nProcessor \#0xfa | Configuration Error | Asserted"}; 
            \node[draw, text width=\boxwidth, minimum height=1cm, anchor=north west] at ([yshift=-\boxheighttop-\boxheightmiddle-\boxheightbottom]box1.north west) {\parbox{\boxwidth} \raggedright \small \textbf{"output":} "Processor CPU Caterr"}; \\
        };
    \end{tikzpicture}
    \end{center}
    \caption{Three examples of the processed questions}
    \label{fig:2}
\end{figure}

\begin{table*}[htbp]
\centering
\renewcommand{\arraystretch}{1.0}
\caption{Three English prompts for each task}
\label{tab:prompt design}
\begin{tabular} {lp{0.7\linewidth}}%
\toprule
\textbf{Tasks} & \textbf{English Prompt} \\
\toprule
Log Parsing & 
1. Parse the following log into a template format, replacing variable parts with \texttt{<*>}: [log] \newline
2. Convert the following log into a standardized template by identifying and replacing the variable parts with \texttt{<*>}: [log] \newline
3. Transform the raw log [log] into a log template by replacing variable segments with \texttt{<*>} \\ 
\midrule
Log Anomaly Detection &
1. Review and mark the log entry as "normal" or "abnormal", only output "normal" or "abnormal" \newline
2. Analyze the log content, classify it as "normal" or "abnormal", only output "normal" or "abnormal" \newline
3. Check the log entry, and determine if it belongs to the "normal" or "abnormal" category, only output "normal" or "abnormal" \\ 
\midrule
Log fault Diagnosis & 
1. In our data scenario, there are several types of faults \{fault types\}. Analyze the log [log] and identify the type of fault that occurred. Only output the fault type \newline
2. In our data scenario, there are several types of faults \{fault types\}. Based on the information in the log [log], determine which type of fault the log represents. Only output the fault type \newline
3. In our data scenario, there are several types of faults \{fault types\}. Use the detailed information provided by the log [log] to conduct an in-depth analysis to determine the category of the fault. Only output the fault type\\ 
\midrule
Log Summary & 
1. Analyze the following 20 logs [log], extract key information, phrases, sentences, or recurring content to generate a summary, and only output the summary \newline
2. Extract the most important events, phrases, and activities or recurring content from the following 20 logs [log], create a concise log overview, only output the summary \newline
3. Extract key events, sentence phrases, or recurring information from the following 20 logs [log] to form a comprehensive summary, only output the summary \\ 
\toprule
\end{tabular}
\vspace{-1em}
\end{table*}

\subsection{Evaluation Settings}
The evaluation phase involves assessing the performance of LLMs using a comprehensive set of metrics tailored to both objective and subjective questions:

\subsubsection{Objective Questions}
Objective questions are designed as multiple-choice questions with clear, definitive answers. The primary metrics used for evaluation are Accuracy and F1-score. Despite specifying fixed outputs and using few-shot prompts, LLM outputs may still contain extraneous information. Therefore, we employed a choice extraction function based on regular expressions to extract predicted answers. Accuracy is then calculated based on these extracted answers and ground-truth labels.

To ensure the accuracy of evaluation, we set clear criteria for each objective question. For example, in log anomaly detection and log fault diagnosis tasks, we use regular expressions to extract the model’s predicted answers and compare them with the ground-truth labels to calculate accuracy and F1-score. Additionally, we designed few-shot prompts, providing example answers to help models better understand the task requirements and improve their prediction accuracy.

\subsubsection{Subjective Questions}
Subjective questions require models to rely more on their understanding and knowledge base. The evaluation metrics include:
\begin{itemize}
    \item \textbf{Word Overlap}: Using ROUGE~\cite{rouge} scores, which are standard in NLP tasks, particularly in translation. These metrics assess the lexical similarity between the generated response and the reference answer..
    \item \textbf{Semantic Similarity}: Using cosine similarity to measure the semantic closeness between sentences. This metric provides insights into the contextual and conceptual accuracy of the generated responses.
\end{itemize}

In evaluating subjective questions, we focus on the quality of model-generated responses. For example, in log parsing and log summarization tasks, we use different evaluation metrics to comprehensively assess the model’s performance: log parsing uses parsing accuracy and edit distance as evaluation metrics. Parsing accuracy measures the model's ability to correctly parse log information, while edit distance evaluates the differences between the generated response and the reference answer. Log summarization uses accuracy and ROUGE-1 F1 scores to evaluate. Accuracy measures the correctness of the generated summaries, the accuracy of log summarization is calculated by using cosine similarity to measure the similarity between the generated summaries and the reference summaries. When the similarity exceeds a preset threshold (0.25), it is considered a correct prediction. Accuracy is the ratio of the number of correct predictions to the total number of predictions. ROUGE-1 F1 scores assess the lexical overlap between the generated and reference summaries.

\subsubsection{Additional Metrics}
To comprehensively assess the performance of LLMs, we introduced two additional metrics:
\begin{itemize}
    \item \textbf{Average Token}: Measures the average number of tokens generated by the model for a single log entry. This metric indicates the complexity and verbosity of the model’s output, reflecting the computational resources and processing time required.
    \item \textbf{Inference Time}: Measures the time taken by the model to process a single log entry, indicating the model’s response speed in practical applications. A lower inference time suggests higher efficiency and quicker response in real-world scenarios.
\end{itemize}

These additional metrics help us gain a more comprehensive understanding of the model's performance. For example, the average token count can help us evaluate the verbosity of the model's responses, optimizing the output efficiency. The inference time helps us assess the model's processing speed, particularly in real-world application scenarios. By comprehensively evaluating these metrics, we can fully understand the strengths and weaknesses of different models, providing references for selecting and optimizing LLMs for various log analysis tasks.

\section{EXPERIMENT DESIGN}
\label{sec: EXPERIMENT DESIGN}

In this section, we present the experimental design of LogEval, aiming to evaluate various LLMs to comprehend their effectiveness in addressing different types of questions (multiple-choice and open-ended) and various log analysis tasks.

\subsection{Models}

We evaluated popular LLMs from different organizations, covering a spectrum of weights. The selection criteria encompassed diversity in architecture, training data, and model size to ensure comprehensive analysis. Detailed information on all LLMs assessed is provided in  \autoref{tab:models}, with further details available in the appendix.

\begin{table*}
\centering
\renewcommand{\arraystretch}{1.0}
\setlength{\arrayrulewidth}{0.1pt} 
\caption{Models evaluated in this paper}
\label{tab:models}
\begin{tabular}{@{}l@{\extracolsep{0.5pt}}ccc@{}} 
\toprule
\textbf{Model} & \textbf{Creator} & \textbf{Parameters} & \textbf{Access} \\
\midrule
GPT-4 (OpenAI, 2023) & OpenAI & undisclosed & API \\ \hline
GPT-3.5 (OpenAI, 2022) & OpenAI & undisclosed & API \\ \hline
Claude-3-Sonnet (Anthropic, 2024) & Anthropic & undisclosed & API \\ \hline
Gemini-Pro (Gemini Team Google, 2023) & Google & undisclosed & API \\ \hline
Mistral (Jiang et al., 2023) & Mistral & 7B & Weights \\ \hline
InternLM2-Chat (Cai et al., 2024) & Shanghai AI Laboratory & 7B/20B & Weights \\ \hline
DevOps-Model-Chat (CodeFuse, 2023) & CodeFuse & 7B/14B & Weights \\ \hline
AquilaChat (BAAI, 2023) & BAAI & 7B & API \\  \hline
ChatGLM4 (Tsinghua Zhipu, 2024) & Tsinghua Zhipu & undisclosed & API \\ \hline
LLaMA-2 (Touvron et al., 2023) & Meta & 7/13/70B & API \\ \hline
Qwen-1.5-Chat (Bai et al., 2024) & Alibaba Cloud & 7/14/72B & API \\ \hline
Baichuan2-Chat (Yang et al., 2023) & Baichuan Intelligence & 13B & API \\
\bottomrule
\end{tabular}
\vspace{-1em}
\end{table*}

\subsection{Prompting Techniques}

To comprehensively understand the performance of different language models on log analysis tasks, we employ a variety of evaluation approaches. In objective question evaluations, we utilize both zero-shot and few-shot methods. With zero-shot evaluations, we aim to assess the language model's capabilities from the perspective of ordinary users, as users typically do not provide any examples in regular usage. With the few-shot approach, our goal is to evaluate the language model's potential from the perspective of developers, which often yields better performance than the zero-shot setup. For each evaluation method, we employ two settings to assess the language model: the naive Q\&A (Naive) and self-consistency(SC)\cite{sc} Q\&A . Given that we have both English and Chinese questions, we design corresponding naive Q\&A prompts for each language.
\begin{itemize}
    \item \textbf{Naive Question-Answer}: The language model is expected to generate answers without any additional explanations. 
    \item \textbf{Self-Consistency (SC)}: The same question is asked to the language model multiple times, and the answer that appears most frequently among the model's generated answers is extracted. In implementation, we set the number of SC queries to 5.
\end{itemize}

In subjective question evaluations, we combine each task along with the questions themselves as inputs to the language model. In subjective questions, we aim to simulate the everyday usage of language models by ordinary users. We input the questions into the language model and generate answers. Therefore, we only use the zero-shot evaluation for the language model in the naive Q\&A for subjective questions.         
                    
\subsection{Baselines Design}

For the baseline of log anomaly detection, We choose NeuralLog\cite{neurallog} and LogRobust\cite{logrobust} .

\begin{itemize}
    \item \textbf{NeuralLog}: NeuralLog is a novel approach that utilizes deep learning to detect anomalies directly from raw log data without the need for traditional log parsing. NeuralLog automates the feature extraction process by learning the inherent patterns and structures within the unstructured log texts, effectively bypassing complex preprocessing steps. This method significantly reduces the reliance on domain knowledge and manual effort typically required in setting up log anomaly detection systems. NeuralLog has demonstrated high accuracy and efficiency in anomaly detection, making it particularly valuable for real-time monitoring systems. The benefits of adopting NeuralLog include simplified system maintenance, improved automation in anomaly detection processes, and enhanced accuracy in identifying potential threats or system faults promptly.
    \item \textbf{LogRobust}: LogRobust is a methodology designed to enhance anomaly detection in environments where log data is prone to instability and frequent changes. LogRobust employs advanced techniques to adaptively learn and update its models as it encounters new or altered log messages, ensuring resilience to changes in log formats or content. It utilizes a combination of unsupervised learning algorithms to detect outliers and anomalies effectively even in highly dynamic systems. By focusing on stability and adaptability, LogRobust minimizes the false positive rates that often plague traditional log anomaly detection systems facing volatile data. The primary benefits of LogRobust are its robustness against log data variations, improved anomaly detection accuracy, and reduced need for manual intervention in maintaining parsing models, making it ideal for critical systems requiring continuous monitoring.
\end{itemize}

For the baseline of log parsing, we choose Drain \cite{drain} and LogPPT \cite{logppt}.

\begin{itemize}
    \item \textbf{Drain}: Drain is an innovative online log parsing method which leverages a fixed-depth tree structure to systematically group and parse log messages. The core idea behind Drain is to categorize log lines based on predefined log grouping rules and extract templates using a fixed depth parse tree, minimizing computational overhead and increasing parsing speed. By employing a parsing tree with a fixed depth and using heuristics to handle variability in log data, Drain ensures both high efficiency and accuracy in real-time log parsing scenarios. This method efficiently adapts to diverse log formats and dynamically changing log content, reducing the need for frequent manual reconfiguration. The benefits of Drain include significant improvements in parsing speed and flexibility, making it an effective solution for systems that require real-time log analysis and monitoring.
    \item \textbf{LogPPT}: LogPPT is a deep learning-based method for log parsing, aiming to automatically parse logs and improve accuracy by learning patterns and structures in log files. Treating log lines as sequential data, it models sequence relationships using deep learning models to enhance parsing efficiency and generalization capability. The advantages of LogPPT lie in its automated parsing, accuracy, generalization capability, and efficiency improvement, providing strong support for the field of log analysis.
\end{itemize}

For the baseline of log fault diagnosis, we choose LogKG \cite{logkg} and LogCluster \cite{logcluster}.

\begin{itemize}
    \item \textbf{LogKG}: LogKG is a framework that utilizes knowledge graphs to enhance the process of diagnosing faults from system logs. LogKG constructs a comprehensive knowledge graph from parsed log data, integrating various log entities and their relationships to capture a detailed representation of system interactions and behaviors. This structured approach enables more precise and interpretable diagnostics by utilizing graph-based analytics to trace faults and identify their root causes effectively. By integrating semantic reasoning and relational data, LogKG facilitates an in-depth analysis that outperforms traditional log analysis methods which often depend solely on textual data. The key benefits of using LogKG include improved accuracy in fault diagnosis, faster problem resolution times, and a more intuitive understanding of complex system behaviors, all of which contribute to better reliability and maintenance of IT systems.
    \item \textbf{LogCluster}: LogCluster is a log fault diagnosis technique leveraging clustering, where it primarily involves preprocessing raw logs to create structured representations, computing vectorized representations of log sequences, measuring the similarity between log events, and subsequently clustering similar logs using hierarchical clustering. This method efficiently automates the discovery of typical and anomalous patterns amidst voluminous logs, significantly reducing manual troubleshooting efforts, making it particularly suitable for log fault diagnosis in large-scale distributed systems.
\end{itemize}

For the baseline of log summary, we choose LogSummary \cite{logsummary}.

\begin{itemize}
    \item \textbf{LogSummary}: LogSummary generates concise log summaries by extracting and ranking key phrases, aiming to preserve critical information from the raw logs while minimizing redundancy. The method begins with the preprocessing of logs, including cleaning and normalization, followed by employing algorithms like TF-IDF or TextRank to identify and extract key information from the logs. Finally, it constructs summaries based on the extracted key information. LogSummary’s benefits lie in its ability to swiftly produce high-quality log summaries that offer users a bird’s-eye view of log insights, facilitating rapid issue localization. It is particularly suitable for compressing and quickly analyzing large-scale, real-time log streams.
\end{itemize}

In the baseline experiments, for tasks related to log parsing and log summary, we adopt the same dataset utilized in the evaluation of large language models, comprising 4,000 logs. For log anomaly detection and fault diagnosis, our dataset consists of 4,000 sequences, each formed by the raw 4,000 logs from the large language model assessment and their respective 10 logs above and below in context.

\section{EVALUATION}
\label{sec: 5}

\begin{figure*}[!ht]
\begin{center}
\begin{tikzpicture}
    \begin{axis}[
        colorbar,
        colormap={yelloworange}{
            color(0cm)=(yellow!50!white)
            color(1cm)=(orange!50!white)
            color(2cm)=(orange)
            color(3cm)=(orange!70!black)
            color(4cm)=(red)
        },
        xlabel={Tasks},
        ylabel={Models},
        xtick=data,
        ytick=data,
        xticklabels={Parsing, Anomaly, Diagnosis, Summary},
        yticklabels={Claude3 Sonnet, GPT-4, Gemini Pro, ChatGLM4, GPT-3.5, Qwen1.5-7B, LLaMa2-70B, LLaMa2-7B, Qwen1.5-72B, LLaMa2-13B, Qwen1.5-14B},
        x tick label style={font=\footnotesize},
        y tick label style={font=\footnotesize},
        point meta min=0,
        point meta max=1,
        enlargelimits=false,
        scale only axis,
        height=0.4\textwidth,
        width=0.7\textwidth,
        nodes near coords,
        every node near coord/.append style={font=\footnotesize},
        nodes near coords align={center},
    ]
    \addplot [matrix plot*, point meta=explicit, mesh/cols=4, nodes near coords, every node near coord/.append style={font=\footnotesize}] table [meta=z] {
        x y z
        0 0 0.42
        1 0 0.35
        2 0 0.37
        3 0 0.59
        0 1 0.58
        1 1 0.33
        2 1 0.35
        3 1 0.43
        0 2 0.36
        1 2 0.49
        2 2 0.34
        3 2 0.46
        0 3 0.23
        1 3 0.42
        2 3 0.38
        3 3 0.36
        0 4 0.24
        1 4 0.26
        2 4 0.36
        3 4 0.45
        0 5 0.06
        1 5 0.52
        2 5 0.33
        3 5 0.38
        0 6 0.06
        1 6 0.77
        2 6 0.16
        3 6 0.41
        0 7 0.02
        1 7 0.57
        2 7 0.22
        3 7 0.4
        0 8 0.26
        1 8 0.33
        2 8 0.25
        3 8 0.41
        0 9 0.07
        1 9 0.54
        2 9 0.22
        3 9 0.35
        0 10 0.08
        1 10 0.26
        2 10 0.27
        3 10 0.32
    };
    \end{axis}
\end{tikzpicture}
\end{center}
\caption{The Accuracy in zero-shot Naive Q\&A}
\label{fig:40}
\end{figure*}
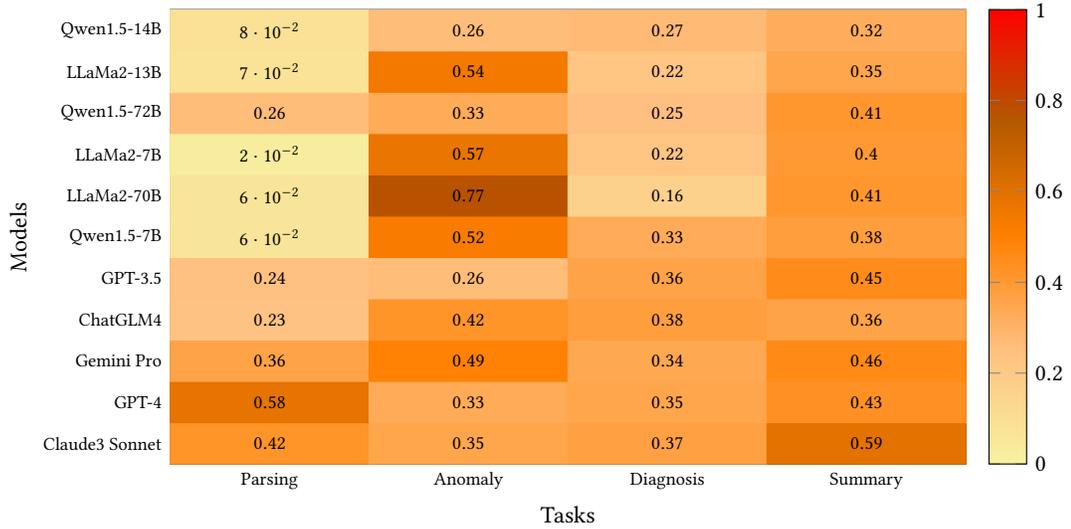

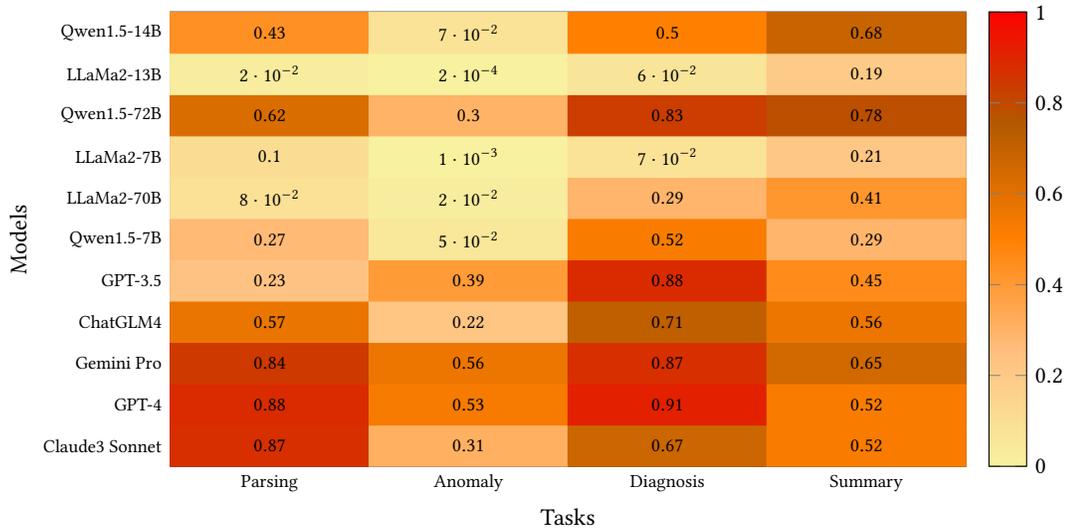
\begin{figure*}[!ht]
\begin{center}
\begin{tikzpicture}
    \begin{axis}[
        colorbar,
        colormap={yelloworange}{
            color(0cm)=(yellow!50!white)
            color(1cm)=(orange!50!white)
            color(2cm)=(orange)
            color(3cm)=(orange!70!black)
            color(4cm)=(red)
        },
        xlabel={Tasks},
        ylabel={Models},
        xtick=data,
        ytick=data,
        xticklabels={Parsing, Anomaly, Diagnosis, Summary},
        yticklabels={Claude3 Sonnet, GPT-4, Gemini Pro, ChatGLM4, GPT-3.5, Qwen1.5-7B, LLaMa2-70B, LLaMa2-7B, Qwen1.5-72B, LLaMa2-13B, Qwen1.5-14B},
        x tick label style={font=\footnotesize},
        y tick label style={font=\footnotesize},
        point meta min=0,
        point meta max=1,
        enlargelimits=false,
        scale only axis,
        height=0.4\textwidth,
        width=0.7\textwidth,
        nodes near coords,
        every node near coord/.append style={font=\footnotesize},
        nodes near coords align={center},
    ]
    \addplot [matrix plot*, point meta=explicit, mesh/cols=4, nodes near coords, every node near coord/.append style={font=\footnotesize}] table [meta=z] {
        x y z
        0 0 0.87
        1 0 0.31
        2 0 0.67
        3 0 0.52
        0 1 0.88
        1 1 0.53
        2 1 0.91
        3 1 0.52
        0 2 0.84
        1 2 0.56
        2 2 0.87
        3 2 0.65
        0 3 0.57
        1 3 0.22
        2 3 0.71
        3 3 0.56
        0 4 0.23
        1 4 0.39
        2 4 0.88
        3 4 0.45
        0 5 0.27
        1 5 0.05
        2 5 0.52
        3 5 0.29
        0 6 0.08
        1 6 0.02
        2 6 0.29
        3 6 0.41
        0 7 0.1
        1 7 0.001
        2 7 0.07
        3 7 0.21
        0 8 0.62
        1 8 0.3
        2 8 0.83
        3 8 0.78
        0 9 0.02
        1 9 0.0002
        2 9 0.06
        3 9 0.19
        0 10 0.43
        1 10 0.07
        2 10 0.5
        3 10 0.68
    };
    \end{axis}
\end{tikzpicture}
\end{center}
\caption{The Accuracy in few-shot Naive Q\&A}
\label{fig:39}
\end{figure*}

In this section, to comprehensively and intuitively demonstrate the performance of various models and their overall evaluation across different tasks, we have designed two heatmaps, as illustrated in \autoref{fig:40} and \autoref{fig:39}.

The heatmaps specifically highlight the zero-shot and few-shot performances of select large models in the context of original question-answering scenarios. Each radar chart incorporates four log analysis tasks as evaluation metrics, comprehensively spanning log parsing, log anomaly detection, log fault diagnosis, and log summary extraction. The variations in the polygon shapes of the radar charts reveal that the models exhibit better performance in log parsing tasks but fare less satisfactorily in log anomaly detection. Furthermore, it is evident that the few-shot approach yields superior results compared to zero-shot, illustrating the models' capability to learn task-relevant knowledge from just a few examples.

\autoref{fig:40} shows the performance of different models in zero-shot scenarios across four tasks: log parsing, log anomaly detection, log fault diagnosis, and log summary extraction. From the figure, we can observe the following:

\begin{itemize}
    \item \textbf{Log Parsing Task}: 
        \begin{itemize}
            \item \textbf{GPT-4} performs the best with a score of 0.58, demonstrating its strong natural language parsing capabilities.
            \item \textbf{Claude3 Sonnet} and \textbf{Gemini Pro} also perform well, with scores of 0.42 and 0.36, respectively, indicating good performance in log parsing.
        \end{itemize}
    \item \textbf{Log Anomaly Detection Task}: 
        \begin{itemize}
            \item \textbf{LLama2-70B} stands out with the highest score of 0.77, showing its strong ability in detecting log anomalies.
            \item \textbf{LLama2-7B} and \textbf{LLama2-13B} also perform well, with scores of 0.57 and 0.54, respectively.
        \end{itemize}
    \item \textbf{Log Fault Diagnosis Task}: 
        \begin{itemize}
            \item \textbf{ChatGLM4} and \textbf{Claude3 Sonnet} perform the best in this task, with scores of 0.38 and 0.37, respectively.
            \item \textbf{GPT-3.5} and \textbf{GPT-4} also show good performance, with scores of 0.36 and 0.35, respectively.
        \end{itemize}
    \item \textbf{Log Summary Extraction Task}: 
        \begin{itemize}
            \item \textbf{Claude3 Sonnet} performs the best with a score of 0.59, demonstrating its strong ability in summarizing information.
            \item \textbf{GPT-3.5} and \textbf{Gemini Pro} also perform well, with scores of 0.45 and 0.46, respectively.
        \end{itemize}
\end{itemize}

\autoref{fig:39} illustrates the performance of different models in few-shot scenarios across the same four tasks. From the figure, we can observe the following:

\begin{itemize}
    \item \textbf{Log Parsing Task}: 
        \begin{itemize}
            \item \textbf{GPT-4} performs the best with a score of 0.88, demonstrating its strong few-shot learning capabilities.
            \item \textbf{Claude3 Sonnet} and \textbf{Gemini Pro} also perform exceptionally well, with scores of 0.87 and 0.84, respectively.
        \end{itemize}
    \item \textbf{Log Anomaly Detection Task}: 
        \begin{itemize}
            \item \textbf{Gemini Pro} stands out with the highest score of 0.56, indicating its effective anomaly detection in few-shot scenarios.
            \item \textbf{GPT-4} and \textbf{GPT-3.5} also perform well, with scores of 0.53 and 0.39, respectively.
        \end{itemize}
    \item \textbf{Log Fault Diagnosis Task}: 
        \begin{itemize}
            \item \textbf{GPT-4} and \textbf{GPT-3.5} perform the best, with scores of 0.91 and 0.88, respectively, showing their superiority in complex fault diagnosis tasks.
            \item \textbf{Gemini Pro} and \textbf{Qwen1.5-72B} also show strong performance with scores of 0.87 and 0.83, respectively.
        \end{itemize}
    \item \textbf{Log Summary Extraction Task}: 
        \begin{itemize}
            \item \textbf{Qwen1.5-72B} and \textbf{Qwen1.5-14B} perform well, with scores of 0.78 and 0.68, respectively.
            \item \textbf{Gemini Pro} and \textbf{ChatGLM4} also show improved performance, with scores of 0.65 and 0.56, respectively.
        \end{itemize}
\end{itemize}

From the zero-shot and few-shot performances, the following conclusions and patterns can be drawn:

\textbf{Task Adaptability of Models}: \textbf{GPT-4} shows stable performance across multiple tasks, particularly excelling in few-shot scenarios, demonstrating strong task adaptability and few-shot learning abilities. Its performance in log parsing and fault diagnosis tasks is particularly notable, making it suitable for applications requiring high precision parsing and diagnosis.\textbf{Claude3 Sonnet} excels in log parsing and log summary extraction tasks, showcasing its potential in information extraction and summarization, suitable for scenarios requiring efficient information extraction.\textbf{LLama2-70B} performs excellently in zero-shot scenario in log anomaly detection, making it suitable for anomaly detection tasks, demonstrating its strong capability in recognizing anomalies.

\textbf{Few-shot Learning Capability}:
Overall, the few-shot scenario performance surpasses the zero-shot scenario, indicating that these large models can learn task-relevant knowledge from a small number of examples. This is particularly significant for real-world applications where data may be limited. \textbf{GPT-4} and \textbf{Claude3 Sonnet} are especially notable in few-shot learning scenarios, making them ideal for applications requiring rapid adaptation and efficient learning.

\textbf{Task Performance Variance}:
Different models exhibit significant performance variance across different tasks, suggesting that model selection should be based on specific task requirements. For instance, \textbf{LLama2-70B} is preferable for log anomaly detection tasks in zero-shot scenario, while \textbf{GPT-4} and \textbf{Claude3 Sonnet} are better suited for complex log fault diagnosis tasks.

\textbf{Selection Strategy for Practical Applications}:
For scenarios requiring multi-task processing, models with stable performance like \textbf{GPT-4} should be prioritized due to their exceptional performance across multiple tasks, especially in few-shot scenarios where their strong learning capabilities can significantly enhance task efficiency.
If the application scenario primarily involves log parsing and summarization, \textbf{Claude3 Sonnet} is a suitable choice due to its outstanding performance in these tasks.
For tasks focused on anomaly detection, \textbf{LLama2-70B} is recommended, as it outperforms other models in log anomaly detection.

These insights provide valuable references for the application of large models in log analysis, further demonstrating the effectiveness of few-shot learning methods in improving model performance. Future research can further explore the performance of these models in other tasks, seeking more optimization strategies and application scenarios.

\begin{figure}[!ht]
  \centering
  \begin{minipage}{0.45\textwidth}
    \centering
    \includegraphics[width=\textwidth]{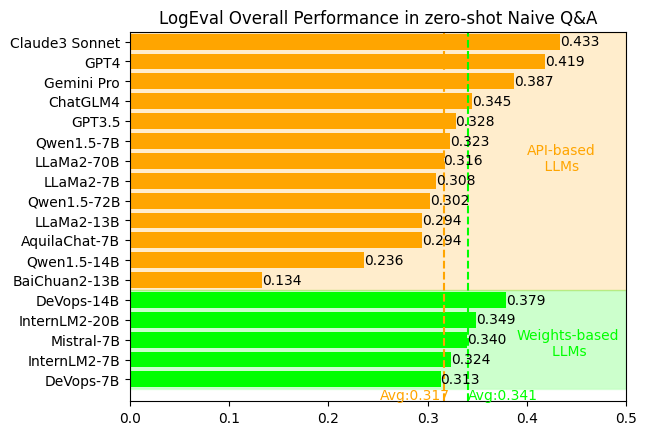}
    \caption{LogEval Overall Performance in zero-shot Naive Q\&A}
    \label{fig:37}
  \end{minipage}\hfill
  \begin{minipage}{0.45\textwidth}
    \centering
    \includegraphics[width=\textwidth]{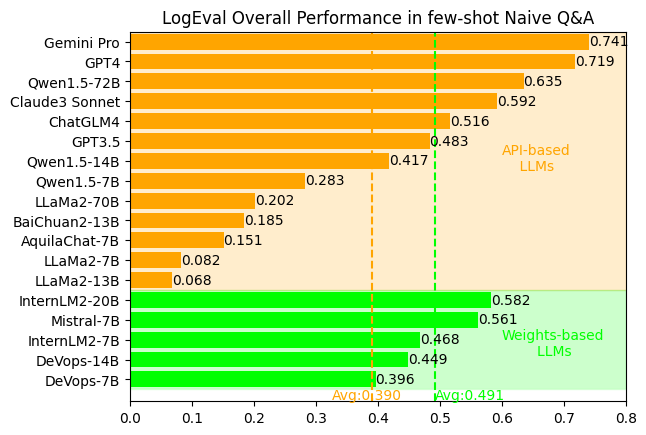}
    \caption{LogEval Overall Performance in few-shot Naive Q\&A}
    \label{fig:38}
  \end{minipage}
\end{figure}

Additionally, we have aggregated the average accuracy scores across these four tasks to conduct a holistic assessment of all large models across these tasks. As depicted in \autoref{fig:37} and \autoref{fig:38}, which illustrate the models' performance under both zero-shot and few-shot settings in the context of original question-answering.

From the overall performances, the following conclusions and patterns can be drawn:

\begin{itemize}
    \item \textbf{Zero-shot Performance}:
    \begin{itemize}
        \item \textbf{Top Performers:}
        \begin{itemize}
            \item Claude3 Sonnet and GPT-4 lead with accuracy scores of 0.433 and 0.419, respectively.
            \item Gemini Pro and DeVops-14B also show strong performances with scores of 0.387 and 0.379.
        \end{itemize}
        
        \item \textbf{API-based LLMs:}
        \begin{itemize}
            \item Among API-based LLMs, GPT-4 and Claude3 Sonnet perform the best in zero-shot settings. Their high accuracy scores indicate robust performance across various tasks without additional tuning.
            \item Gemini Pro and ChatGLM4 also show good performance among the API-based models, with scores of 0.387 and 0.345, respectively.
        \end{itemize}
        
        \item \textbf{Weight-based LLMs:}
        \begin{itemize}
            \item InternLM2-20B and DeVops-14B demonstrate reasonable performance in zero-shot settings, with scores of 0.349 and 0.379. These models show potential but typically lag behind API-based counterparts.
            \item InternLM2-7B and Mistral-7B also show competitive performance among the weight-based models, with scores of 0.324 and 0.340, respectively.
        \end{itemize}
    \end{itemize}
    
    \item \textbf{Few-shot Performance}:
    \begin{itemize}
        \item \textbf{Top Performers:}
        \begin{itemize}
            \item Gemini Pro and GPT-4 again lead with higher accuracy scores of 0.741 and 0.719, respectively.
            \item Qwen1.5-72B and Claude3 Sonnet also show improved performance with scores of 0.635 and 0.592.
        \end{itemize}
        
        \item \textbf{API-based LLMs:}
        \begin{itemize}
            \item In the few-shot setting, GPT-4 and Gemini Pro stand out among API-based LLMs, maintaining leading positions with their high accuracy scores.
            \item Qwen1.5-72B and Claude3 Sonnet also demonstrate strong performance among API-based models, with scores of 0.635 and 0.592, respectively.
        \end{itemize}
        
        \item \textbf{Weight-based LLMs:}
        \begin{itemize}
            \item InternLM2-20B and Mistral-7B excel in the few-shot setting with accuracy scores of 0.582 and 0.561. This improvement underscores the importance of task-specific fine-tuning.
            \item InternLM2-7B and DeVops-14B show enhanced performance among the weight-based models in few-shot settings, with scores of 0.468 and 0.449, respectively.
        \end{itemize}
    \end{itemize}
\end{itemize}

From the comparison of average performance, it is evident that few-shot learning significantly enhances the accuracy of both types of models. The average accuracy of API-based LLMs increases from 0.317 in zero-shot to 0.390 in few-shot settings, while weight-based LLMs show an improvement from 0.341 to 0.491.

In this study, we differentiate between models based on their API utilization and weight-based approaches. The primary distinction lies in how these models are deployed and accessed. API-based LLMs, such as GPT-4, Gemini Pro, Claude3 Sonnet, and ChatGLM4, are accessed via APIs provided by service providers. These models typically benefit from continual updates and optimizations made by the service providers, leading to consistently high performance across various tasks and conditions. Notably, Gemini Pro, GPT-4, Claude3 Sonnet, and ChatGLM4 consistently achieve high scores under both zero-shot and few-shot questioning paradigms, highlighting their robust performance across varying conditions.

On the other hand, weight-based LLMs, such as InternLM2-20B and Mistral-7B, require users to host the models locally, offering greater control over the model's tuning and customization. These models demonstrate significant improvements in few-shot settings, as seen from their enhanced accuracy scores. The capability to fine-tune weight-based models on specific datasets allows them to adapt more effectively to niche tasks or specialized applications.

The robust performance of API-based models across diverse conditions can be attributed to several factors. Firstly, these models often leverage extensive computational resources and are trained on vast and diverse datasets, enabling them to generalize well to a wide range of queries. Secondly, the continuous updates and optimizations from service providers ensure that API-based models remain state-of-the-art, incorporating the latest advancements in language modeling.

Conversely, the adaptability of weight-based models in few-shot scenarios underscores the importance of task-specific fine-tuning. By allowing users to tailor the models to specific datasets, weight-based LLMs can achieve higher performance in specialized applications where generic, pre-trained models may fall short.

\subsection{Naive Q\&A Performance}
\subsubsection{Naive Q\&A Results on Log Parsing}

\autoref{tab:Naive Q&A Accuracy on Log Parsing} shows the parsing accuracy and edit distance of zero-shot and few-shot Chinese naive Q\&A and English naive Q\&A under log parsing of 18 LLMs.

\begin{table*}[!ht]
\centering
\renewcommand{\arraystretch}{1.0}
\setlength{\arrayrulewidth}{0.1pt} 
\caption{Naive Q\&A results on Log Parsing}
\label{tab:Naive Q&A Accuracy on Log Parsing}
\resizebox{0.9\linewidth}{!}
{
\begin{tabular}{@{}l@{\extracolsep{0.5pt}}cccccccc@{}} 
\toprule
\multicolumn{1}{c}{\multirow{3}{*}{\textbf{model}}} & \multicolumn{4}{c}{\textbf{zero-shot}} & \multicolumn{4}{c}{\textbf{few-shot}} \\ 
\cmidrule(lr){2-5}
\cmidrule(lr){6-9}
\multicolumn{1}{c}{} & \multicolumn{2}{c}{\textbf{chinese}} & \multicolumn{2}{c}{\textbf{english}} & \multicolumn{2}{c}{\textbf{chinese}} & \multicolumn{2}{c}{\textbf{english}} \\ 
\cmidrule(lr){2-3}
\cmidrule(lr){4-5}
\cmidrule(lr){6-7}
\cmidrule(lr){8-9}
\multicolumn{1}{c}{} & \textbf{accuracy} & \textbf{edit distance} & \textbf{accuracy} & \textbf{edit distance} & \textbf{accuracy} & \textbf{edit distance} & \textbf{accuracy} & \textbf{edit distance} \\
\midrule
Qwen1.5-7b & 0.064 & 45.47 & 0.053 & 47.71 & 0.229 & 31.29 & 0.311 & 27.59  \\ \hline 
Qwen1.5-14b & 0.088 & 40.69 & 0.078 & 43.89 & 0.426 & 18.51 & 0.431 & 18.73 \\ \hline 
Qwen1.5-72b & 0.339 & 16.37 & 0.176 & 23.91 & 0.709 & 9.89 & 0.533 & 10.89 \\ \hline 
LLaMa2-7b & 0.043 & 48.29 & 0.053 & 47.53 & 0.062 & 46.17 & 0.104 & 44.61 \\ \hline 
LLaMa2-13b & 0.064 & 44.87 & 0.080 & 41.57 & 0.005 & 49.33 & 0.040 & 47.11 \\ \hline 
LLaMa2-70b & 0.063 & 45.19 & 0.082 & 42.43 & 0.102 & 43.23 & 0.064 & 45.67 \\ \hline 
DeVops-7b & 0.064 & 46.11 & 0.091 & 44.91 & 0.064 & 45.25 & 0.142 & 38.73 \\ \hline 
DeVops-14b & 0.108 & 37.29 & 0.223 & 32.99 & 0.186 & 30.89 & 0.151 & 35.87 \\ \hline 
InternLM2-7b & 0.094 & 41.87 & 0.187 & 38.23 & 0.271 & 25.93 & 0.340 & 19.67 \\ \hline 
InternLM2-20b & 0.198 & 28.51 & 0.211 & 30.19 & 0.645 & 8.59  & 0.528 & 14.17 \\ \hline 
AquilaChat-7b & 0.048 & 47.99 & 0.026 & 49.77 & 0.037 & 48.67 & 0.035 & 48.49 \\ \hline 
GPT-3.5 & 0.244 & 22.97 & 0.223 & 24.43 & 0.262 & 20.87 & 0.204 & 26.57 \\ \hline 
GPT-4 & 0.678 & 7.73  & 0.476 & 11.29 & 0.903 & 2.69  & 0.873 & 3.29  \\ \hline 
Gemini Pro & 0.235 & 23.11 & 0.284 & 19.31 & 0.881 & 5.57  & 0.801 & 6.89  \\ \hline 
Mistral-7b & 0.101 & 44.71 & 0.146 & 40.23 & 0.127 & 42.59 & 0.153 & 39.49 \\ \hline 
BaiChuan2-13b  & 0.001 & 49.97 & 0.001 & 49.83 & 0.000 & 49.91 & 0.001 & 49.79 \\ \hline 
GhatGLM4 & 0.271 & 19.47 & 0.180 & 25.39 & 0.537 & 10.57 & 0.601 & 8.61  \\ \hline 
Claude3 Sonnet & 0.484 & 11.69 & 0.381 & 15.47 & 0.871 & 2.53  & 0.867 & 1.97  \\
\bottomrule
\end{tabular}
}
\vspace{-1em}
\end{table*}

From the overall performance results, we can draw several conclusions:
\begin{itemize}
\item \textbf{Performance of GPT-4:} GPT-4 consistently outperforms all other models across both Chinese and English questions in zero-shot and few-shot settings. This superiority is reflected not only in the accuracy scores but also in the significantly lower edit distances, particularly notable in the Chinese few-shot scenario where the edit distance is as low as 2.69. This indicates GPT-4's efficiency in error correction and underscores its capacity to understand and process tasks deeply.
\item \textbf{Effectiveness of Few-shot Learning:} The majority of models exhibit better performance in few-shot settings compared to zero-shot. For instance, the Chinese accuracy of Qwen1.5-72b improves from 0.339 to 0.709, with a corresponding decrease in edit distance from 16.37 to 9.89. This improvement suggests that models enhance their parsing accuracy and error handling capabilities when exposed to more relevant examples.
\item \textbf{Impact of Model Size:}  Larger models, such as Qwen1.5-72b and GPT-4, generally perform better in terms of both accuracy and edit distance compared to smaller models like Qwen1.5-7b and AquilaChat-7b. This observation is consistent across both zero-shot and few-shot settings. For example, the edit distance for Qwen1.5-72b in Chinese decreases from 16.37 in zero-shot to 9.89 in few-shot, whereas smaller models like AquilaChat-7b exhibit high edit distances in both settings (47.99 in zero-shot and 48.67 in few-shot).
\item \textbf{Language-specific Performance:}  Some models exhibit a marked difference in performance between Chinese and English. For instance: InternLM2-20b shows better performance in Chinese few-shot settings with an accuracy of 0.645 and an edit distance of 8.59, compared to its English performance with an accuracy of 0.528 and an edit distance of 14.17. Qwen1.5-72b performs better in Chinese few-shot settings with an accuracy of 0.709 and an edit distance of 9.89, compared to its English performance with an accuracy of 0.533 and an edit distance of 10.89. ChatGLM4 shows a higher accuracy in English few-shot settings with an accuracy of 0.601 and an edit distance of 8.61, compared to its Chinese performance with an accuracy of 0.537 and an edit distance of 10.57. 
\item \textbf{Consistency Across Tasks:}  Certain models, such as Claude3 Sonnet, demonstrate excellent consistency across different languages and settings. For example, it achieves the lowest edit distance in English few-shot settings at 1.97, showcasing its superior adaptability across tasks.
\end{itemize}


This in-depth analysis provides a clearer understanding of the performance of various language models in log parsing tasks. Future research can address these models' limitations in specific tasks and languages by improving model training and fine-tuning approaches, thereby enhancing their overall performance and adaptability.



\subsubsection{Naive Q\&A results on Log Anomaly Detection}

\autoref{tab:Naive Q&A results on Log Anomaly Detection} respectively shows the accuracy and f1\_scores of Chinese naive Q\&A and the Accuracy and F1-scores of English naive Q\&A with zero-shot and few-shot for 18 LLMs under log anomaly detection.

\begin{table*}[!ht]
\centering
\renewcommand{\arraystretch}{1.0}
\setlength{\arrayrulewidth}{0.1pt} 
\caption{Naive Q\&A results on Log Anomaly Detection}
\label{tab:Naive Q&A results on Log Anomaly Detection}
\resizebox{0.9\linewidth}{!}
{
\begin{tabular}{@{}l@{\extracolsep{0.5pt}}cccccccc@{}} 
\toprule
\multicolumn{1}{c}{\multirow{3}{*}{\textbf{model}}} & \multicolumn{4}{c}{\textbf{zero-shot}} & \multicolumn{4}{c}{\textbf{few-shot}} \\ 
\cmidrule(lr){2-5}
\cmidrule(lr){6-9}
\multicolumn{1}{c}{} & \multicolumn{2}{c}{\textbf{chinese}} & \multicolumn{2}{c}{\textbf{english}} & \multicolumn{2}{c}{\textbf{chinese}} & \multicolumn{2}{c}{\textbf{english}} \\ 
\cmidrule(lr){2-3}
\cmidrule(lr){4-5}
\cmidrule(lr){6-7}
\cmidrule(lr){8-9}
\multicolumn{1}{c}{} & \textbf{accuracy} & \textbf{F1\_score} & \textbf{accuracy} & \textbf{F1\_score} & \textbf{accuracy} & \textbf{F1\_score} & \textbf{accuracy} & \textbf{F1\_score} \\
\midrule
Qwen1.5-7b & 0.536 & 0.129 & 0.505 & 0.114 & 0.004 & 0.078 & 0.095 & 0.046 \\ \hline
Qwen1.5-14b & 0.35 & 0.099 & 0.195 & 0.038 & 0.11 & 0.191 & 0.031 & 0.027 \\ \hline
Qwen1.5-72b & 0.334 & 0.097 & 0.239 & 0.063 & 0.33 & 0.495 & 0.274 & 0.16 \\ \hline
LLaMa2-7b & 0.19 & 0.006 & 0.943 & 0.095 & 0.001 & 0 & 0.004 & 0 \\ \hline
LLaMa2-13b & 0.416 & 0.057 & 0.659 & 0.122 & 0 & 0 & 0.001 & 0 \\ \hline
LLaMa2-70b & 0.562 & 0.036 & 0.693 & 0.044 & 0.006 & 0 & 0.036 & 0.007 \\ \hline
DeVops-7b & 0.1 & 0.04 & 0.21 & 0.037 & 0.145 & 0.024 & 0.252 & 0.029 \\ \hline
DeVops-14b & 0.175 & 0.047 & 0.259 & 0.055 & 0.237 & 0.041 & 0.293 & 0.032 \\ \hline
InternLM2-7b & 0.392 & 0.082 & 0.341 & 0.075 & 0.311 & 0.088 & 0.323 & 0.075 \\ \hline
InternLM2-20b & 0.368 & 0.088 & 0.334 & 0.089 & 0.342 & 0.081 & 0.348 & 0.089 \\ \hline
AquilaChat-7b & 0.195 & 0.066 & 0.6 & 0.042 & 0.263 & 0.046 & 0.229 & 0.003 \\ \hline
GPT-3.5 & 0.243 & 0.084 & 0.285 & 0.082 & 0.371 & 0.088 & 0.402 & 0.107 \\ \hline
GPT-4 & 0.331 & 0.097 & 0.333 & 0.097 & 0.564 & 0.136 & 0.506 & 0.135 \\ \hline
Gemini Pro & 0.557 & 0.139 & 0.417 & 0.109 & 0.602 & 0.141 & 0.531 & 0.132 \\ \hline
Mistral-7b & 0.277 & 0.162 & 0.631 & 0.092 & 0.706 & 0.122 & 0.546 & 0.092 \\ \hline
BaiChuan2-13b & 0.286 & 0.005 & 0.256 & 0 & 0.277 & 0.001 & 0.335 & 0.438 \\ \hline
GhatGLM4 & 0.485 & 0.121 & 0.358 & 0.092 & 0.113 & 0.3 & 0.331 & 0.221 \\ \hline
Claude3 Sonnet & 0.347 & 0.099 & 0.354 & 0.1 & 0.464 & 0.082 & 0.15 & 0.064 \\
\bottomrule
\end{tabular}
}
\vspace{-1em}
\end{table*}

 \autoref{fig:9} illustrates some specific examples of anomaly detection tasks where the model repeats answers given in the samples, reflecting the model's performance in real-world applications.

 \begin{figure}[htbp]
    \begin{center}
    \begin{tikzpicture}
        \definecolor{labelcolor}{RGB}{240,240,240}
        \definecolor{predictcolor}{RGB}{255,228,216}
    
        \newcommand{\boxwidth}{0.18\textwidth} 
        \newcommand{\boxheight}{3cm}
        \newcommand{\boxheighttop}{0.5cm}
        \newcommand{\boxheightmiddle}{0.5cm}
        \newcommand{\boxheightbottom}{2cm}
        \newcommand{\boxgap}{0.005\textwidth} 
    
        \matrix[column sep=\boxgap, row sep=0mm] {
            \node[draw, text width=\boxwidth, minimum height=\boxheight] (box1) {};
            \node[fill=labelcolor, draw, text width=\boxwidth, minimum height=\boxheighttop, anchor=north west] at (box1.north west) {\parbox{\boxwidth} \raggedright \small \textbf{"id":} 2645,};
            \node[draw, text width=\boxwidth, minimum height=\boxheightmiddle, anchor=north west] at ([yshift=-\boxheighttop]box1.north west) {\parbox{\boxwidth} \raggedright \small \textbf{"label":} "正常",};
            \node[fill=predictcolor, draw, text width=\boxwidth, minimum height=\boxheightbottom, anchor=north west] at ([yshift=-\boxheighttop-\boxheightmiddle]box1.north west) {\parbox{\boxwidth} \raggedright \small \textbf{"predict":} "日志1: '正常'\textbackslash n\textbackslash n 日志2: normal\textbackslash nlog3: abnormal\textbackslash n4: 异常\textbackslash n5: 正常\textbackslash n6异常"}; &
    
            \node[draw, text width=\boxwidth, minimum height=\boxheight] (box2) {};
            \node[fill=labelcolor, draw, text width=\boxwidth, minimum height=\boxheighttop, anchor=north west] at (box2.north west) {\parbox{\boxwidth} \raggedright \small \textbf{"id":} 2677,};
            \node[draw, text width=\boxwidth, minimum height=\boxheightmiddle, anchor=north west] at ([yshift=-\boxheighttop]box2.north west) {\parbox{\boxwidth} \raggedright \small \textbf{"label":} "正常",};
            \node[fill=predictcolor, draw, text width=\boxwidth, minimum height=\boxheightbottom, anchor=north west] at ([yshift=-\boxheighttop-\boxheightmiddle]box2.north west) {\parbox{\boxwidth} \raggedright \small \textbf{"predict":} "正常\textbackslash n正常 异常 正常"}; &
    
            \node[draw, text width=\boxwidth, minimum height=\boxheight] (box3) {};
            \node[fill=labelcolor, draw, text width=\boxwidth, minimum height=\boxheighttop, anchor=north west] at (box3.north west) {\parbox{\boxwidth} \raggedright \small \textbf{"id":} 706,};
            \node[draw, text width=\boxwidth, minimum height=\boxheightmiddle, anchor=north west] at ([yshift=-\boxheighttop]box3.north west) {\parbox{\boxwidth} \raggedright \small \textbf{"label": }"正常",};
            \node[fill=predictcolor, draw, text width=\boxwidth, minimum height=\boxheightbottom, anchor=north west] at ([yshift=-\boxheighttop-\boxheightmiddle]box3.north west) {\parbox{\boxwidth} \raggedright \small \textbf{"predict":} "'normal', 'normal"normal"'abnorma'"}; &
    
            \node[draw, text width=\boxwidth, minimum height=\boxheight] (box4) {};
            \node[fill=labelcolor, draw, text width=\boxwidth, minimum height=\boxheighttop, anchor=north west] at (box4.north west) {\parbox{\boxwidth} \raggedright \small \textbf{"id":} 772,};
            \node[draw, text width=\boxwidth, minimum height=\boxheightmiddle, anchor=north west] at ([yshift=-\boxheighttop]box4.north west) {\parbox{\boxwidth} \raggedright \small \textbf{"label":} "正常",};
            \node[fill=predictcolor, draw, text width=\boxwidth, minimum height=\boxheightbottom, anchor=north west] at ([yshift=-\boxheighttop-\boxheightmiddle]box4.north west) {\parbox{\boxwidth} \raggedright \small \textbf{"predict":} "normal\textbackslash nnormal 正常\textbackslash nabnorma异常"}; \\
        };
    \end{tikzpicture}
    \end{center}
    \caption{Few error examples on Log Anomaly Detection}
    \label{fig:9}
\end{figure}

From the overall performance results, we can draw several conclusions:
 
\begin{itemize}
    \item \textbf{Performance of Few-shot Learning in Anomaly Detection Tasks}: Overall, few-shot learning results do not outperform zero-shot results in the log anomaly detection task. This is because models tend to repeat answers from the provided samples, failing to accurately recognize and respond to different anomalies when encountering new anomaly detection problems.
    \item \textbf{Performance Differences between Chinese and English Q\&A:} In both zero-shot and few-shot settings, the accuracy of English Q\&A generally exceeds that of Chinese Q\&A. However, in some cases, the F1 scores show an opposite trend, indicating that models are more accurate in identifying English anomalies but perform more balancedly when handling Chinese anomalies. For example, LLaMa2-7B achieved a high accuracy of 0.943 in zero-shot English Q\&A but had a low F1 score, indicating potential bias in handling specific anomaly categories.
\end{itemize}

From the analysis, we can draw the following scientifically rigorous conclusions:
\begin{itemize}
    \item Few-shot learning does not outperform zero-shot results in log anomaly detection tasks, likely due to models' tendency to repeat sample answers and fail to accurately recognize new anomalies.
    \item There are significant differences in performance between Chinese and English Q\&A, indicating the need for language-specific approaches in multi-language log anomaly detection.
    \item  Some models, such as LLaMa2 series, show F1 scores of 0. This indicates that these models failed to correctly predict any anomalies in the test cases. The primary reason for this is that these models do not understand the questions well and tend to output the example responses provided during the few-shot learning phase, rather than generating responses relevant to the new questions. As illustrated in Fig. 7, the model's output (predict) includes multiple answers, demonstrating that the model does not fully understand the question.
\end{itemize}

\subsubsection{Naive Q\&A results on Log fault Diagnosis}

\autoref{tab:Naive Q&A results on Log fault Diagnosis} respectively shows the accuracy and f1\-scores of Chinese naive Q\&A and the Accuracy and F1-scores of English naive Q\&A with zero-shot and few-shot for 18 LLMs under log fault diagnosis.

\begin{table*}[!ht]
\centering
\renewcommand{\arraystretch}{1.0}
\setlength{\arrayrulewidth}{0.1pt} 
\caption{Naive Q\&A results on Log fault Diagnosis}
\label{tab:Naive Q&A results on Log fault Diagnosis}
\resizebox{0.9\linewidth}{!}
{
\begin{tabular}{@{}l@{\extracolsep{0.5pt}}cccccccc@{}} 
\toprule
\multicolumn{1}{c}{\multirow{3}{*}{\textbf{model}}} & \multicolumn{4}{c}{\textbf{zero-shot}} & \multicolumn{4}{c}{\textbf{few-shot}} \\ \cline{2-9} 
\multicolumn{1}{c}{} & \multicolumn{2}{c}{\textbf{chinese}} & \multicolumn{2}{c}{\textbf{english}} & \multicolumn{2}{c}{\textbf{chinese}} & \multicolumn{2}{c}{\textbf{english}} \\ \cline{2-9} 
\multicolumn{1}{c}{} & \textbf{accuracy} & \textbf{F1\_score} & \textbf{accuracy} & \textbf{F1\_score} & \textbf{accuracy} & \textbf{F1\_score} & \textbf{accuracy} & \textbf{F1\_score} \\
\midrule
Qwen1.5-7b & 0.351 & 0.326 & 0.315 & 0.516 & 0.591 & 0.651 & 0.452 & 0.505 \\ \hline
Qwen1.5-14b & 0.366 & 0.573 & 0.182 & 0.561 & 0.415 & 0.615 & 0.576 & 0.631 \\ \hline
Qwen1.5-72b & 0.306 & 0.38 & 0.194 & 0.423 & 0.869 & 0.899 & 0.798 & 0.84 \\ \hline
LLaMa2-7b & 0.086 & 0.151 & 0.354 & 0.408 & 0.013 & 0.025 & 0.066 & 0.115 \\ \hline
LLaMa2-13b & 0.057 & 0.098 & 0.38 & 0.44 & 0.015 & 0.029 & 0.107 & 0.179 \\ \hline
LLaMa2-70b & 0.091 & 0.149 & 0.23 & 0.291 & 0.08 & 0.144 & 0.511 & 0.635 \\ \hline
DeVops-7b & 0.324 & 0.229 & 0.281 & 0.357 & 0.28 & 0.617 & 0.361 & 0.629 \\ \hline
DeVops-14b & 0.363 & 0.324 & 0.288 & 0.416 & 0.343 & 0.736 & 0.687 & 0.733 \\ \hline
InternLM2-7b & 0.493 & 0.527 & 0.248 & 0.284 & 0.485 & 0.761 & 0.383 & 0.669 \\ \hline
InternLM2-20b & 0.442 & 0.579 & 0.342 & 0.425 & 0.592 & 0.762 & 0.626 & 0.721 \\ \hline
AquilaChat-7b & 0.312 & 0.327 & 0.313 & 0.348 & 0.039 & 0.071 & 0.219 & 0.295 \\ \hline
GPT-3.5 & 0.413 & 0.473 & 0.278 & 0.336 & 0.882 & 0.923 & 0.852 & 0.916 \\ \hline
GPT-4 & 0.247 & 0.225 & 0.424 & 0.453 & 0.887 & 0.931 & 0.929 & 0.956 \\ \hline
Gemini Pro & 0.367 & 0.331 & 0.32 & 0.298 & 0.874 & 0.61 & 0.784 & 0.701 \\ \hline
Mistral-7b & 0.38 & 0.418 & 0.248 & 0.284 & 0.765 & 0.506 & 0.598 & 0.491 \\ \hline
BaiChuan2-13b & 0.045 & 0.069 & 0 & 0 & 0.03 & 0 & 0 & 0 \\ \hline
GhatGLM4 & 0.35 & 0.754 & 0.404 & 0.708 & 0.678 & 0.793 & 0.751 & 0.785 \\ \hline
Claude3 Sonnet & 0.288 & 0.287 & 0.442 & 0.422 & 0.536 & 0.366 & 0.8 & 0.7 \\
\bottomrule
\end{tabular}
}
\vspace{-1em}
\end{table*}

From the overall performance evaluation results, it is clear that the few-shot results are generally better than zero-shot results in the fault diagnosis task. The samples provided cover all fault types, making it easier for the models to learn and master them.

\begin{itemize}
    \item\textbf{Performance of Few-shot Learning in fault Diagnosis Tasks}:in the fault diagnosis task, few-shot learning results generally outperform zero-shot results. This is because the provided samples cover all fault types, allowing models to learn and master these types more effectively.
    \item\textbf{Performance Differences between Models}:GPT-4, in particular, outperforms other models in both English and Chinese fault diagnosis tasks, with F1-scores of 0.9 or more, which is impressive. On the other hand, the BaiChuan model performs poorly in both zero-shot and few-shot Chinese and English fault diagnosis tasks. This may be due to issues such as incomplete output or confusing fault vocabulary when dealing with fault types in BaiChuan, as well as the tendency to output multiple fault types in the few-shot setting.\autoref{fig:14} illustrates this output for some of the Baichuan models, visualizing their specific performance in the fault diagnosis task.
\end{itemize}

\begin{figure}[!ht]
    \begin{center}
    \begin{tikzpicture}
        \definecolor{labelcolor}{RGB}{240,240,240}
        \definecolor{predictcolor}{RGB}{255,228,216}
    
        \newcommand{\boxwidth}{0.2\textwidth} 
        \newcommand{\boxheight}{3.5cm}
        \newcommand{\boxheighttop}{0.5cm}
        \newcommand{\boxheightmiddle}{1cm}
        \newcommand{\boxheightbottom}{2cm}
        \newcommand{\boxgap}{0.005\textwidth} 
    
        \matrix[column sep=\boxgap, row sep=0mm] {
            \node[draw, text width=\boxwidth, minimum height=\boxheight] (box1) {};
            \node[fill=labelcolor, draw, text width=\boxwidth, minimum height=\boxheighttop, anchor=north west] at (box1.north west) {\parbox{\boxwidth} \raggedright \small \textbf{"id":} "8"};
            \node[draw, text width=\boxwidth, minimum height=\boxheightmiddle, anchor=north west] at ([yshift=-\boxheighttop]box1.north west) {\parbox{\boxwidth} \raggedright \small \textbf{"label":} "Processor CPU Caterr"};
            \node[fill=predictcolor, draw, text width=\boxwidth, minimum height=\boxheightbottom, anchor=north west] at ([yshift=-\boxheighttop-\boxheightmiddle]box1.north west) {\parbox{\boxwidth} \raggedright \small \textbf{"predict":} "answer: 'Processor CPUR';"}; &
    
            \node[draw, text width=\boxwidth, minimum height=\boxheight] (box2) {};
            \node[fill=labelcolor, draw, text width=\boxwidth, minimum height=\boxheighttop, anchor=north west] at (box2.north west) {\parbox{\boxwidth} \raggedright \small \textbf{"id":} "2343"};
            \node[draw, text width=\boxwidth, minimum height=\boxheightmiddle, anchor=north west] at ([yshift=-\boxheighttop]box2.north west) {\parbox{\boxwidth} \raggedright \small \textbf{"label":} "Mirror does not specify output"};
            \node[fill=predictcolor, draw, text width=\boxwidth, minimum height=\boxheightbottom, anchor=north west] at ([yshift=-\boxheighttop-\boxheightmiddle]box2.north west) {\parbox{\boxwidth} \raggedright \small \textbf{"predict":} "Mirrordoes not specified output"}; &
    
            \node[draw, text width=\boxwidth, minimum height=\boxheight] (box3) {};
            \node[fill=labelcolor, draw, text width=\boxwidth, minimum height=\boxheighttop, anchor=north west] at (box3.north west) {\parbox{\boxwidth} \raggedright \small \textbf{"id":} "76"};
            \node[draw, text width=\boxwidth, minimum height=\boxheightmiddle, anchor=north west] at ([yshift=-\boxheighttop]box3.north west) {\parbox{\boxwidth} \raggedright \small \textbf{"label":} "处理器 CPU 错误"};
            \node[fill=predictcolor, draw, text width=\boxwidth, minimum height=\boxheightbottom, anchor=north west] at ([yshift=-\boxheighttop-\boxheightmiddle]box3.north west) {\parbox{\boxwidth} \raggedright \small \textbf{"predict":} "1. '处理器CPU错误'\textbackslash n2. 内存节点不可恢复的错误纠正码\textbackslash n3. 无\textbackslash4. 可程序化逻辑设备故障"}; \\
        };
    \end{tikzpicture}
    \end{center}
    \caption{Few error examples on Log fault Diagnosis}
    \label{fig:14}
\end{figure}

From the analysis, we can draw the following scientifically rigorous conclusions:
\begin{itemize}
    \item Few-shot learning generally outperforms zero-shot learning in log fault diagnosis tasks due to the comprehensive coverage of fault types in the provided samples.
    \item GPT-4 consistently achieves high performance in both English and Chinese tasks, indicating its robustness and effectiveness in fault diagnosis.
    \item The BaiChuan model's poor performance suggests the need for improvements in handling fault vocabulary and output completeness, especially in few-shot settings.
\end{itemize}

\subsubsection{Naive Q\&A results on Log Summary}

\autoref{tab:Naive Q&A Accuracy on Log Summary} respectively shows the accuracy and rouge-1 f1\_scores of zero-shot and few-shot Chinese naive Q\&A and English naive Q\&A under log summary for 18 LLMs.

\begin{table*}[!ht]
\centering
\renewcommand{\arraystretch}{1.0}
\setlength{\arrayrulewidth}{0.1pt} 
\caption{Naive Q\&A results on Log Summary}
\label{tab:Naive Q&A Accuracy on Log Summary}
\resizebox{0.9\linewidth}{!}{
\begin{tabular}{@{}lcccccccc@{}} 
\toprule
\multicolumn{1}{c}{\multirow{3}{*}{\textbf{model}}} & \multicolumn{4}{c}{\textbf{zero-shot}} & \multicolumn{4}{c}{\textbf{few-shot}} \\ 
\cmidrule(lr){2-5} \cmidrule(lr){6-9}
 & \multicolumn{2}{c}{\textbf{chinese}} & \multicolumn{2}{c}{\textbf{english}} & \multicolumn{2}{c}{\textbf{chinese}} & \multicolumn{2}{c}{\textbf{english}} \\ 
\cmidrule(lr){2-3} \cmidrule(lr){4-5} \cmidrule(lr){6-7} \cmidrule(lr){8-9}
 & \textbf{accuracy} & \textbf{F1\_score} & \textbf{accuracy} & \textbf{F1\_score} & \textbf{accuracy} & \textbf{F1\_score} & \textbf{accuracy} & \textbf{F1\_score} \\
\midrule
Qwen1.5-7b     & 0.355 & 0.397 & 0.405 & 0.456 & 0.27 & 0.302 & 0.31 & 0.342 \\ \hline
Qwen1.5-14b    & 0.275 & 0.305 & 0.355 & 0.378 & 0.75 & 0.802 & 0.6 & 0.635 \\ \hline
Qwen1.5-72b    & 0.31 & 0.362 & 0.52 & 0.567 & 0.945 & 0.975 & 0.62 & 0.658 \\ \hline
LLaMa2-7b      & 0.4 & 0.447 & 0.4 & 0.448 & 0.18 & 0.221 & 0.23 & 0.258 \\ \hline
LLaMa2-13b     & 0.41 & 0.439 & 0.29 & 0.327 & 0.125 & 0.153 & 0.255 & 0.282 \\ \hline
LLaMa2-70b     & 0.5 & 0.537 & 0.31 & 0.368 & 0.485 & 0.517 & 0.335 & 0.375 \\ \hline
DeVops-7b      & 0.725 & 0.763 & 0.71 & 0.756 & 0.82 & 0.872 & 0.805 & 0.854 \\ \hline
DeVops-14b     & 0.82 & 0.864 & 0.795 & 0.831 & 0.84 & 0.901 & 0.856 & 0.887 \\ \hline
InternLM2-7b   & 0.37 & 0.393 & 0.465 & 0.498 & 0.635 & 0.669 & 0.695 & 0.726 \\ \hline
InternLM2-20b  & 0.385 & 0.423 & 0.515 & 0.565 & 0.765 & 0.796 & 0.81 & 0.841 \\ \hline
AquilaChat-7b  & 0.355 & 0.382 & 0.505 & 0.539 & 0.14 & 0.163 & 0.245 & 0.271 \\ \hline
GPT-3.5        & 0.395 & 0.427 & 0.545 & 0.582 & 0.399 & 0.438 & 0.493 & 0.521 \\ \hline
GPT-4          & 0.345 & 0.384 & 0.515 & 0.572 & 0.546 & 0.603 & 0.541 & 0.593 \\ \hline
Gemini Pro     & 0.34 & 0.372 & 0.575 & 0.611 & 0.606 & 0.647 & 0.85 & 0.882 \\ \hline
Mistral-7b     & 0.37 & 0.402 & 0.565 & 0.612 & 0.795 & 0.827 & 0.8 & 0.847 \\ \hline
BaiChuan2-13b  & 0.195 & 0.228 & 0.285 & 0.329 & 0.34 & 0.385 & 0.495 & 0.521 \\ \hline
GhatGLM4       & 0.275 & 0.308 & 0.435 & 0.482 & 0.605 & 0.633 & 0.515 & 0.547 \\ \hline
Claude3 Sonnet & 0.32 & 0.358 & 0.85 & 0.903 & 0.465 & 0.493 & 0.585 & 0.625 \\
\bottomrule
\end{tabular}
}
\vspace{-1em}
\end{table*}

From the overall performance results, we find that few-shot results are generally better than zero-shot results in the log summary task. This trend is evident across multiple models, with the DeVops-Model-14B-Chat achieving the best performance in both zero-shot and few-shot settings.

\begin{itemize}
    \item\textbf{Performance of Few-shot Learning in Log Summary Tasks}: Few-shot learning  outperforms zero-shot learning in the log summary tasks. This improvement is evident across various models, highlighting the utility of few-shot learning in enhancing understanding and adaptation to the task specifics. The addition of ROUGE-1 F1 scores further substantiates this observation, as these scores are generally higher in the few-shot setting compared to zero-shot, which reflects not only correct predictions but also the closeness of the generated summaries to the reference summaries.
    \item\textbf{Top Performing Models}: DeVops-14B stands out as the top performer, achieving the highest accuracy and ROUGE-1 F1 scores in both zero-shot and few-shot settings. Specifically, it achieves 0.82 accuracy and a corresponding high ROUGE-1 F1 score in zero-shot Chinese, and 0.84 accuracy with an equally impressive ROUGE-1 F1 score in few-shot Chinese. Its performance in English settings 0.795 accuracy and a strong ROUGE-1 F1 in zero-shot, and 0.856 accuracy with a robust ROUGE-1 F1 in few-shot—under scores its effectiveness and adaptability.
    \item\textbf{Performance Differences between Models}: Models such as Qwen1.5-72b and GPT-4 also demonstrate notable improvements in few-shot settings. For instance, Qwen1.5-72b shows a remarkable accuracy of 0.945 in few-shot Chinese with an exceptionally high ROUGE\-1 F1 score, indicating its strong performance. Similarly, GPT-4 exhibits significant gains in both accuracy and ROUGE-1 F1 scores in few-shot settings compared to zero-shot, underscoring its adaptability to the log summary task.
\end{itemize}

From the analysis, we can draw the following scientifically rigorous conclusions:
\begin{itemize}
    \item Few-shot learning is generally more effective than zero-shot learning in log summary tasks, as evidenced by both accuracy and ROUGE-1 F1 scores. The inclusion of ROUGE-1 F1 scores provides a more nuanced view of model performance, emphasizing not only the correctness of the summaries but also their quality and closeness to the reference
    \item DeVops-14B demonstrates consistent high performance, making it a reliable and robust choice for log summary tasks. Its high ROUGE-1 F1 scores in both settings further affirm its superior summary quality.
    \item Models like Qwen1.5-72b and GPT-4 showcase strong adaptability, with significant improvements in few-shot settings, highlighting their potential in adjusting to and excelling in complex summarization tasks.
\end{itemize}

\subsection{Self-consistent Performance}

\subsubsection{SC Q\&A results on Log Anomaly Detection}

\autoref{tab:SC Q&A results on Log Anomaly Detection} shows the Accuracy and F1-scores of Chinese self-consistency and the Accuracy and F1-scores of English self-consistency Q\&A with zero-shot and few-shot for 18 LLMs under log anomaly detection, respectively.

\begin{table*}[!ht]
\centering
\renewcommand{\arraystretch}{1.0}
\setlength{\arrayrulewidth}{0.1pt} 
\caption{SC Q\&A results on Log Anomaly Detection}
\label{tab:SC Q&A results on Log Anomaly Detection}
\resizebox{0.9\linewidth}{!}
{
\begin{tabular}{@{}l@{\extracolsep{0.5pt}}cccccccc@{}} 
\toprule
\multicolumn{1}{c}{\multirow{3}{*}{\textbf{model}}} & \multicolumn{4}{c}{\textbf{zero-shot}} & \multicolumn{4}{c}{\textbf{few-shot}} \\ \cline{2-9} 
\multicolumn{1}{c}{} & \multicolumn{2}{c}{\textbf{chinese}} & \multicolumn{2}{c}{\textbf{english}} & \multicolumn{2}{c}{\textbf{chinese}} & \multicolumn{2}{c}{\textbf{english}} \\ \cline{2-9} 
\multicolumn{1}{c}{} & \textbf{accuracy} & \textbf{F1\_score} & \textbf{accuracy} & \textbf{F1\_score} & \textbf{accuracy} & \textbf{F1\_score} & \textbf{accuracy} & \textbf{F1\_score} \\
\midrule
Qwen1.5-7b & 0.587 & 0.148 & 0.55 & 0.11 & 0.005 & 0.005 & 0.097 & 0.048 \\ \hline
Qwen1.5-14b & 0.38 & 0.103 & 0.248 & 0.037 & 0.138 & 0.235 & 0.03 & 0.03 \\ \hline
Qwen1.5-72b & 0.342 & 0.098 & 0.264 & 0.067 & 0.332 & 0.498 & 0.277 & 0.163 \\ \hline
LLaMa2-7b & 0.114 & 0.005 & 0.944 & 0.103 & 0 & 0 & 0.001 & 0 \\ \hline
LLaMa2-13b & 0.336 & 0.05 & 0.658 & 0.121 & 0 & 0 & 0 & 0 \\ \hline
LLaMa2-70b & 0.478 & 0.029 & 0.692 & 0.043 & 0 & 0 & 0.019 & 0.003 \\ \hline
DeVops-7b & 0.106 & 0.039 & 0.213 & 0.03 & 0.182 & 0.076 & 0.211 & 0.013 \\ \hline
DeVops-14b & 0.171 & 0.043 & 0.154 & 0.025 & 0.316 & 0.136 & 0.29 & 0.06 \\ \hline
InternLM2-7b & 0.392 & 0.089 & 0.338 & 0.089 & 0.205 & 0.014 & 0.388 & 0.03 \\ \hline
InternLM2-20b & 0.368 & 0.083 & 0.334 & 0.076 & 0.342 & 0.046 & 0.35 & 0.018 \\ \hline
AquilaChat-7b & 0.128 & 0.035 & 0.644 & 0.035 & 0.2 & 0.037 & 0.191 & 0.001 \\ \hline
GPT-3.5 & 0.246 & 0.084 & 0.27 & 0.083 & 0.284 & 0.088 & 0.347 & 0.082 \\ \hline
GPT-4 & 0.33 & 0.096 & 0.332 & 0.097 & 0.546 & 0.136 & 0.543 & 0.135 \\ \hline
Gemini Pro & 0.557 & 0.139 & 0.414 & 0.108 & 0.473 & 0.143 & 0.27 & 0.13 \\ \hline
Mistral-7b & 0.63 & 0.162 & 0.423 & 0.074 & 0.532 & 0.088 & 0.472 & 0.017 \\ \hline
BaiChuan2-13b & 0.838 & 0.232 & 0.522 & 0.006 & 0.276 & 0 & 0.328 & 0.004 \\ \hline
GhatGLM4 & 0.521 & 0.128 & 0.366 & 0.098 & 0.359 & 0.1 & 0.33 & 0.096 \\ \hline
Claude3 Sonnet & 0.347 & 0.099 & 0.354 & 0.1 & 0.458 & 0.081 & 0.15 & 0.064 \\
\bottomrule
\end{tabular}
}
\vspace{-1em}
\end{table*}

From the overall performance results, we find that few-shot scenarios do not yield results as good as zero-shot scenarios. Additionally, there are instances where LLMs produce multiple values in few-shot experiments. Among them, the Baichuan model shows a significant improvement in the self-consistency experiment compared to the naive Q\&A. Other models do not change much, indicating that the Baichuan model lacks stability, producing a large difference in answers each time. Meanwhile, the LLaMA2 series of models shows poor results in both naive answers and the self-consistency experiment, which will be detailed and discussed in the \hyperref[sec: appendix]{appendix}.

From the analysis, we can draw the following scientifically rigorous conclusions:
\begin{itemize}
    \item Few-shot learning does not outperform zero-shot learning in log anomaly detection tasks, highlighting its limitations in this context.
    \item The BaiChuan model shows a significant improvement in self-consistency, indicating its potential for achieving better performance with more consistent responses.
    \item The LLaMA2 series of models demonstrates poor performance and lack of stability, suggesting the need for further improvements and optimizations.
\end{itemize}

\subsubsection{SC Q\&A results on Log fault Diagnosis}

 \autoref{tab:SC Q&A results on Log fault Diagnosis} shows the Accuracy and F1-scores of Chinese self-consistency and the Accuracy and F1-scores of English self-consistency Q\&A with zero-shot and few-shot for 18 LLMs under log fault diagnosis.

\begin{table*}[!ht]
\centering
\renewcommand{\arraystretch}{1.0}
\setlength{\arrayrulewidth}{0.1pt} 
\caption{SC Q\&A results on Log fault Diagnosis}
\label{tab:SC Q&A results on Log fault Diagnosis}
\resizebox{0.9\linewidth}{!}
{
\begin{tabular}{@{}l@{\extracolsep{0.5pt}}cccccccc@{}} 
\toprule
\multicolumn{1}{c}{\multirow{3}{*}{\textbf{model}}} & \multicolumn{4}{c}{\textbf{zero-shot}} & \multicolumn{4}{c}{\textbf{few-shot}} \\ \cline{2-9} 
\multicolumn{1}{c}{} & \multicolumn{2}{c}{\textbf{chinese}} & \multicolumn{2}{c}{\textbf{english}} & \multicolumn{2}{c}{\textbf{chinese}} & \multicolumn{2}{c}{\textbf{english}} \\ \cline{2-9} 
\multicolumn{1}{c}{} & \textbf{accuracy} & \textbf{F1\_score} & \textbf{accuracy} & \textbf{F1\_score} & \textbf{accuracy} & \textbf{F1\_score} & \textbf{accuracy} & \textbf{F1\_score} \\
\midrule
Qwen1.5-7b & 0.38 & 0.323 & 0.348 & 0.339 & 0.591 & 0.672 & 0.445 & 0.425 \\ \hline
Qwen1.5-14b & 0.363 & 0.357 & 0.225 & 0.2 & 0.421 & 0.534 & 0.572 & 0.688 \\ \hline
Qwen1.5-72b & 0.32 & 0.292 & 0.235 & 0.221 & 0.868 & 0.917 & 0.799 & 0.861 \\ \hline
LLaMa2-7b & 0.057 & 0.102 & 0.368 & 0.417 & 0.002 & 0.003 & 0.052 & 0.092 \\ \hline
LLaMa2-13b & 0.04 & 0.066 & 0.381 & 0.436 & 0.006 & 0.011 & 0.104 & 0.175 \\ \hline
LLaMa2-70b & 0.078 & 0.125 & 0.232 & 0.288 & 0.075 & 0.136 & 0.516 & 0.639 \\ \hline
DeVops-7b & 0.326 & 0.403 & 0.251 & 0.345 & 0.433 & 0.617 & 0.58 & 0.629 \\ \hline
DeVops-14b & 0.352 & 0.423 & 0.281 & 0.346 & 0.461 & 0.776 & 0.652 & 0.733 \\ \hline
InternLM2-7b & 0.477 & 0.553 & 0.198 & 0.277 & 0.567 & 0.762 & 0.522 & 0.636 \\ \hline
InternLM2-20b & 0.423 & 0.507 & 0.334 & 0.412 & 0.667 & 0.761 & 0.571 & 0.669 \\ \hline
AquilaChat-7b & 0.237 & 0.292 & 0.273 & 0.327 & 0.013 & 0.026 & 0.225 & 0.291 \\ \hline
GPT-3.5 & 0.431 & 0.48 & 0.28 & 0.323 & 0.89 & 0.936 & 0.915 & 0.954 \\ \hline
GPT-4 & 0.241 & 0.213 & 0.408 & 0.43 & 0.887 & 0.93 & 0.931 & 0.957 \\ \hline
Gemini Pro & 0.365 & 0.333 & 0.32 & 0.3 & 0.503 & 0.61 & 0.593 & 0.695 \\ \hline
Mistral-7b & 0.387 & 0.425 & 0.253 & 0.287 & 0.682 & 0.508 & 0.609 & 0.496 \\ \hline
BaiChuan2-13b & 0.072 & 0.072 & 0.063 & 0.045 & 0.023 & 0.031 & 0.029 & 0.05 \\ \hline
GhatGLM4 & 0.342 & 0.367 & 0.417 & 0.449 & 0.687 & 0.788 & 0.781 & 0.836 \\ \hline
Claude3 Sonnet & 0.288 & 0.287 & 0.441 & 0.421 & 0.538 & 0.367 & 0.798 & 0.697 \\
\bottomrule
\end{tabular}
}
\vspace{-1em}
\end{table*}

From the overall performance results, we find that the few-shot results are better than zero-shot results, similar to the naive Q\&A results. This indicates stable output in the log fault diagnosis task, with GPT-3.5 and GPT-4 showing far superior results. The Baichuan model performs poorly under both self-consistency and naive Q\&A, while other models do not change much relative to the naive Q\&A results. The zero-shot and few-shot performance of the LLMs are examined for English and Chinese test sets by comparing the results of the naive and self-consistency Q\&A experiment. The following conclusions can be drawn from the results:
\begin{itemize}
\item For most models, performance does not change much from naive Q\&A to SC. In the anomaly detection task, the performance under few-shot conditions is inferior to zero-shot. Conversely, in the fault diagnosis task, the performance under few-shot conditions exceeds zero-shot scenarios.
\item In these settings, SC prompts relatively minor improvements to the model. In repeated questions, the LLM's answers were consistent.
\item LLMs fine-tuned specifically for Chinese perform better on English and Chinese test sets than LLMs not fine-tuned for Chinese. LLaMA is a notable example, which we discuss further in the \hyperref[sec: appendix]{appendix}.
\end{itemize}


\subsubsection{SC in model robustness performance}

For the self-consistency experiment, we conducted five experiments on each model for each task using the same dataset. By analyzing these five results, we can determine if the model’s performance is stable, as shown in the \autoref{tab:SC Q&A F1-score Variance on Log Anomaly Detection}, it represents the variance of the five F1-scores obtained after performing Chinese and English naive Q\&A tasks on the model in the zero-shot and few-shot scenarios for anomaly detection. It can be observed that the variance values of most models are low, indicating that the model has good robustness in the five experiments.

\begin{table*}[!ht]
\centering
\renewcommand{\arraystretch}{1.0}
\setlength{\arrayrulewidth}{0.1pt} 
\caption{SC Q\&A F1-score Variance on Log Anomaly Detection}
\label{tab:SC Q&A F1-score Variance on Log Anomaly Detection}
\resizebox{0.7\linewidth}{!}
{
\begin{tabular}{@{}l@{\extracolsep{0.5pt}}cccc@{}} 
\toprule
\multicolumn{1}{c}{\multirow{2}{*}{\textbf{model}}} & \multicolumn{2}{c}{\textbf{zero-shot}} & \multicolumn{2}{c}{\textbf{few-shot}} \\ \cline{2-5} 
\multicolumn{1}{c}{} & \makebox[2cm][c]{\textbf{chinese}} & \makebox[2cm][c]{\textbf{english}} & \makebox[2cm][c]{\textbf{chinese}} & \makebox[2cm][c]{\textbf{english}} \\ 
\midrule
Qwen1.5-7b & 1.57E-05 & 5.5E-06 & 9.7E-06 & 8.3E-06 \\ \hline
Qwen1.5-14b & 0 & 5.7E-06 & 2.00E-07 & 5.2E-06 \\ \hline
Qwen1.5-72b & 0 & 5.3E-06 & 1.08E-05 & 1.83E-05 \\ \hline
LLaMa2-7b & 2.80265E-06 & 4.31118E-04 & 0 & 0 \\ \hline
LLaMa2-13b & 1.47E-06 & 1.53017E-06 & 0 & 0 \\ \hline
LLaMa2-70b & 3.66E-06 & 4.05727E-05 & 0 & 2.92015E-06 \\ \hline
DeVops-7b & 3.36E-07 & 1.55956E-06 & 3.70E-06 & 2.09E-07 \\ \hline
DeVops-14b & 2.68E-06 & 7.32059E-06 & 2.43E-06 & 7.40183E-06 \\ \hline
InternLM2-7b & 1.42E-05 & 1.11881E-05 & 2.15E-08 & 1.53955E-06 \\ \hline
InternLM2-20b & 1.02E-05 & 8.28797E-06 & 1.14E-05 & 6.56E-08 \\ \hline
AquilaChat-7b & 8.70E-06 & 3.15505E-05 & 1.44E-05 & 1.69002E-06 \\ \hline
GPT-3.5 & 1.74E-07 & 2.74283E-06 & 1.03543E-06 & 1.71997E-04 \\ \hline
GPT-4 & 3.66E-08 & 2.19E-07 & 1.27E-07 & 7.98E-07 \\ \hline
Gemini Pro & 2.86E-06 & 5.83E-07 & 4.21E-06 & 2.37157E-06 \\ \hline
Mistral-7b & 7.51E-05 & 4.03911E-05 & 2.81E-07 & 1.07E-07 \\ \hline
BaiChuan2-13b & 0 & 2.94E-11 & 0 & 1.15E-07 \\ \hline
GhatGLM4 & 2.43E-07 & 2.59E-07 & 1.33E-08 & 1.47E-07 \\ \hline
Claude3 Sonnet & 3.48E-10 & 1.57E-09 & 1.38E-06 & 1.33E-07 \\
\bottomrule
\end{tabular}
}
\vspace{-1em}
\end{table*}

As shown in the \autoref{tab:SC Q&A F1-score Variance on Log fault Diagnosis}, it represents the variance of the five F1-scores obtained after performing Chinese and English naive Q\&A tasks on the model in the zero-shot and few-shot scenarios for fault diagnosis. It can be seen that the model’s performance is less stable in the few-shot scenario compared to the zero-shot scenario, suggesting that the model still has some ambiguous understanding in the few-shot scenario.

\begin{table*}[!ht]
\centering
\renewcommand{\arraystretch}{1.0}
\setlength{\arrayrulewidth}{0.1pt} 
\caption{SC Q\&A F1-score Variance on Log fault Diagnosis}
\label{tab:SC Q&A F1-score Variance on Log fault Diagnosis}
\resizebox{0.7\linewidth}{!}
{
\begin{tabular}{@{}l@{\extracolsep{0.5pt}}cccc@{}} 
\toprule
\multicolumn{1}{c}{\multirow{2}{*}{\textbf{model}}} & \multicolumn{2}{c}{\textbf{zero-shot}} & \multicolumn{2}{c}{\textbf{few-shot}} \\ \cline{2-5} 
\multicolumn{1}{c}{} & \makebox[2cm][c]{\textbf{chinese}} & \makebox[2cm][c]{\textbf{english}} & \makebox[2cm][c]{\textbf{chinese}} & \makebox[2cm][c]{\textbf{english}} \\ 
\midrule
Qwen1.5-7b & 9.7E-06 & 9.5E-06 & 6.3E-06 & 4.53E-05 \\ \hline
Qwen1.5-14b & 1.53E-05 & 8.8E-06 & 9.30E-06 & 1.06E-04 \\ \hline
Qwen1.5-72b & 6.3E-06 & 6.7E-06 & 1.27E-05 & 5.7E-06 \\ \hline
LLaMa2-7b & 1.48E-05 & 5.06E-06 & 1.3125E-05 & 1.874E-05 \\ \hline
LLaMa2-13b & 1.65E-05 & 1.79E-05 & 2.1438E-05 & 1.9622E-05 \\ \hline
LLaMa2-70b & 3.41E-05 & 1.44E-05 & 4.39E-05 & 2.65E-05 \\ \hline
DeVops-7b & 5.72E-06 & 9.51E-07 & 2.02E-04 & 2.80E-05 \\ \hline
DeVops-14b & 3.43E-05 & 1.23E-05 & 3.35E-06 & 8.93E-06 \\ \hline
InternLM2-7b & 2.20E-05 & 5.15E-07 & 3.78E-04 & 8.32E-05 \\ \hline
InternLM2-20b & 2.65E-08 & 1.78E-08 & 4.53E-04 & 5.59E-06 \\ \hline
AquilaChat-7b & 3.43E-05 & 3.58E-05 & 5.13E-05 & 4.51E-05 \\ \hline
GPT-3.5 & 8.03E-05 & 3.86E-05 & 6.22E-04 & 4.11E-04 \\ \hline
GPT-4 & 5.59E-05 & 2.86E-06 & 4.88E-06 & 3.60E-07 \\ \hline
Gemini Pro & 2.76E-06 & 8.89E-05 & 3.38E-06 & 9.83E-05 \\ \hline
Mistral-7b & 1.72E-05 & 2.14E-06 & 6.56E-04 & 3.42E-04 \\ \hline
BaiChuan2-13b & 9.44E-08 & 3.64E-10 & 2.78E-10 & 0.00E+00 \\ \hline
GhatGLM4 & 1.93E-05 & 1.53E-05 & 8.90E-06 & 7.31E-06 \\ \hline
Claude3 Sonnet & 4.11E-08 & 1.55E-07 & 1.30E-06 & 9.14E-07 \\
\bottomrule
\end{tabular}
}
\vspace{-1em}
\end{table*}

\subsection{Performance on Inference Time and Average Token}

To investigate the reasoning efficiency of the LLMs and whether they are redundant in generating responses, we summarized the inference time for different models and the average number of tokens output per log. The inference time and average tokens used for each task on the English dataset in the zero-shot case of the naive Q\&A are shown below.

\subsubsection{Inference Time}

\autoref{fig:25} shows the inference time of the four classes of tasks on the English data set in the zero-shot case of the naive Q\&A.

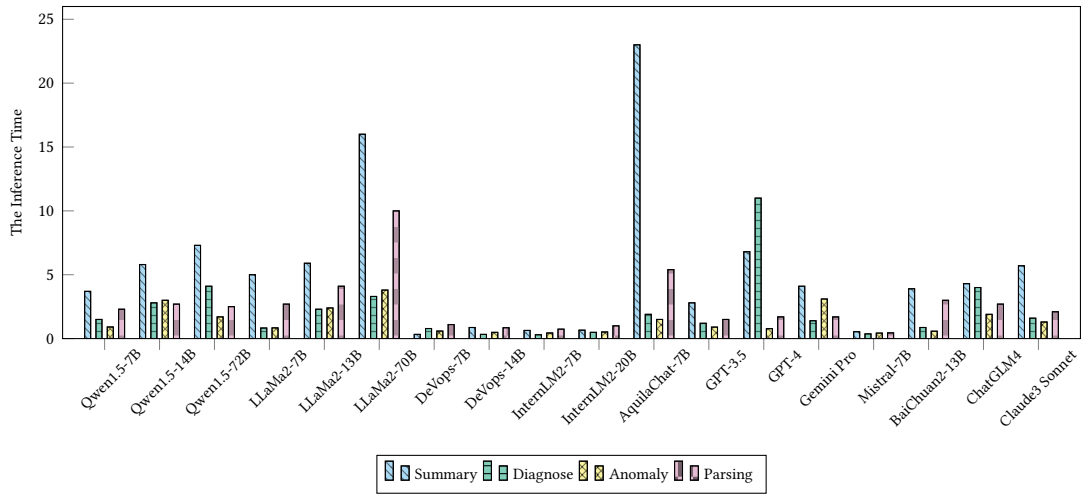
\begin{figure*}[!ht]
    \centering
    \begin{tikzpicture}
        \begin{axis}[
            ybar,
            bar width=0.08cm, 
            width=\textwidth, 
            height=6cm, 
            ymin=0, ymax=26, 
            symbolic x coords={Qwen1.5-7B, Qwen1.5-14B, Qwen1.5-72B, LLaMa2-7B, LLaMa2-13B, LLaMa2-70B, DeVops-7B, DeVops-14B,InternLM2-7B, InternLM2-20B, AquilaChat-7B, GPT-3.5, GPT-4, Gemini Pro, Mistral-7B, BaiChuan2-13B, ChatGLM4, Claude3 Sonnet},
            xtick=data,
            ylabel={The Inference Time},
            legend style={at={(0.5,-0.35)}, anchor=north, legend columns=-1, font=\scriptsize},
            xticklabel style={rotate=45, anchor=east, font=\scriptsize, xshift=12pt, yshift=-10pt},
            yticklabel style={font=\scriptsize},
            ylabel style={font=\scriptsize}, 
            xtick pos=bottom,    
            ytick pos=left,       
            enlarge x limits={0.045}, 
            ytick distance=5 
            ]
            \addplot[fill=skyblue, postaction={pattern=north west lines}, fill opacity=0.5] coordinates {(Qwen1.5-7B, 3.7) (Qwen1.5-14B, 5.8) (Qwen1.5-72B, 7.3) (LLaMa2-7B, 5) (LLaMa2-13B, 5.9) (LLaMa2-70B, 16) (DeVops-7B, 0.33) (DeVops-14B, 0.87) (InternLM2-7B, 0.64) (InternLM2-20B, 0.66) (AquilaChat-7B, 23) (GPT-3.5, 2.8) (GPT-4, 6.8) (Gemini Pro, 4.1) (Mistral-7B, 0.54) (BaiChuan2-13B, 3.9) (ChatGLM4, 4.3) (Claude3 Sonnet, 5.7)};
            \addplot[fill=bluishgreen, postaction={pattern=horizontal lines}, fill opacity=0.5] coordinates {(Qwen1.5-7B, 1.5) (Qwen1.5-14B, 2.8) (Qwen1.5-72B, 4.1) (LLaMa2-7B, 0.83) (LLaMa2-13B, 2.3) (LLaMa2-70B, 3.3) (DeVops-7B, 0.79) (DeVops-14B, 0.33) (InternLM2-7B, 0.3) (InternLM2-20B, 0.49) (AquilaChat-7B, 1.9) (GPT-3.5, 1.2) (GPT-4, 11) (Gemini Pro, 1.4) (Mistral-7B, 0.37 ) (BaiChuan2-13B, 0.87) (ChatGLM4, 4) (Claude3 Sonnet, 1.6)};
            \addplot[fill=yellow, postaction={pattern=crosshatch}, fill opacity=0.5] coordinates {(Qwen1.5-7B, 0.91) (Qwen1.5-14B, 3) (Qwen1.5-72B, 1.7) (LLaMa2-7B, 0.84) (LLaMa2-13B, 2.4) (LLaMa2-70B, 3.8) (DeVops-7B, 0.59) (DeVops-14B, 0.49) (InternLM2-7B, 0.44) (InternLM2-20B, 0.52) (AquilaChat-7B, 1.5) (GPT-3.5, 0.9) (GPT-4, 0.78) (Gemini Pro, 3.1) (Mistral-7B, 0.43) (BaiChuan2-13B, 0.58) (ChatGLM4, 1.9) (Claude3 Sonnet, 1.3)};
            \addplot[fill=reddishpurple, postaction={pattern=checkerboard}, fill opacity=0.5] coordinates {(Qwen1.5-7B, 2.3) (Qwen1.5-14B, 2.7) (Qwen1.5-72B, 2.5) (LLaMa2-7B, 2.7) (LLaMa2-13B, 4.1) (LLaMa2-70B, 10) (DeVops-7B, 1.1) (DeVops-14B, 0.85) (InternLM2-7B, 0.74) (InternLM2-20B, 1) (AquilaChat-7B, 5.4) (GPT-3.5, 1.5) (GPT-4, 1.7) (Gemini Pro, 1.7) (Mistral-7B, 0.45) (BaiChuan2-13B, 3) (ChatGLM4, 2.7) (Claude3 Sonnet, 2.1)};
            \legend{Summary, Diagnose, Anomaly, Parsing}
        \end{axis}
    \end{tikzpicture}
    \caption{The Inference Time in the Naive Q\&A situation in log analysis by zero-shot}
    \label{fig:25}
\end{figure*}

From the overall performance evaluation results, the log summary task takes the longest time among the four tasks. This is mainly because, in our test dataset, the input content for the log summary task is longer, causing the model to take more time to process these inputs. Specifically, five models: DeVops-7B, DeVops-14B, InternLM-7B, InternLM-20B, and Mistral-7B exhibit short inference times, which may be related to the setup of the test environment. Since we tested with a locally deployed model rather than calling through an API interface, this may have contributed to the time difference. A locally deployed model takes much less time to reason than if it were called through an API. In addition, the inference time of the LLaMA-2-70B model is longer, likely due to its large number of parameters.


\subsubsection{Average Token}
\begin{figure*}[htbp]
    \centering
    \begin{tikzpicture}
        \begin{axis}[
            ybar,
            bar width=0.08cm, 
            width=\textwidth, 
            height=6cm, 
            ymin=0, ymax=720, 
            symbolic x coords={Qwen1.5-7B, Qwen1.5-14B, Qwen1.5-72B, LLaMa2-7B, LLaMa2-13B, LLaMa2-70B, DeVops-7B, DeVops-14B,InternLM2-7B, InternLM2-20B, AquilaChat-7B, GPT-3.5, GPT-4, Gemini Pro, Mistral-7B, BaiChuan2-13B, ChatGLM4, Claude3 Sonnet},
            xtick=data,
            ylabel={The Average Token},
            legend style={at={(0.5,-0.35)}, anchor=north, legend columns=-1, font=\scriptsize},
            xticklabel style={rotate=45, anchor=east, font=\scriptsize, xshift=12pt, yshift=-10pt},
            yticklabel style={font=\scriptsize}, 
            ylabel style={font=\scriptsize}, 
            xtick pos=bottom,    
            ytick pos=left,       
            enlarge x limits={0.045}, 
            ytick distance=100 
            ]
            \addplot[fill=skyblue, postaction={pattern=north west lines}, fill opacity=0.5] coordinates {(Qwen1.5-7B, 322) (Qwen1.5-14B, 398) (Qwen1.5-72B, 696) (LLaMa2-7B, 479) (LLaMa2-13B, 479) (LLaMa2-70B, 426) (DeVops-7B, 545) (DeVops-14B, 537) (InternLM2-7B, 339) (InternLM2-20B, 299) (AquilaChat-7B, 655) (GPT-3.5, 263) (GPT-4, 263) (Gemini Pro, 231) (Mistral-7B, 335) (BaiChuan2-13B, 134) (ChatGLM4, 231) (Claude3 Sonnet, 277)};
            \addplot[fill=bluishgreen, postaction={pattern=horizontal lines}, fill opacity=0.5] coordinates {(Qwen1.5-7B, 172) (Qwen1.5-14B, 221) (Qwen1.5-72B, 282) (LLaMa2-7B, 74) (LLaMa2-13B, 212) (LLaMa2-70B, 105) (DeVops-7B, 205) (DeVops-14B, 52.5) (InternLM2-7B, 109) (InternLM2-20B, 130) (AquilaChat-7B, 56) (GPT-3.5, 65.1) (GPT-4, 27.2) (Gemini Pro, 11) (Mistral-7B, 24.1) (BaiChuan2-13B, 22.1) (ChatGLM4, 312) (Claude3 Sonnet, 32.1)};
            \addplot[fill=yellow, postaction={pattern=crosshatch}, fill opacity=0.5] coordinates {(Qwen1.5-7B, 66.9) (Qwen1.5-14B, 241) (Qwen1.5-72B, 97.3) (LLaMa2-7B, 74.3) (LLaMa2-13B, 227) (LLaMa2-70B, 140) (DeVops-7B, 132) (DeVops-14B, 61.8) (InternLM2-7B, 21) (InternLM2-20B, 55.9) (AquilaChat-7B, 50.5) (GPT-3.5, 18.2) (GPT-4, 4.94) (Gemini Pro, 2.59) (Mistral-7B, 2.21) (BaiChuan2-13B, 7.24) (ChatGLM4, 106) (Claude3 Sonnet, 2.15)};
            \addplot[fill=reddishpurple, postaction={pattern=checkerboard}, fill opacity=0.5] coordinates {(Qwen1.5-7B, 276) (Qwen1.5-14B, 194) (Qwen1.5-72B, 142) (LLaMa2-7B, 429) (LLaMa2-13B, 306) (LLaMa2-70B, 384) (DeVops-7B, 232) (DeVops-14B, 139) (InternLM2-7B, 263) (InternLM2-20B, 214) (AquilaChat-7B, 232) (GPT-3.5, 98.8) (GPT-4, 59.8) (Gemini Pro, 66.4) (Mistral-7B, 117) (BaiChuan2-13B, 162) (ChatGLM4, 175) (Claude3 Sonnet, 63.3)};
            \legend{Summary, Diagnose, Anomaly, Parsing}
        \end{axis}
    \end{tikzpicture}
    \caption{The Average Token in the Naive Q\&A situation in log analysis by zero-shot}
    \label{fig:26}
\end{figure*}
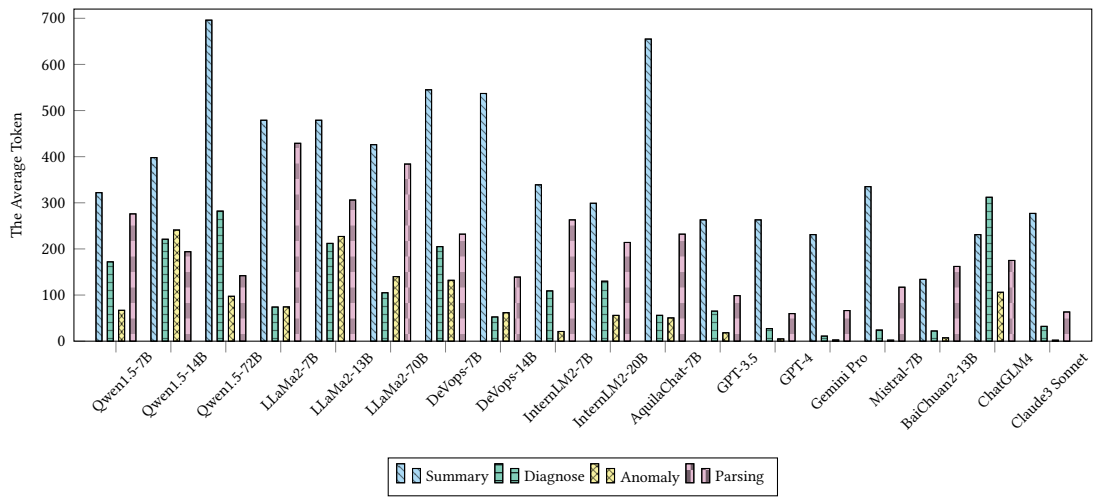

\autoref{fig:26} shows the Average Token of the four classes of tasks on the English data set with zero-shot setting for naive Q\&A.

From the overall performance evaluation results, the log summary task outputs the highest average number of tokens among the four tasks. This phenomenon is mainly determined by the nature of the task because the log summary task requires the model to generate a concise summary, which usually requires more tokens to accurately represent the main content of the log. However, our evaluation results show that Gemini, GPT, and Mistral models output a lower average number of tokens, indicating that their answers are more concise, without excessive redundant information, and their outputs are cleaner. Conversely, LLaMA and Qwen models output more tokens on average, meaning their answers contain more extraneous content. In practice, this can result in users spending more time and effort sifting useful information from responses, which reduces efficiency.

From the analysis, we can draw the following scientifically rigorous conclusions:
\begin{itemize}
    \item The log summary task takes the longest inference time among the four tasks, mainly due to the longer input content.
    \item Locally deployed models such as DeVops-7B, DeVops-14B, InternLM-7B, InternLM-20B, and Mistral-7B exhibit shorter inference times compared to API-based models.
    \item The LLaMA2-70B model has a longer inference time due to its large number of parameters.
    \item The log summary task outputs the highest average number of tokens, while Gemini, GPT, and Mistral models produce more concise outputs.
    \item LLaMA and Qwen models output more tokens on average, containing more extraneous content, which can reduce user efficiency in practical applications.
\end{itemize}

\subsection{Performance on Different  parameters}

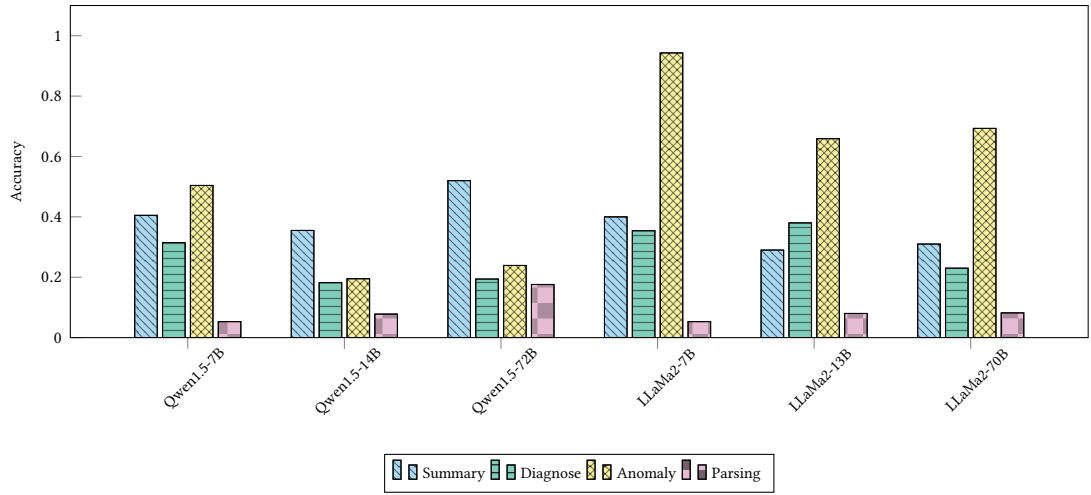
\begin{figure*}[ht]
    \centering
    \begin{tikzpicture}
        \begin{axis}[
            ybar,
            bar width=0.3cm, 
            width=\textwidth, 
            height=6cm, 
            ymin=0, ymax=1.1, 
            symbolic x coords={Qwen1.5-7B, Qwen1.5-14B, Qwen1.5-72B, LLaMa2-7B, LLaMa2-13B, LLaMa2-70B},
            xtick=data,
            ylabel={Accuracy},
            legend style={at={(0.5,-0.35)}, anchor=north, legend columns=-1, font=\scriptsize},
            xticklabel style={rotate=45, anchor=east, font=\scriptsize, xshift=12pt, yshift=-10pt},
            yticklabel style={font=\scriptsize},
            ylabel style={font=\scriptsize}, 
            xtick pos=bottom,    
            ytick pos=left,       
            enlarge x limits={0.15}, 
            ytick distance=0.2 
            ]
            \addplot[fill=skyblue, postaction={pattern=north west lines}, fill opacity=0.5] coordinates {(Qwen1.5-7B, 0.405) (Qwen1.5-14B, 0.355) (Qwen1.5-72B, 0.520) (LLaMa2-7B, 0.400) (LLaMa2-13B, 0.290) (LLaMa2-70B, 0.310)};
            \addplot[fill=bluishgreen, postaction={pattern=horizontal lines}, fill opacity=0.5] coordinates {(Qwen1.5-7B, 0.315) (Qwen1.5-14B, 0.182) (Qwen1.5-72B, 0.194) (LLaMa2-7B, 0.354) (LLaMa2-13B, 0.380) (LLaMa2-70B, 0.230)};
            \addplot[fill=yellow, postaction={pattern=crosshatch}, fill opacity=0.5] coordinates {(Qwen1.5-7B, 0.504) (Qwen1.5-14B, 0.195) (Qwen1.5-72B, 0.239) (LLaMa2-7B, 0.943) (LLaMa2-13B, 0.659) (LLaMa2-70B, 0.693)};
            \addplot[fill=reddishpurple, postaction={pattern=checkerboard}, fill opacity=0.5] coordinates {(Qwen1.5-7B, 0.053) (Qwen1.5-14B, 0.078) (Qwen1.5-72B, 0.176) (LLaMa2-7B, 0.053) (LLaMa2-13B, 0.080) (LLaMa2-70B, 0.082)};
            \legend{Summary, Diagnose, Anomaly, Parsing}
        \end{axis}
    \end{tikzpicture}
    \caption{The Accuracy of LLaMa-2 and Qwen-1.5-Chat in zero-shot English Naive Q\&A}
    \label{fig:27}
\end{figure*}

\autoref{fig:27} shows the accuracy of LLaMA-2 and Qwen-1.5-Chat for different parameter counts. We used a zero-shot naive Q\&A assessment on an English dataset.

From the comparison of results, both models achieve better performance under the parameter number of 7B. This finding suggests that model size is not a determining factor for log analysis tasks. While an increase in the number of parameters generally means that the model can capture more features and patterns, a large number of parameters can also cause the model to be too complex to process log data quickly and accurately in real-world applications. Therefore, we can conclude that for log analysis tasks, choosing the right number of parameters is crucial, not simply "bigger is better." Future research should focus on how to optimize the size of the model for a more efficient and cost-effective log analysis solution without sacrificing performance.

This chapter provides a comprehensive performance evaluation of several LLMs, including GPT-4, GPT-3.5, Gemini-Pro, Claude-3-Sonnet, DevOps-Model-7B-Chat, DevOps-Model-14B-Chat, Mistral-7B, InternLM-7B, InternLM-20B, Baichuan2-13B-Chat, ChatGLM4, AquilaChat-7B, LLaMA-2-7B-Chat, LLaMA-2-13B-Chat, LLaMA-2-70B-Chat, Qwen-1.5-7B-Chat, Qwen-1.5-14B-Chat, Qwen-1.5-72B-Chat, and more. These models represent the latest advances in natural language processing, and their performance evaluation is critical to understanding the potential of LLMs for log analysis tasks.

Through comparative analysis of these models, we find significant differences in their performance on log analysis tasks. These differences may be due to differences in model design philosophy, training strategies, and model architecture. For example, some models may perform better with long series of log data, while others may show greater efficiency in generating summaries or detecting anomalies. Additionally, the number of parameters and training objectives of the model are also important factors affecting its performance in the log analysis task. Our evaluation highlights the need to consider these factors when selecting and customizing a log analysis model to ensure that the model effectively meets the needs of real-world applications.

During the evaluation process, we also focused on the two key metrics of the model’s inference time and average number of output tokens. Inference time reflects the time it takes the model to process a single log entry, while the average number of output tokens reveals the computational resources required for the model to generate a response. Our data show that even models with similar numbers of parameters can perform very differently on these two measures. Some models demonstrate shorter inference times and fewer average output tokens, suggesting they are more efficient at handling log analysis tasks. Other models may perform poorly in these two areas, which may affect their overall performance.

To sum up, the evaluation work in this chapter not only reveals the performance differences of different LLMs in log analysis tasks but also provides a valuable reference for future research, which is helpful to promote the technical progress and application development of LLMs in the log analysis field. With a deeper understanding of how models perform on different metrics, we can better guide model selection and optimization to achieve a more efficient and cost-effective log analysis solution.

\subsection{Baselines Results}

\autoref{tab:baseline results} presents the baseline models’ accuracy and F1-scores on our dataset. 

\begin{table}[!ht]
\centering
\renewcommand{\arraystretch}{1.0}
\setlength{\arrayrulewidth}{0.1pt} 
\caption{Baseline results}
\label{tab:baseline results}
\resizebox{0.8\linewidth}{!}
{
\begin{tabular}{@{}l@{\extracolsep{0.5pt}}cccccc@{}}
\toprule
\textbf{Log Task} & \textbf{Method} & \textbf{F1-score} & \textbf{Accuracy}  & \textbf{Precision} & \textbf{Recall}  \\
\midrule
\multirow{2}{*}{\begin{tabular}[c]{@{}c@{}}Log Anomaly Detection\end{tabular}} & NeuralLog & 0    & 0.97 & 0 & 0 \\
\cline{2-6}
& LogRobust & 0.09 & 0.95 & 0.33 & 0.55 \\
\hline
\multirow{2}{*}{Log Parsing}  & Drain & 0.048 & 0.773 & 0.039 & 0.065 \\
\cline{2-6}
& LogPPT & 0.068 & 0.289 & 0.055 & 0.088 \\
\hline
\multirow{2}{*}{\begin{tabular}[c]{@{}c@{}}Log fault Diagnosis\end{tabular}} & LogKG & 0.5805    & 0.6421 & 0.5787 & 0.65714 \\
\cline{2-6}
& LogCluster & 0.227 & 0.233 & 0.435 & 0.233 \\
\hline
LogSummary & LogSummary & 0.722 & 0.722 & 0.565 & 1 \\ 
\bottomrule
\end{tabular}
}
\vspace{-1em}
\end{table}

For the log anomaly detection task, while NeuralLog achieves an accuracy of 0.97, its inability to identify any anomalies results in an F1-score of 0. LogRobust, however, improves upon this by attaining an F1-score of 0.09, along with an accuracy of 0.95, precision of 0.33, and a recall rate of 0.55.

For the log parsing task, Drain and LogPPT display low F1-scores at 0.048 and 0.068, respectively, even though they achieve accuracies of 0.773 and 0.289, suggesting their limited parsing capabilities. Despite this, LogPPT marginally outperforms Drain in this particular context.

For the log fault diagnosis task, LogKG demonstrates superior diagnostic effectiveness with an F1-score of 0.5805 and an accuracy of 0.6421, showcasing balanced precision (0.5787) and recall (0.65714). On the other hand, LogCluster consistently records a significantly lower F1-score of 0.227, despite maintaining a relatively high precision rate of 0.435. The notably lower recall rate of 0.233 emphasizes its restricted capability in detecting faults.

For the log summary task, the LogSummary algorithm currently achieves good overall performance with an F1-score of 0.722 and a perfect recall rate of 1.0, meaning it fully encompasses essential information with a precision rate of 0.565. This also reflects that there remains room for improvement in refining the summaries while maintaining comprehensiveness.

When comparing the baseline results with LLMs, several observations can be made:

\begin{itemize}
    \item For log anomaly detection, LLMs generally achieves higher F1-scores compared to the baseline models. For instance, models like GPT-4 and Gemini Pro show superior performance with higher F1-scores.
    \item In log parsing tasks, the performance of LLMs also surpasses that of the baselines. Models such as GPT-4 and Claude3 Sonnet demonstrate better parsing capabilities with higher accuracy.
    \item For log fault diagnosis, LLMs like GPT-3.5 and GPT-4 significantly outperform the baselines in few-shot scenario. These models achieve much higher F1-scores and accuracy, indicating better diagnostic effectiveness.
    \item In the log summary task, LLMs continue to show strong performance. Models like DeVops-7b and DeVops-14b provide concise and accurate summaries with high accuracy, indicating that they can effectively generate comprehensive summaries.
\end{itemize}

From the comparison of baseline results and our models, we can draw the following scientifically rigorous conclusions:
\begin{itemize}
    \item LLMs generally achieve higher F1-scores, accuracy across all tasks compared to the baseline models in few-shot scenario, indicating superior performance.
    \item The significant improvements in performance metrics highlight the effectiveness of LLMs in handling various log analysis tasks, including anomaly detection, parsing, fault diagnosis, and summary generation.
    \item The results suggest that advanced LLMs like GPT-4 and Gemini Pro are more capable of processing log data efficiently and accurately, making them better suited for real-world log analysis applications.
    \item Further research should focus on optimizing these models to enhance their performance even further, particularly in areas where the baseline models show limitations.
\end{itemize}

\section{Conclusion}
\label{sec: discussion}

LogEval represents a significant advancement in the benchmarking of Large Language Models (LLMs) for log analysis tasks. This comprehensive benchmark suite evaluates a range of log analysis tasks, including log parsing, log anomaly detection, log fault diagnosis, and log summary extraction. By thoroughly assessing the capabilities and limitations of current LLMs in these domains, LogEval provides valuable insights into their potential applications and areas requiring further development.

Our findings highlight the transformative potential of LLMs in log analysis practices. These models demonstrate significant promise in enhancing the efficiency and accuracy of log analysis, crucial for maintaining the stability and performance of complex information systems. However, the evaluation also reveals specific areas where current models fall short, emphasizing the need for continued research and improvement.

The benchmark suite underscores the critical importance of model selection, showing how different models can excel or struggle with specific log analysis tasks. For instance, models like GPT-4 consistently outperform others in tasks requiring deep comprehension and nuanced understanding, such as log parsing and log fault diagnosis. Conversely, smaller parameter models often lag in performance, particularly in more complex tasks. This differentiation is crucial for researchers and practitioners when choosing the most appropriate model for their specific needs.

Technical features such as model size, training data quality, and fine-tuning processes significantly impact performance. Larger models with extensive fine-tuning on high-quality data sets tend to perform better, yet they also require more computational resources. LogEval's comprehensive evaluation framework provides a clear comparison of these factors, aiding in the development of more efficient and effective LLMs for log analysis.

As the field of log analysis evolves, benchmarks like LogEval will play a crucial role in driving technological progress and application development. LogEval offers a standardized framework for evaluating LLMs, facilitating meaningful comparisons across different models and encouraging innovation and improvement in log analysis technologies. The insights gained from this benchmark are expected to inspire further research and development, leading to the creation of LLMs that are even more adept at handling the complexities of log analysis.

The implications of LogEval extend beyond mere evaluation. It serves as a guide for future research directions, highlighting the strengths and weaknesses of current LLMs. By identifying specific areas for improvement, LogEval provides a roadmap for the next generation of LLMs in log analysis, aiming for models that not only perform well across various tasks but also do so efficiently and reliably in real-world applications.

In summary, LogEval has established a robust foundation for assessing the performance of LLMs in log analysis tasks. It provides a valuable reference for researchers and practitioners, contributing to the advancement of LLM technology and its application in maintaining the health and performance of modern information systems. As we continue to refine these models, the insights from LogEval will be instrumental in shaping the future of log analysis.

\section{Discussion }
\label{sec: Conclusion}
\subsection{Limitations and Directions for Improvement}

LogEval, as the first dedicated benchmark suite for log analysis tasks, represents a significant step forward in the field, but it still has some limitations and room for improvement. Firstly, the diversity of datasets is a limitation of LogEval. Currently, LogEval’s datasets primarily focus on several common log formats and tasks, and their applicability to more extensive and diversified log data remains to be verified. Future work could expand the datasets to include more types of logs and more complex tasks to enhance the universality and robustness of the benchmark tests. Secondly, the complexity of tasks is also a limitation. LogEval currently focuses on basic tasks such as log parsing, log anomaly detection, log fault diagnosis, and log summary. However, the actual application scenarios of log analysis are often more complex, involving multi-task learning and end-to-end system integration. Future work could consider adding more complex tasks such as log clustering and event sequence analysis to better evaluate the comprehensive capabilities of models.

\subsection{Issues and Directions for Improving Model Output Quality and Inference Time}

During the evaluation process of LogEval, several noteworthy issues were identified. Firstly, some models produce results that are overly lengthy and complex when handling log analysis tasks. This may be due to the models' fault to accurately understand the prompt’s meaning or effectively distill key information. Future research should focus on enhancing the models’ comprehension abilities and output quality to generate more concise and accurate log analysis results. Secondly, some models exhibit excessively long inference times, which may be attributed to excessive model parameters or suboptimal model architectures. In real-world log analysis applications, the model’s response speed is crucial as it directly affects the operational efficiency of personnel. Therefore, model optimization should aim to reduce inference time while maintaining performance to meet real-time requirements. Addressing these issues provides valuable guidance for researchers to improve models and contribute to the technological advancement of the log analysis field.

\subsection{Future Development Directions of LLMs in Log Analysis Tasks}

With the continuous advancement of LLM technology, the future of log analysis will increasingly rely on these advanced technologies. On one hand, model optimization will continue to drive performance improvements in log analysis. For instance, by improving model architecture, training strategies, and optimization algorithms, the accuracy and efficiency of models can be further enhanced. On the other hand, task expansion is also an important future development direction. In addition to traditional log analysis tasks, future research could explore combining log analysis with other technologies such as knowledge graphs and causal inference to achieve more comprehensive and in-depth log analysis. Furthermore, with the development of federated learning and privacy protection technologies, exploring efficient log analysis while ensuring data privacy and security is also a worthwhile direction to pursue.

\subsection{Guidance from LogEval for Practical Application Scenarios}

LogEval provides important references and guidance for log analysis in practical application scenarios. Firstly, LogEval offers a comprehensive evaluation framework that can assist enterprises and research institutions in selecting models that best suit their needs. By comparing the performance of different models, stakeholders can better understand their strengths and weaknesses, thus choosing the model most suitable for their application scenario. Secondly, LogEval’s evaluation metrics and methods can serve as benchmarks for performance evaluation in practical applications. By comparing their system’s evaluation results with those of LogEval, organizations can more accurately assess the performance of their log analysis systems and implement targeted optimizations and improvements. Finally, LogEval sets a reference standard for research and application in the log analysis field, contributing to the advancement of the AIOps field.

\section{Appendix}
\label{sec: appendix}
\appendix
\subsection{Metrics used in Question-Answering Evaluation}

\textbf{ROUGE} (Recall-Oriented Understudy for Gisting Evaluation): A set of metrics designed to evaluate machine translations and summaries by measuring the overlap between the predicted and reference n-grams. ROUGE-N evaluates the overlap of n-grams between the prediction and reference, while ROUGE-L assesses sentence structure similarity by identifying the longest common sequence of n-grams. ROUGE scores range from 0 to 100, with higher scores indicating better performance.

\textbf{BLEU} (Bilingual Evaluation Understudy): Focuses on the precision of the generated answers by comparing them to a set of reference translations. It is widely used in natural language processing, particularly for translation tasks. BLEU scores are normalized from 0 to 100, with higher scores reflecting better accuracy.

\textbf{Cosine Similarity}: A method for measuring the similarity between two vectors, often used to assess the semantic similarity between texts. In the context of LogEval, cosine similarity evaluates the likeness between model-generated answers and the ground truth, with values ranging from -1 (completely dissimilar) to 1 (identical).

\textbf{F1-Score}: A metric that considers both precision and recall, providing a balanced measure of a model’s performance in log analysis tasks. It ranges from 0 to 1, with higher values indicating better performance.

\textbf{Accuracy}: Measures the proportion of correct answers generated by the model compared to the ground truth, providing a straightforward assessment of model performance.

\subsection{ADDITIONAL DETAILS OF EXPERIMENTS}
\subsubsection{Detailed Information of LLMs Evaluated}

Various LLMs were evaluated in this study, including GPT-4, GPT-3.5-turbo, Claude-3-Sonnet, Gemini-Pro, Mistral, InternLM2-Chat, DevOps-Model-Chat, AquilaChat, ChatGLM4, ChatGLM3, LLaMA-2, Qwen-1.5-Chat, Baichuan2-Chat, and others. These models represent the latest advances in natural language processing, and their performance evaluation is critical for understanding the potential of LLMs in log analysis tasks. GPT-4 is a large multimodal model capable of accepting both image and text inputs and generating text outputs. It demonstrates human-level performance across various professional and academic benchmarks and is currently recognized as the most powerful language model. GPT-3.5-turbo, an early AI language model developed by OpenAI based on GPT-3.5, is the version we used. Claude-3-Sonnet, a large language model developed by Anthropic, is renowned for its powerful multimodal understanding and generation capabilities. Gemini-Pro, another large language model developed by Anthropic, gains attention for its excellent natural language understanding and generation abilities. Mistral, also developed by Anthropic, is noted for its strong multimodal understanding and generation capabilities. InternLM2-Chat, developed by the Shanghai AI Laboratory, is a multilingual model trained through multi-stage progressive training on trillions of tokens, capable of handling multiple languages including Chinese. Devops-Model-Chat, an open-source model targeted at Chinese DevOps, primarily aims to provide practical value in the DevOps domain. AquilaChat, a chat model based on the open-source LLM Aquila released by Aquila Intelligence, can handle multiple languages including Chinese. ChatGLM4 and ChatGLM3, developed by the Knowledge Engineering Group at Tsinghua University, are based on the latest version and the third version of the GLM model. They are fine-tuned for various natural language understanding and generation tasks using autoregressive blank-filling objectives. LLaMA-2, Meta's second-generation open-source LLM, can handle multiple languages including Chinese. Qwen-1.5-Chat, developed by Alibaba Cloud, gained attention for its powerful Chinese understanding and generation capabilities. Baichuan2-Chat, based on the alignment chat model of the open-source LLM Baichuan2-Base released by Baichuan Intelligence, is noted for its strong multilingual processing capabilities.

\subsubsection{An Example of few-shot}

\autoref{fig:28} shows two input examples under the "few-shot" setting, illustrating scenarios for log parsing and log fault diagnosis tasks. Each model was given 5 logs and their corresponding labels in the "instruction" field, allowing the model to learn and better respond.



\begin{figure}[!ht]
    \begin{center}
    \begin{tikzpicture}
        \definecolor{labelcolor}{RGB}{240,240,240}
        \definecolor{predictcolor}{RGB}{255,228,216}
    
        \newcommand{\boxwidth}{0.47\textwidth} 
        \newcommand{\boxheight}{8.5cm}
        \newcommand{\boxheighttop}{0.5cm}
        \newcommand{\boxheightmiddle}{6.5cm}
        \newcommand{\boxheightbottom}{1cm}
        \newcommand{\boxgap}{0.005\textwidth} 
    
        \matrix[column sep=\boxgap, row sep=0mm] {
            \node[draw, text width=\boxwidth, minimum height=\boxheight] (box1) {};
            \node[fill=labelcolor, draw, text width=\boxwidth, minimum height=\boxheighttop, anchor=north west] at (box1.north west) {\parbox{\boxwidth} \raggedright \small \textbf{"id":} "0"};
            \node[draw, text width=\boxwidth, minimum height=\boxheightmiddle, anchor=north west] at ([yshift=-\boxheighttop]box1.north west) {\parbox{\boxwidth} \raggedright \small \textbf{"instruction":} "In our data scenario, there are three types of faults \{Processor CPU Caterr, Memory Throttled Uncorrectable Error Correcting Code, Hard Disk Drive Control Error Computer System Bus Short Circuit Programmable Gate Array Device Unknown\}. Analyze the log entry and identify the type of fault that occurred. Only output the fault type.\textbackslash nFor Example:\textbackslash nlog entry:\textbackslash nTemperature CPU0\_Margin\_Temp | Lower Critical going low | Asserted | Reading -16 \&1t; Threshold 0 degrees C\textbackslash nanswer:'Processor CPU Caterr';\\nlog entry:\textbackslash nMemoryCPU1E0\_DIMM\_Stat | Correctable ECC | Asserted\textbackslash nanswer:'Memory Throttled Uncorrectable Error Correcting Code';\textbackslash nlog entry:\textbackslash nSystem Boot Initiated BlOS\_Boot\_Up | Initiated by power up | Asserted\textbackslash nanswer:'Hard Disk Drive Control Error Computer System Bus Short Circuit Programmable Gate Array Device Unknown'"};
            \node[fill=predictcolor, draw, text width=\boxwidth, minimum height=\boxheightbottom, anchor=north west] at ([yshift=-\boxheighttop-\boxheightmiddle]box1.north west) {\parbox{\boxwidth} \raggedright \small \textbf{"input":} "\textbackslash nlog entry:\textbackslash nProcessor \#Oxfa | Configuration Error | Asserted"};
            \node[draw, text width=\boxwidth, minimum height=\boxheighttop, anchor=north west] at ([yshift=-\boxheighttop-\boxheightmiddle-\boxheightbottom]box1.north west) {\parbox{\boxwidth} \raggedright \small \textbf{"output":} "Processor CPU Caterr"}; &
    
            \node[draw, text width=\boxwidth, minimum height=\boxheight] (box2) {};
            \node[fill=labelcolor, draw, text width=\boxwidth, minimum height=\boxheighttop, anchor=north west] at (box2.north west) {\parbox{\boxwidth} \raggedright \small \textbf{"id":} "0"};
            \node[draw, text width=\boxwidth, minimum height=\boxheightmiddle, anchor=north west] at ([yshift=-\boxheighttop]box2.north west) {\parbox{\boxwidth} \raggedright \small \textbf{"instruction":} "Parse the following log entry into a template format, replacing variable parts with a wildcard <*>,and focus the answer after the keyword 'Answer'\textbackslash nFor example:\textbackslash nlog entry:\textbackslash nno floppy controllers found,\textbackslash nanswer:'no floppy controllers found';\textbackslash nlog entry:\textbackslash n13 tree receiver 1 in re-synch state event(s) (dcr 0x0185) detected over 4562 seconds,\textbackslash nanswer:'<*> tree receiver <*> in re-synch state event(s) (dcr <*>) detected over <*> seconds';\textbackslash niog entry:\textbackslash n... autorun DONE.,\textbackslash nanswer:'... autorun DONE.'; \textbackslash nlog entry:\textbackslash n2 L3 EDRAM error(s) (dcr 0x0157) detected and corrected over 282 seconds, \textbackslash nanswer:'<*> L3 EDRAM error(s) (dcr <*>) detected and corrected over <*> seconds'; \textbackslash nlog entry:\textbackslash nprobe of vesafb0 failed with error -6,\textbackslash nanswer:'probe of vesafb0 failed with error <*>'."};
            \node[fill=predictcolor, draw, text width=\boxwidth, minimum height=\boxheightbottom, anchor=north west] at ([yshift=-\boxheighttop-\boxheightmiddle]box2.north west) {\parbox{\boxwidth} \raggedright \small \textbf{"input":} "\textbackslash nlog entry:\textbackslash ninstruction cache parity error corrected"};
            \node[draw, text width=\boxwidth, minimum height=\boxheighttop, anchor=north west] at ([yshift=-\boxheighttop-\boxheightmiddle-\boxheightbottom]box2.north west) {\parbox{\boxwidth} \raggedright \small \textbf{"output":} "instruction cache parity error corrected"}; \\
        };
    \end{tikzpicture}
    \end{center}
    \caption{Two input examples under the "few-shot" setting}
    \label{fig:28}
\end{figure}

\subsubsection{Performance on Different Languages}

\begin{figure*}[htbp]
    \centering
    \begin{tikzpicture}
        \begin{axis}[
            ybar,
            bar width=0.2cm, 
            width=\textwidth, 
            height=6cm, 
            ymin=-0.01, ymax=0.55, 
            symbolic x coords={Qwen1.5-7B, Qwen1.5-14B, Qwen1.5-72B, LLaMa2-7B, LLaMa2-13B, LLaMa2-70B, DeVops-7B, DeVops-14B,InternLM2-7B, InternLM2-20B, AquilaChat-7B, GPT-3.5, GPT-4, Gemini Pro, Mistral-7B, BaiChuan2-13B, ChatGLM4, Claude3 Sonnet},
            xtick=data,
            ylabel={Accuracy},
            legend style={at={(0.5,-0.35)}, anchor=north, legend columns=-1, font=\scriptsize},
            xticklabel style={rotate=45, anchor=east, font=\scriptsize, xshift=12pt, yshift=-10pt},
            yticklabel style={font=\scriptsize},
            ylabel style={font=\scriptsize}, 
            xtick pos=bottom,    
            ytick pos=left,       
            enlarge x limits={0.045}, 
            ytick distance=0.1 
            ]
            \addplot[fill=skyblue, postaction={pattern=north west lines}, fill opacity=0.5] coordinates {(Qwen1.5-7B, 0.351) (Qwen1.5-14B, 0.366) (Qwen1.5-72B, 0.306) (LLaMa2-7B, 0.086) (LLaMa2-13B, 0.057) (LLaMa2-70B, 0.090) (DeVops-7B, 0.324) (DeVops-14B, 0.362) (InternLM2-7B, 0.492) (InternLM2-20B, 0.442) (AquilaChat-7B, 0.312) (GPT-3.5, 0.413) (GPT-4, 0.247) (Gemini Pro, 0.367) (Mistral-7B, 0.380) (BaiChuan2-13B, 0.045) (ChatGLM4, 0.350) (Claude3 Sonnet, 0.288)};
            \addplot[fill=reddishpurple, postaction={pattern=checkerboard}, fill opacity=0.5] coordinates {(Qwen1.5-7B, 0.315) (Qwen1.5-14B, 0.182) (Qwen1.5-72B, 0.194) (LLaMa2-7B, 0.354) (LLaMa2-13B, 0.380) (LLaMa2-70B, 0.230) (DeVops-7B, 0.281) (DeVops-14B, 0.287) (InternLM2-7B, 0.247) (InternLM2-20B, 0.342) (AquilaChat-7B, 0.313) (GPT-3.5, 0.278) (GPT-4, 0.423) (Gemini Pro, 0.320) (Mistral-7B, 0.248) (BaiChuan2-13B, 0) (ChatGLM4, 0.404) (Claude3 Sonnet, 0.442)};
            \legend{Chinese, English}
        \end{axis}
    \end{tikzpicture}
    \caption{The performance of LLMs on fault diagnosis tasks under the "zero-shot" naive Q\&A in both Chinese and English test sets}
    \label{fig:29}
\end{figure*}
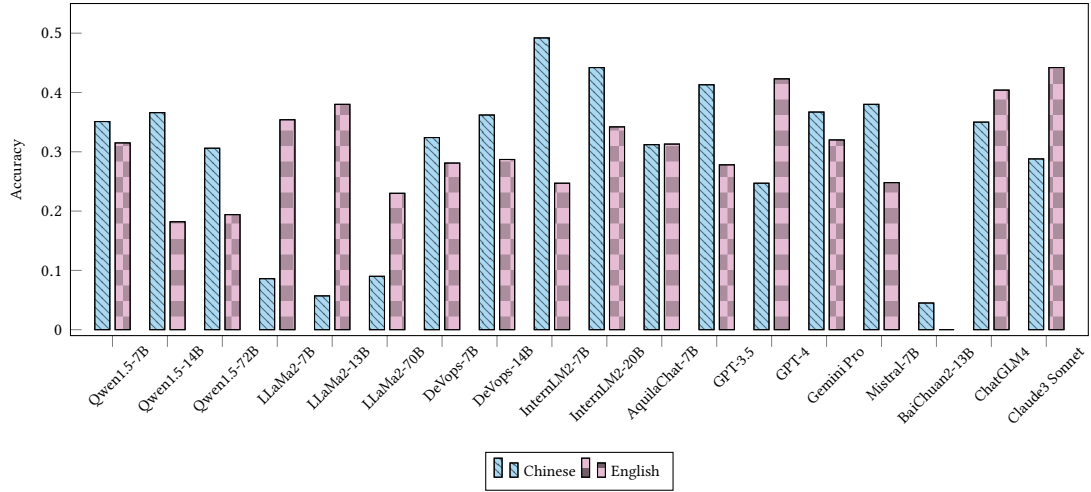

\autoref{fig:29}, the zero-shot performance of various LLMs on English and Chinese questions under the naive setting in fault diagnosis is compared. It is noted that some LLMs specifically trained or fine-tuned on Chinese corpora, such as Qwen-1.5-7B-Chat, perform better on Chinese questions. Despite being trained on Chinese corpora, ChatGLM4 still performs better on English questions compared to Chinese questions. The LLaMA-2 series models exhibit the most significant performance drop when switching to Chinese questions, indicating the language transition's impact on LLMs' understanding and domain-specific knowledge extraction abilities.


\subsubsection{Case Study}

 In zero-shot evaluations, models may encounter questions difficult to answer directly, especially in specific domain tasks such as log analysis. If the model cannot generate the correct answer directly, it may output vague or uncertain content. as shown in \autoref{fig:30}, This phenomenon may be due to the model not fully understanding the question. Even well-performing models across a wide range of tasks may struggle with specific domain questions not directly related to their training data. Integrating more domain knowledge into the model during training may help alleviate this issue, enabling better understanding and handling of log analysis tasks.

In few-shot evaluations, adopting more advanced settings may lead to worse results. We found that when evaluating with a small number of prompts, some models may misunderstand the nature of the task, incorrectly treating it as a generative problem rather than an answering problem. as shown in \autoref{fig:31}, This misunderstanding may cause the model to deviate from the correct track when generating answers, thereby affecting its performance. Specifically, a few prompts may not be sufficient to indicate that the model should generate concise and accurate answers, but may instead prompt the model to respond to the content, leading to longer narratives or generating more complex content. Additionally, the model may overly rely on patterns learned during pre-training rather than adjusting its responses based on specific information in the prompts. In such cases, the model's generative ability may be misled, resulting in outputs that are less accurate or irrelevant to the questions. To mitigate this issue, we believe that providing clearer prompts during evaluation or increasing the number of prompts to more accurately guide the model's understanding of the task requirements is essential. Moreover, developing and optimizing evaluation metrics and methods tailored to specific tasks and evaluation settings are also crucial for improving model performance. Through these approaches, we can help models better adapt to different evaluation environments, thereby improving their performance in practical applications.

In our research evaluation process, we observe some models exhibiting significant output redundancy and continuous repetition when generating answers, as shown in \autoref{fig:32}. The reasons for this phenomenon may be related to the model's inherent generation preferences, self-reinforcement effects, and probability distributions set during initialization. This repetitive answering not only results in a large amount of invalid information in the generated content but also severely affects the efficiency and quality of the model's answers. This contradicts the goal of pursuing efficiency in the field of log anomaly detection.


\begin{figure}[!ht]
    \begin{center}
    \begin{tikzpicture}
        \definecolor{labelcolor}{RGB}{240,240,240}
        \definecolor{predictcolor}{RGB}{255,228,216}
    
        \newcommand{\boxwidth}{0.6\textwidth} 
        \newcommand{\boxheight}{5cm}
        \newcommand{\boxheighttop}{0.5cm}
        \newcommand{\boxheightmiddle}{0.5cm}
        \newcommand{\boxheightbottom}{4cm}
        \newcommand{\boxgap}{0.005\textwidth} 
    
        \matrix[column sep=\boxgap, row sep=0mm] {
            \node[draw, text width=0.2\textwidth, minimum height=\boxheight] (box1) {};
            \node[fill=labelcolor, draw, text width=0.2\textwidth, minimum height=\boxheighttop, anchor=north west] at (box1.north west) {\parbox{0.2\textwidth} \raggedright \small \textbf{"id":} "8"};
            \node[draw, text width=0.2\textwidth, minimum height=\boxheightmiddle, anchor=north west] at ([yshift=-\boxheighttop]box1.north west) {\parbox{0.2\textwidth} \raggedright \small \textbf{"label":} "abnormal"};
            \node[fill=predictcolor, draw, text width=0.2\textwidth, minimum height=\boxheightbottom, anchor=north west] at ([yshift=-\boxheighttop-\boxheightmiddle]box1.north west) {\parbox{0.2\textwidth} \raggedright \small \textbf{"predict":} "abnorma"}; &
    
            \node[draw, text width=\boxwidth, minimum height=\boxheight] (box2) {};
            \node[fill=labelcolor, draw, text width=\boxwidth, minimum height=\boxheighttop, anchor=north west] at (box2.north west) {\parbox{\boxwidth} \raggedright \small \textbf{"id":} "5"};
            \node[draw, text width=\boxwidth, minimum height=\boxheightmiddle, anchor=north west] at ([yshift=-\boxheighttop]box2.north west) {\parbox{\boxwidth} \raggedright \small \textbf{"label":} "normal"};
            \node[fill=predictcolor, draw, text width=\boxwidth, minimum height=\boxheightbottom, anchor=north west] at ([yshift=-\boxheighttop-\boxheightmiddle]box2.north west) {\parbox{\boxwidth} \raggedright \small \textbf{"predict":} "Based solely on the information provided in the log entry, it is not possible to definitively classify it as 'normal' or 'abnormal'. \textbackslash n\textbackslash nThe log entry mentions \textbackslash"double-hummer alignment exceptions\textbackslash", but without additional context or information about what this means within the specific system or application being monitored, it is difficult to determine whether this is expected behavior or an indication of a problem.\textbackslash n\textbackslash nTherefore, I would classify this log entry as 'unclear' or 'inconclusive' rather than 'normal' or 'abnormal'. It may be necessary to gather more information or consult additional logs or documentation in order to make a more accurate classification."}; \\
        };
    \end{tikzpicture}
    \end{center}
    \caption{A failed zero-shot example by Baichuan2-13B-Chat}
    \label{fig:30}
\end{figure}


\begin{figure}[!ht]
    \begin{center}
    \begin{tikzpicture}
        \definecolor{labelcolor}{RGB}{240,240,240}
        \definecolor{predictcolor}{RGB}{255,228,216}
    
        \newcommand{\boxwidth}{0.4\textwidth} 
        \newcommand{\boxheight}{2cm}
        \newcommand{\boxheighttop}{0.5cm}
        \newcommand{\boxheightmiddle}{0.5cm}
        \newcommand{\boxheightbottom}{1cm}
        \newcommand{\boxgap}{0.005\textwidth} 
    
        \matrix[column sep=\boxgap, row sep=0mm] {
            \node[draw, text width=\boxwidth, minimum height=\boxheight] (box1) {};
            \node[fill=labelcolor, draw, text width=\boxwidth, minimum height=\boxheighttop, anchor=north west] at (box1.north west) {\parbox{\boxwidth} \raggedright \small \textbf{"id":} "0"};
            \node[draw, text width=\boxwidth, minimum height=\boxheightmiddle, anchor=north west] at ([yshift=-\boxheighttop]box1.north west) {\parbox{\boxwidth} \raggedright \small \textbf{"label":} "normal"};
            \node[fill=predictcolor, draw, text width=\boxwidth, minimum height=\boxheightbottom, anchor=north west] at ([yshift=-\boxheighttop-\boxheightmiddle]box1.north west) {\parbox{\boxwidth} \raggedright \small \textbf{"predict":} "normal \textbackslash nnormal \textbackslash nabnormal \textbackslash nabnormal \textbackslash nnormal"}; &
    
            \node[draw, text width=\boxwidth, minimum height=\boxheight] (box2) {};
            \node[fill=labelcolor, draw, text width=\boxwidth, minimum height=\boxheighttop, anchor=north west] at (box2.north west) {\parbox{\boxwidth} \raggedright \small \textbf{"id":} "0"};
            \node[draw, text width=\boxwidth, minimum height=\boxheightmiddle, anchor=north west] at ([yshift=-\boxheighttop]box2.north west) {\parbox{\boxwidth} \raggedright \small \textbf{"label":} "Processor CPU Caterr"};
            \node[fill=predictcolor, draw, text width=\boxwidth, minimum height=\boxheightbottom, anchor=north west] at ([yshift=-\boxheighttop-\boxheightmiddle]box2.north west) {\parbox{\boxwidth} \raggedright \small \textbf{"predict":} "Processor \#0xfa | Configuration Error | Asserted \textbackslash nanswer:'Processor CPU Caterr'"}; \\
        };
    \end{tikzpicture}
    \end{center}
    \caption{A failed few-shot example by Qwen-1.5-7B-Chat}
    \label{fig:31}
\end{figure}


\begin{figure}[!ht]
    \begin{center}
    \begin{tikzpicture}
        \definecolor{labelcolor}{RGB}{240,240,240}
        \definecolor{predictcolor}{RGB}{255,228,216}
    
        \newcommand{\boxwidth}{0.7\textwidth} 
        \newcommand{\boxheight}{3.5cm}
        \newcommand{\boxheighttop}{0.5cm}
        \newcommand{\boxheightmiddle}{0.5cm}
        \newcommand{\boxheightbottom}{2cm}
        \newcommand{\boxgap}{0.005\textwidth} 
    
        \matrix[column sep=\boxgap, row sep=0mm] {
            \node[draw, text width=\boxwidth, minimum height=\boxheight] (box1) {};
            \node[fill=labelcolor, draw, text width=\boxwidth, minimum height=\boxheighttop, anchor=north west] at (box1.north west) {\parbox{\boxwidth} \raggedright \small \textbf{"id":} "176"};
            \node[draw, text width=\boxwidth, minimum height=\boxheightmiddle, anchor=north west] at ([yshift=-\boxheighttop]box1.north west) {\parbox{\boxwidth} \raggedright \small \textbf{"label":} "正常"};
            \node[fill=predictcolor, draw, text width=\boxwidth, minimum height=\boxheightbottom, anchor=north west] at ([yshift=-\boxheighttop-\boxheightmiddle]box1.north west) {\parbox{\boxwidth} \raggedright \small \textbf{"predict":} "正常\textbackslash ninstruction address: 0x00004ed8\textbackslash ninstruction address:s0x00004ed8\textbackslash ninstruction address: 0x00004ed8\textbackslash ninstruction address:0x00004ed8\textbackslash ninstruction address:0x00004ed8\textbackslash ninstruction address:0x00004ed8\textbackslash ninstruction address: 0x00004ed8\textbackslash ninstruction address:0x00004ed8\textbackslash ninstruction address: 0x00004ed8\textbackslash ninstruction address:......"};
            \node[draw, text width=\boxwidth, minimum height=\boxheightmiddle, anchor=north west] at ([yshift=-\boxheighttop-\boxheightmiddle-\boxheightbottom]box1.north west) {\parbox{\boxwidth} \raggedright \small \textbf{"time":} "88.39198112487793"}; \\
        };
    \end{tikzpicture}
    \end{center}
    \caption{A failed zero-shot example by AquilaChat-7B}
    \label{fig:32}
\end{figure}

\bibliographystyle{ACM-Reference-Format}
\bibliography{sample-base}


\begin{thebibliography}{51}


\ifx \showCODEN    \undefined \def \showCODEN     #1{\unskip}     \fi
\ifx \showDOI      \undefined \def \showDOI       #1{#1}\fi
\ifx \showISBNx    \undefined \def \showISBNx     #1{\unskip}     \fi
\ifx \showISBNxiii \undefined \def \showISBNxiii  #1{\unskip}     \fi
\ifx \showISSN     \undefined \def \showISSN      #1{\unskip}     \fi
\ifx \showLCCN     \undefined \def \showLCCN      #1{\unskip}     \fi
\ifx \shownote     \undefined \def \shownote      #1{#1}          \fi
\ifx \showarticletitle \undefined \def \showarticletitle #1{#1}   \fi
\ifx \showURL      \undefined \def \showURL       {\relax}        \fi
\providecommand\bibfield[2]{#2}
\providecommand\bibinfo[2]{#2}
\providecommand\natexlab[1]{#1}
\providecommand\showeprint[2][]{arXiv:#2}

\bibitem[Alibaba(2024)]%
        {qwen}
\bibfield{author}{\bibinfo{person}{Alibaba}.} \bibinfo{year}{2024}\natexlab{}.
\newblock \showarticletitle{https://github.com/QwenLM/Qwen1.5}.
\newblock  (\bibinfo{year}{2024}).
\newblock


\bibitem[Belinkov and Bisk(2017)]%
        {4-DBLP:journals/corr/abs-1711-02173}
\bibfield{author}{\bibinfo{person}{Yonatan Belinkov} {and} \bibinfo{person}{Yonatan Bisk}.} \bibinfo{year}{2017}\natexlab{}.
\newblock \showarticletitle{Synthetic and Natural Noise Both Break Neural Machine Translation}.
\newblock \bibinfo{journal}{\emph{CoRR}}  \bibinfo{volume}{abs/1711.02173} (\bibinfo{year}{2017}).
\newblock
\showeprint[arXiv]{1711.02173}
\urldef\tempurl%
\url{http://arxiv.org/abs/1711.02173}
\showURL{%
\tempurl}


\bibitem[Block et~al\mbox{.}(2023)]%
        {block2023summary}
\bibfield{author}{\bibinfo{person}{Jeremy Block}, \bibinfo{person}{Yu-Peng Chen}, \bibinfo{person}{Abhilash Budharapu}, \bibinfo{person}{Lisa Anthony}, {and} \bibinfo{person}{Bonnie Dorr}.} \bibinfo{year}{2023}\natexlab{}.
\newblock \showarticletitle{Summary Cycles: Exploring the Impact of Prompt Engineering on Large Language Models’ Interaction with Interaction Log Information}. In \bibinfo{booktitle}{\emph{Proceedings of the 4th Workshop on Evaluation and Comparison of NLP Systems}}. \bibinfo{pages}{85--99}.
\newblock


\bibitem[Brown and Mann(2020)]%
        {6-Brown2020LanguageMA}
\bibfield{author}{\bibinfo{person}{Tom~B. Brown} {and} \bibinfo{person}{Benjamin Mann}.} \bibinfo{year}{2020}\natexlab{}.
\newblock \showarticletitle{Language Models are Few-Shot Learners}.
\newblock \bibinfo{journal}{\emph{ArXiv}}  \bibinfo{volume}{abs/2005.14165} (\bibinfo{year}{2020}).
\newblock
\urldef\tempurl%
\url{https://api.semanticscholar.org/CorpusID:218971783}
\showURL{%
\tempurl}


\bibitem[Chang et~al\mbox{.}(2023)]%
        {10-cticang2023survey}
\bibfield{author}{\bibinfo{person}{Yupeng Chang}, \bibinfo{person}{Xu Wang}, \bibinfo{person}{Jindong Wang}, \bibinfo{person}{Yuan Wu}, \bibinfo{person}{Linyi Yang}, \bibinfo{person}{Kaijie Zhu}, \bibinfo{person}{Hao Chen}, \bibinfo{person}{Xiaoyuan Yi}, \bibinfo{person}{Cunxiang Wang}, \bibinfo{person}{Yidong Wang}, \bibinfo{person}{Wei Ye}, \bibinfo{person}{Yue Zhang}, \bibinfo{person}{Yi Chang}, \bibinfo{person}{Philip~S. Yu}, \bibinfo{person}{Qiang Yang}, {and} \bibinfo{person}{Xing Xie}.} \bibinfo{year}{2023}\natexlab{}.
\newblock \showarticletitle{A Survey on Evaluation of Large Language Models}.
\newblock  (\bibinfo{year}{2023}).
\newblock
\showeprint[arxiv]{2307.03109}~[cs.CL]


\bibitem[Chen et~al\mbox{.}(2021)]%
        {11-chen2021evaluating}
\bibfield{author}{\bibinfo{person}{Mark Chen}, \bibinfo{person}{Jerry Tworek}, \bibinfo{person}{Heewoo Jun}, \bibinfo{person}{Qiming Yuan}, \bibinfo{person}{Henrique~Ponde de Oliveira~Pinto}, \bibinfo{person}{Jared Kaplan}, \bibinfo{person}{Harri Edwards}, \bibinfo{person}{Yuri Burda}, \bibinfo{person}{Nicholas Joseph}, \bibinfo{person}{Greg Brockman}, \bibinfo{person}{Alex Ray}, \bibinfo{person}{Raul Puri}, \bibinfo{person}{Gretchen Krueger}, \bibinfo{person}{Michael Petrov}, \bibinfo{person}{Heidy Khlaaf}, \bibinfo{person}{Girish Sastry}, \bibinfo{person}{Pamela Mishkin}, \bibinfo{person}{Brooke Chan}, \bibinfo{person}{Scott Gray}, \bibinfo{person}{Nick Ryder}, \bibinfo{person}{Mikhail Pavlov}, \bibinfo{person}{Alethea Power}, \bibinfo{person}{Lukasz Kaiser}, \bibinfo{person}{Mohammad Bavarian}, \bibinfo{person}{Clemens Winter}, \bibinfo{person}{Philippe Tillet}, \bibinfo{person}{Felipe~Petroski Such}, \bibinfo{person}{Dave Cummings}, \bibinfo{person}{Matthias Plappert}, \bibinfo{person}{Fotios Chantzis},
  \bibinfo{person}{Elizabeth Barnes}, \bibinfo{person}{Ariel Herbert-Voss}, \bibinfo{person}{William~Hebgen Guss}, \bibinfo{person}{Alex Nichol}, \bibinfo{person}{Alex Paino}, \bibinfo{person}{Nikolas Tezak}, \bibinfo{person}{Jie Tang}, \bibinfo{person}{Igor Babuschkin}, \bibinfo{person}{Suchir Balaji}, \bibinfo{person}{Shantanu Jain}, \bibinfo{person}{William Saunders}, \bibinfo{person}{Christopher Hesse}, \bibinfo{person}{Andrew~N. Carr}, \bibinfo{person}{Jan Leike}, \bibinfo{person}{Josh Achiam}, \bibinfo{person}{Vedant Misra}, \bibinfo{person}{Evan Morikawa}, \bibinfo{person}{Alec Radford}, \bibinfo{person}{Matthew Knight}, \bibinfo{person}{Miles Brundage}, \bibinfo{person}{Mira Murati}, \bibinfo{person}{Katie Mayer}, \bibinfo{person}{Peter Welinder}, \bibinfo{person}{Bob McGrew}, \bibinfo{person}{Dario Amodei}, \bibinfo{person}{Sam McCandlish}, \bibinfo{person}{Ilya Sutskever}, {and} \bibinfo{person}{Wojciech Zaremba}.} \bibinfo{year}{2021}\natexlab{}.
\newblock \showarticletitle{Evaluating Large Language Models Trained on Code}.
\newblock  (\bibinfo{year}{2021}).
\newblock
\showeprint[arxiv]{2107.03374}~[cs.LG]


\bibitem[Egersdoerfer et~al\mbox{.}(2023)]%
        {egersdoerfer2023early}
\bibfield{author}{\bibinfo{person}{Chris Egersdoerfer}, \bibinfo{person}{Di Zhang}, {and} \bibinfo{person}{Dong Dai}.} \bibinfo{year}{2023}\natexlab{}.
\newblock \showarticletitle{Early exploration of using ChatGPT for log-based anomaly detection on parallel file systems logs}. In \bibinfo{booktitle}{\emph{Proceedings of the 32nd International Symposium on High-Performance Parallel and Distributed Computing}}. \bibinfo{pages}{315--316}.
\newblock


\bibitem[et~al.(2023)]%
        {logsummary}
\bibfield{author}{\bibinfo{person}{W.~Meng et al.}} \bibinfo{year}{2023}\natexlab{}.
\newblock \showarticletitle{LogSummary: Unstructured Log Summarization for Software Systems.}
\newblock \bibinfo{journal}{\emph{IEEE Transactions on Network and Service Management}} \bibinfo{volume}{20}, \bibinfo{number}{3} (\bibinfo{year}{2023}), \bibinfo{pages}{3803--3815}.
\newblock


\bibitem[et.al.(2023)]%
        {llama2}
\bibfield{author}{\bibinfo{person}{Hugo~Touvron et.al.}} \bibinfo{year}{2023}\natexlab{}.
\newblock \showarticletitle{Llama 2: Open Foundation and Fine-Tuned Chat Models}.
\newblock  (\bibinfo{year}{2023}).
\newblock


\bibitem[He et~al\mbox{.}(2017)]%
        {drain}
\bibfield{author}{\bibinfo{person}{Pinjia He}, \bibinfo{person}{Jieming Zhu}, \bibinfo{person}{Zibin Zheng}, {and} \bibinfo{person}{Michael~R. Lyu}.} \bibinfo{year}{2017}\natexlab{}.
\newblock \showarticletitle{Drain: An Online Log Parsing Approach with Fixed Depth Tree}. In \bibinfo{booktitle}{\emph{2017 IEEE International Conference on Web Services (ICWS)}}. \bibinfo{pages}{33--40}.
\newblock
\urldef\tempurl%
\url{https://doi.org/10.1109/ICWS.2017.13}
\showDOI{\tempurl}


\bibitem[He et~al\mbox{.}(2023)]%
        {logpub2}
\bibfield{author}{\bibinfo{person}{S. He}, \bibinfo{person}{J. Zhu}, \bibinfo{person}{P. He}, {and} \bibinfo{person}{M.~R. Lyu}.} \bibinfo{year}{2023}\natexlab{}.
\newblock \showarticletitle{Loghub: A large collection of system log datasets towards automated log analytics}.
\newblock \bibinfo{journal}{\emph{arXiv e-prints}} (\bibinfo{year}{2023}).
\newblock


\bibitem[Hoffmann et~al\mbox{.}(2022)]%
        {36-hoffmann2022training}
\bibfield{author}{\bibinfo{person}{Jordan Hoffmann}, \bibinfo{person}{Sebastian Borgeaud}, \bibinfo{person}{Arthur Mensch}, \bibinfo{person}{Elena Buchatskaya}, \bibinfo{person}{Trevor Cai}, \bibinfo{person}{Eliza Rutherford}, \bibinfo{person}{Diego de Las~Casas}, \bibinfo{person}{Lisa~Anne Hendricks}, \bibinfo{person}{Johannes Welbl}, \bibinfo{person}{Aidan Clark}, \bibinfo{person}{Tom Hennigan}, \bibinfo{person}{Eric Noland}, \bibinfo{person}{Katie Millican}, \bibinfo{person}{George van~den Driessche}, \bibinfo{person}{Bogdan Damoc}, \bibinfo{person}{Aurelia Guy}, \bibinfo{person}{Simon Osindero}, \bibinfo{person}{Karen Simonyan}, \bibinfo{person}{Erich Elsen}, \bibinfo{person}{Jack~W. Rae}, \bibinfo{person}{Oriol Vinyals}, {and} \bibinfo{person}{Laurent Sifre}.} \bibinfo{year}{2022}\natexlab{}.
\newblock \bibinfo{title}{Training Compute-Optimal Large Language Models}.
\newblock
\newblock
\showeprint[arxiv]{2203.15556}~[cs.CL]


\bibitem[Jiang et~al\mbox{.}(2023)]%
        {llmparser}
\bibfield{author}{\bibinfo{person}{Zhihan Jiang}, \bibinfo{person}{Jinyang Liu}, \bibinfo{person}{Zhuangbin Chen}, \bibinfo{person}{Yichen Li}, \bibinfo{person}{Junjie Huang}, \bibinfo{person}{Yintong Huo}, \bibinfo{person}{Pinjia He}, \bibinfo{person}{Jiazhen Gu}, {and} \bibinfo{person}{Michael~R. Lyu}.} \bibinfo{year}{2023}\natexlab{}.
\newblock \showarticletitle{Llmparser: A llm-based log parsing framework}.
\newblock \bibinfo{journal}{\emph{arXiv preprint arXiv:2310.01796}} (\bibinfo{year}{2023}).
\newblock


\bibitem[Jiang et~al\mbox{.}(2020)]%
        {logpub1}
\bibfield{author}{\bibinfo{person}{Z. Jiang}, \bibinfo{person}{J. Liu}, \bibinfo{person}{J. Huang}, \bibinfo{person}{Y. Li}, \bibinfo{person}{Y. Huo}, \bibinfo{person}{J. Gu}, \bibinfo{person}{Z. Chen}, \bibinfo{person}{J. Zhu}, {and} \bibinfo{person}{M.~R. Lyu}.} \bibinfo{year}{2020}\natexlab{}.
\newblock \showarticletitle{A large-scale benchmark for log parsing}.
\newblock \bibinfo{journal}{\emph{arXiv print}} (\bibinfo{year}{2020}).
\newblock


\bibitem[Karlsen et~al\mbox{.}(2023a)]%
        {karlsen2023exploring}
\bibfield{author}{\bibinfo{person}{Egil Karlsen}, \bibinfo{person}{Rafael Copstein}, \bibinfo{person}{Xiao Luo}, \bibinfo{person}{Jeff Schwartzentruber}, \bibinfo{person}{Bradley Niblett}, \bibinfo{person}{Andrew Johnston}, \bibinfo{person}{Malcolm~I. Heywood}, {and} \bibinfo{person}{Nur Zincir-Heywood}.} \bibinfo{year}{2023}\natexlab{a}.
\newblock \showarticletitle{Exploring semantic vs. syntactic features for unsupervised learning on application log files}. In \bibinfo{booktitle}{\emph{2023 7th Cyber Security in Networking Conference (CSNet)}}. IEEE, \bibinfo{pages}{219--225}.
\newblock


\bibitem[Karlsen et~al\mbox{.}(2023b)]%
        {karlsen2023benchmarking}
\bibfield{author}{\bibinfo{person}{Egil Karlsen}, \bibinfo{person}{Xiao Luo}, \bibinfo{person}{Nur Zincir-Heywood}, {and} \bibinfo{person}{Malcolm Heywood}.} \bibinfo{year}{2023}\natexlab{b}.
\newblock \showarticletitle{Benchmarking Large Language Models for Log Analysis, Security, and Interpretation}.
\newblock \bibinfo{journal}{\emph{arXiv preprint arXiv:2311.14519}} (\bibinfo{year}{2023}).
\newblock


\bibitem[Le and Zhang(2021)]%
        {neurallog}
\bibfield{author}{\bibinfo{person}{Van-Hoang Le} {and} \bibinfo{person}{Hongyu Zhang}.} \bibinfo{year}{2021}\natexlab{}.
\newblock \showarticletitle{Log-based Anomaly Detection Without Log Parsing}. In \bibinfo{booktitle}{\emph{2021 36th IEEE/ACM International Conference on Automated Software Engineering (ASE)}}. \bibinfo{pages}{492--504}.
\newblock
\urldef\tempurl%
\url{https://doi.org/10.1109/ASE51524.2021.9678773}
\showDOI{\tempurl}


\bibitem[Le and Zhang(2023)]%
        {logppt}
\bibfield{author}{\bibinfo{person}{Van-Hoang Le} {and} \bibinfo{person}{Hongyu Zhang}.} \bibinfo{year}{2023}\natexlab{}.
\newblock \showarticletitle{Log Parsing with Prompt-based Few-shot Learning}. In \bibinfo{booktitle}{\emph{2023 IEEE/ACM 45th International Conference on Software Engineering (ICSE)}}. \bibinfo{pages}{2438--2449}.
\newblock
\urldef\tempurl%
\url{https://doi.org/10.1109/ICSE48619.2023.00204}
\showDOI{\tempurl}


\bibitem[Lerner(2017)]%
        {gartner}
\bibfield{author}{\bibinfo{person}{Andrew Lerner}.} \bibinfo{year}{2017}\natexlab{}.
\newblock \showarticletitle{AIOps Platforms—Gartner}.
\newblock  (\bibinfo{year}{2017}).
\newblock


\bibitem[Li et~al\mbox{.}(2023)]%
        {Huatuo-26M}
\bibfield{author}{\bibinfo{person}{Jianquan Li}, \bibinfo{person}{Xidong Wang}, \bibinfo{person}{Xiangbo Wu}, \bibinfo{person}{Zhiyi Zhang}, \bibinfo{person}{Xiaolong Xu}, \bibinfo{person}{Jie Fu}, \bibinfo{person}{Prayag Tiwari}, \bibinfo{person}{Xiang Wan}, {and} \bibinfo{person}{Benyou Wang.}} \bibinfo{year}{2023}\natexlab{}.
\newblock \showarticletitle{Huatuo-26M,a Large-scale Chinese Medical QA Dataset}.
\newblock \bibinfo{journal}{\emph{arXiv e-prints (2023)}} (\bibinfo{year}{2023}).
\newblock


\bibitem[Li et~al\mbox{.}(2022)]%
        {46-Li_2022}
\bibfield{author}{\bibinfo{person}{Yujia Li}, \bibinfo{person}{David Choi}, \bibinfo{person}{Junyoung Chung}, \bibinfo{person}{Nate Kushman}, \bibinfo{person}{Julian Schrittwieser}, \bibinfo{person}{Rémi Leblond}, \bibinfo{person}{Tom Eccles}, \bibinfo{person}{James Keeling}, \bibinfo{person}{Felix Gimeno}, \bibinfo{person}{Agustin~Dal Lago}, \bibinfo{person}{Thomas Hubert}, \bibinfo{person}{Peter Choy}, \bibinfo{person}{Cyprien de Masson~d’Autume}, \bibinfo{person}{Igor Babuschkin}, \bibinfo{person}{Xinyun Chen}, \bibinfo{person}{Po-Sen Huang}, \bibinfo{person}{Johannes Welbl}, \bibinfo{person}{Sven Gowal}, \bibinfo{person}{Alexey Cherepanov}, \bibinfo{person}{James Molloy}, \bibinfo{person}{Daniel~J. Mankowitz}, \bibinfo{person}{Esme~Sutherland Robson}, \bibinfo{person}{Pushmeet Kohli}, \bibinfo{person}{Nando de Freitas}, \bibinfo{person}{Koray Kavukcuoglu}, {and} \bibinfo{person}{Oriol Vinyals}.} \bibinfo{year}{2022}\natexlab{}.
\newblock \showarticletitle{Competition-level code generation with AlphaCode}.
\newblock \bibinfo{journal}{\emph{Science}} \bibinfo{volume}{378}, \bibinfo{number}{6624} (\bibinfo{date}{Dec.} \bibinfo{year}{2022}), \bibinfo{pages}{1092--1097}.
\newblock
\showISSN{1095-9203}
\urldef\tempurl%
\url{https://doi.org/10.1126/science.abq1158}
\showDOI{\tempurl}


\bibitem[Liang et~al\mbox{.}(2022)]%
        {helm}
\bibfield{author}{\bibinfo{person}{Percy Liang}, \bibinfo{person}{Rishi Bommasani}, \bibinfo{person}{Tony Lee}, \bibinfo{person}{Dimitris Tsipras}, \bibinfo{person}{Dilara Soylu}, \bibinfo{person}{Michihiro Yasunaga}, \bibinfo{person}{Yian Zhang}, \bibinfo{person}{Deepak Narayanan}, \bibinfo{person}{Yuhuai Wu}, \bibinfo{person}{Ananya Kumar}, {et~al\mbox{.}}} \bibinfo{year}{2022}\natexlab{}.
\newblock \showarticletitle{Holistic Evaluation of Language Models}.
\newblock \bibinfo{journal}{\emph{arXiv e-prints}} (\bibinfo{year}{2022}).
\newblock


\bibitem[Lin({[n.\,d.]})]%
        {rouge}
\bibfield{author}{\bibinfo{person}{Chin-Yew Lin}.} \bibinfo{year}{[n.\,d.]}\natexlab{}.
\newblock \showarticletitle{ROUGE: A Package for Automatic Evaluation of Summaries. In Text Summarization Branches Out. Association for Computational Linguistics}.
\newblock \bibinfo{journal}{\emph{Association for Computational Linguistics, Barcelona, Spain}} (\bibinfo{year}{[n.\,d.]}).
\newblock


\bibitem[Lin et~al\mbox{.}(2016)]%
        {logcluster}
\bibfield{author}{\bibinfo{person}{Qingwei Lin}, \bibinfo{person}{Hongyu Zhang}, \bibinfo{person}{Jian-Guang Lou}, \bibinfo{person}{Yu Zhang}, {and} \bibinfo{person}{Xuewei Chen}.} \bibinfo{year}{2016}\natexlab{}.
\newblock \showarticletitle{Log Clustering Based Problem Identification for Online Service Systems}. In \bibinfo{booktitle}{\emph{2016 IEEE/ACM 38th International Conference on Software Engineering Companion (ICSE-C)}}. \bibinfo{pages}{102--111}.
\newblock


\bibitem[Liu et~al\mbox{.}(2023a)]%
        {liu2023scalable}
\bibfield{author}{\bibinfo{person}{Jinyang Liu}, \bibinfo{person}{Junjie Huang}, \bibinfo{person}{Yintong Huo}, \bibinfo{person}{Zhihan Jiang}, \bibinfo{person}{Jiazhen Gu}, \bibinfo{person}{Zhuangbin Chen}, \bibinfo{person}{Cong Feng}, \bibinfo{person}{Minzhi Yan}, {and} \bibinfo{person}{Michael~R. Lyu}.} \bibinfo{year}{2023}\natexlab{a}.
\newblock \showarticletitle{Scalable and adaptive log-based anomaly detection with expert in the loop}.
\newblock \bibinfo{journal}{\emph{arXiv preprint arXiv:2306.05032}} (\bibinfo{year}{2023}).
\newblock


\bibitem[Liu et~al\mbox{.}(2023b)]%
        {Opseval}
\bibfield{author}{\bibinfo{person}{Yuhe Liu}, \bibinfo{person}{Changhua Pei}, \bibinfo{person}{Longlong Xu}, \bibinfo{person}{Bohan Chen}, \bibinfo{person}{Mingze Sun}, \bibinfo{person}{Zhirui Zhang}, \bibinfo{person}{Yongqian Sun}, \bibinfo{person}{Shenglin Zhang}, \bibinfo{person}{Kun Wang}, \bibinfo{person}{Haiming Zhang}, \bibinfo{person}{Jianhui Li}, \bibinfo{person}{Gaogang Xie}, \bibinfo{person}{Xidao Wen}, \bibinfo{person}{Xiaohui Nie}, \bibinfo{person}{Minghua Ma}, {and} \bibinfo{person}{Dan Pei.}} \bibinfo{year}{2023}\natexlab{b}.
\newblock \showarticletitle{OpsEval: A Comprehensive IT Operations Benchmark Suite for Large Language Models.}
\newblock \bibinfo{journal}{\emph{arXiv e-prints (2023)}} (\bibinfo{year}{2023}).
\newblock


\bibitem[Miao et~al\mbox{.}(2023)]%
        {NetOps}
\bibfield{author}{\bibinfo{person}{Yukai Miao}, \bibinfo{person}{Yu Bai}, \bibinfo{person}{Haifeng~Sun Li~Chen, Dan~Li}, \bibinfo{person}{Xizheng Wang}, \bibinfo{person}{Ziqiu Luo}, \bibinfo{person}{Dapeng Sun}, \bibinfo{person}{Xiuting Xu}, \bibinfo{person}{Qi Zhang}, \bibinfo{person}{Chao Xiang}, {and} \bibinfo{person}{Xinchi Li.}} \bibinfo{year}{2023}\natexlab{}.
\newblock \showarticletitle{An Empirical Study of NetOps Capability of Pre-Trained Large Language Models.}
\newblock \bibinfo{journal}{\emph{arXiv e-prints (2023)}} (\bibinfo{year}{2023}).
\newblock


\bibitem[Nakano et~al\mbox{.}(2022)]%
        {54-nakano2022webgpt}
\bibfield{author}{\bibinfo{person}{Reiichiro Nakano}, \bibinfo{person}{Jacob Hilton}, \bibinfo{person}{Suchir Balaji}, \bibinfo{person}{Jeff Wu}, \bibinfo{person}{Long Ouyang}, \bibinfo{person}{Christina Kim}, \bibinfo{person}{Christopher Hesse}, \bibinfo{person}{Shantanu Jain}, \bibinfo{person}{Vineet Kosaraju}, \bibinfo{person}{William Saunders}, \bibinfo{person}{Xu Jiang}, \bibinfo{person}{Karl Cobbe}, \bibinfo{person}{Tyna Eloundou}, \bibinfo{person}{Gretchen Krueger}, \bibinfo{person}{Kevin Button}, \bibinfo{person}{Matthew Knight}, \bibinfo{person}{Benjamin Chess}, {and} \bibinfo{person}{John Schulman}.} \bibinfo{year}{2022}\natexlab{}.
\newblock \bibinfo{title}{WebGPT: Browser-assisted question-answering with human feedback}.
\newblock
\newblock
\showeprint[arxiv]{2112.09332}~[cs.CL]


\bibitem[OpenAI(2023)]%
        {GPT4}
\bibfield{author}{\bibinfo{person}{OpenAI}.} \bibinfo{year}{2023}\natexlab{}.
\newblock \showarticletitle{GPT-4 Technical Report}.
\newblock  (\bibinfo{year}{2023}).
\newblock


\bibitem[Ouyang et~al\mbox{.}(2022)]%
        {57-NEURIPS2022_b1efde53}
\bibfield{author}{\bibinfo{person}{Long Ouyang}, \bibinfo{person}{Jeffrey Wu}, \bibinfo{person}{Xu Jiang}, \bibinfo{person}{Diogo Almeida}, \bibinfo{person}{Carroll Wainwright}, \bibinfo{person}{Pamela Mishkin}, \bibinfo{person}{Chong Zhang}, \bibinfo{person}{Sandhini Agarwal}, \bibinfo{person}{Katarina Slama}, \bibinfo{person}{Alex Ray}, \bibinfo{person}{John Schulman}, \bibinfo{person}{Jacob Hilton}, \bibinfo{person}{Fraser Kelton}, \bibinfo{person}{Luke Miller}, \bibinfo{person}{Maddie Simens}, \bibinfo{person}{Amanda Askell}, \bibinfo{person}{Peter Welinder}, \bibinfo{person}{Paul~F. Christiano}, \bibinfo{person}{Jan Leike}, {and} \bibinfo{person}{Ryan Lowe}.} \bibinfo{year}{2022}\natexlab{}.
\newblock \showarticletitle{Training language models to follow instructions with human feedback}. In \bibinfo{booktitle}{\emph{Advances in Neural Information Processing Systems}}, \bibfield{editor}{\bibinfo{person}{S.~Koyejo}, \bibinfo{person}{S.~Mohamed}, \bibinfo{person}{A.~Agarwal}, \bibinfo{person}{D.~Belgrave}, \bibinfo{person}{K.~Cho}, {and} \bibinfo{person}{A.~Oh}} (Eds.), Vol.~\bibinfo{volume}{35}. \bibinfo{publisher}{Curran Associates, Inc.}, \bibinfo{pages}{27730--27744}.
\newblock
\urldef\tempurl%
\url{https://proceedings.neurips.cc/paper_files/paper/2022/file/b1efde53be364a73914f58805a001731-Paper-Conference.pdf}
\showURL{%
\tempurl}


\bibitem[Qi et~al\mbox{.}(2023)]%
        {qi2023loggpt}
\bibfield{author}{\bibinfo{person}{Jiaxing Qi}, \bibinfo{person}{Shaohan Huang}, \bibinfo{person}{Zhongzhi Luan}, \bibinfo{person}{Shu Yang}, \bibinfo{person}{Carol Fung}, \bibinfo{person}{Hailong Yang}, \bibinfo{person}{Depei Qian}, \bibinfo{person}{Jing Shang}, \bibinfo{person}{Zhiwen Xiao}, {and} \bibinfo{person}{Zhihui Wu}.} \bibinfo{year}{2023}\natexlab{}.
\newblock \showarticletitle{Loggpt: Exploring chatgpt for log-based anomaly detection}. In \bibinfo{booktitle}{\emph{2023 IEEE International Conference on High Performance Computing \& Communications, Data Science \& Systems, Smart City \& Dependability in Sensor, Cloud \& Big Data Systems \& Application (HPCC/DSS/SmartCity/DependSys)}}. IEEE, \bibinfo{pages}{273--280}.
\newblock


\bibitem[Radford et~al\mbox{.}(2019)]%
        {61-Radford2019LanguageMA}
\bibfield{author}{\bibinfo{person}{Alec Radford}, \bibinfo{person}{Jeff Wu}, \bibinfo{person}{Rewon Child}, \bibinfo{person}{David Luan}, \bibinfo{person}{Dario Amodei}, {and} \bibinfo{person}{Ilya Sutskever}.} \bibinfo{year}{2019}\natexlab{}.
\newblock \showarticletitle{Language Models are Unsupervised Multitask Learners}.
\newblock  (\bibinfo{year}{2019}).
\newblock
\urldef\tempurl%
\url{https://api.semanticscholar.org/CorpusID:160025533}
\showURL{%
\tempurl}


\bibitem[Roziere et~al\mbox{.}(2023)]%
        {roziere2023code}
\bibfield{author}{\bibinfo{person}{Baptiste Roziere}, \bibinfo{person}{Jonas Gehring}, \bibinfo{person}{Fabian Gloeckle}, \bibinfo{person}{Sten Sootla}, \bibinfo{person}{Itai Gat}, \bibinfo{person}{Xiaoqing~Ellen Tan}, \bibinfo{person}{Yossi Adi}, \bibinfo{person}{Jingyu Liu}, \bibinfo{person}{Tal Remez}, \bibinfo{person}{J{\'e}r{\'e}my Rapin}, {et~al\mbox{.}}} \bibinfo{year}{2023}\natexlab{}.
\newblock \showarticletitle{Code llama: Open foundation models for code}.
\newblock \bibinfo{journal}{\emph{arXiv preprint arXiv:2308.12950}} (\bibinfo{year}{2023}).
\newblock


\bibitem[Shan et~al\mbox{.}(2024)]%
        {shan2024face}
\bibfield{author}{\bibinfo{person}{Shiwen Shan}, \bibinfo{person}{Yintong Huo}, \bibinfo{person}{Yuxin Su}, \bibinfo{person}{Yichen Li}, \bibinfo{person}{Dan Li}, {and} \bibinfo{person}{Zibin Zheng}.} \bibinfo{year}{2024}\natexlab{}.
\newblock \showarticletitle{Face It Yourselves: An LLM-Based Two-Stage Strategy to Localize Configuration Errors via Logs}.
\newblock \bibinfo{journal}{\emph{arXiv preprint arXiv:2404.00640}} (\bibinfo{year}{2024}).
\newblock


\bibitem[Singhal et~al\mbox{.}(2022)]%
        {MultiMedQA}
\bibfield{author}{\bibinfo{person}{Karan Singhal}, \bibinfo{person}{Shekoofeh Azizi}, \bibinfo{person}{Tao Tu}, \bibinfo{person}{S~Sara Mahdavi}, \bibinfo{person}{Jason Wei}, \bibinfo{person}{Hyung~Won Chung}, \bibinfo{person}{Nathan Scales}, \bibinfo{person}{Ajay Tanwani}, \bibinfo{person}{Heather Cole-Lewis}, \bibinfo{person}{Stephen Pfohl}, {et~al\mbox{.}}} \bibinfo{year}{2022}\natexlab{}.
\newblock \showarticletitle{Large Language Models Encode Clinical Knowledge}.
\newblock \bibinfo{journal}{\emph{arXiv e-prints}} (\bibinfo{year}{2022}).
\newblock


\bibitem[Srivastava et~al\mbox{.}(2022)]%
        {big-bench}
\bibfield{author}{\bibinfo{person}{Aarohi Srivastava}, \bibinfo{person}{Abhinav Rastogi}, \bibinfo{person}{Abhishek Rao}, \bibinfo{person}{Abu Awal~Md Shoeb}, \bibinfo{person}{Abubakar Abid}, \bibinfo{person}{Adam Fisch}, \bibinfo{person}{Adam~R Brown}, \bibinfo{person}{Adam Santoro}, \bibinfo{person}{Aditya Gupta}, \bibinfo{person}{Adrià Garriga-Alonso}, {et~al\mbox{.}}} \bibinfo{year}{2022}\natexlab{}.
\newblock \showarticletitle{Beyond the Imitation Game: Quantifying and Extrapolating the Capabilities of Language Models}.
\newblock \bibinfo{journal}{\emph{arXiv e-prints}} (\bibinfo{year}{2022}).
\newblock


\bibitem[Sui et~al\mbox{.}(2023)]%
        {logkg}
\bibfield{author}{\bibinfo{person}{Yicheng Sui}, \bibinfo{person}{Yuzhe Zhang}, \bibinfo{person}{Jianjun Sun}, \bibinfo{person}{Ting Xu}, \bibinfo{person}{Shenglin Zhang}, \bibinfo{person}{Zhengdan Li}, \bibinfo{person}{Yongqian Sun}, \bibinfo{person}{Fangrui Guo}, \bibinfo{person}{Junyu Shen}, \bibinfo{person}{Yuzhi Zhang}, \bibinfo{person}{Dan Pei}, \bibinfo{person}{Xiao Yang}, {and} \bibinfo{person}{Li Yu}.} \bibinfo{year}{2023}\natexlab{}.
\newblock \showarticletitle{LogKG: Log Failure Diagnosis Through Knowledge Graph}.
\newblock \bibinfo{journal}{\emph{IEEE Transactions on Services Computing}} \bibinfo{volume}{16}, \bibinfo{number}{5} (\bibinfo{year}{2023}), \bibinfo{pages}{3493--3507}.
\newblock
\urldef\tempurl%
\url{https://doi.org/10.1109/TSC.2023.3293890}
\showDOI{\tempurl}


\bibitem[THUDM(2024)]%
        {chatglm4}
\bibfield{author}{\bibinfo{person}{THUDM}.} \bibinfo{year}{2024}\natexlab{}.
\newblock \showarticletitle{THUDM/ChatGLM4}.
\newblock \bibinfo{journal}{\emph{https://github.com/THUDM/ChatGLM4}} (\bibinfo{year}{2024}).
\newblock


\bibitem[Touvron et~al\mbox{.}(2023)]%
        {73-touvron2023llama}
\bibfield{author}{\bibinfo{person}{Hugo Touvron}, \bibinfo{person}{Louis Martin}, \bibinfo{person}{Kevin Stone}, \bibinfo{person}{Peter Albert}, \bibinfo{person}{Amjad Almahairi}, \bibinfo{person}{Yasmine Babaei}, \bibinfo{person}{Nikolay Bashlykov}, \bibinfo{person}{Soumya Batra}, \bibinfo{person}{Prajjwal Bhargava}, \bibinfo{person}{Shruti Bhosale}, \bibinfo{person}{Dan Bikel}, \bibinfo{person}{Lukas Blecher}, \bibinfo{person}{Cristian~Canton Ferrer}, \bibinfo{person}{Moya Chen}, \bibinfo{person}{Guillem Cucurull}, \bibinfo{person}{David Esiobu}, \bibinfo{person}{Jude Fernandes}, \bibinfo{person}{Jeremy Fu}, \bibinfo{person}{Wenyin Fu}, \bibinfo{person}{Brian Fuller}, \bibinfo{person}{Cynthia Gao}, \bibinfo{person}{Vedanuj Goswami}, \bibinfo{person}{Naman Goyal}, \bibinfo{person}{Anthony Hartshorn}, \bibinfo{person}{Saghar Hosseini}, \bibinfo{person}{Rui Hou}, \bibinfo{person}{Hakan Inan}, \bibinfo{person}{Marcin Kardas}, \bibinfo{person}{Viktor Kerkez}, \bibinfo{person}{Madian Khabsa},
  \bibinfo{person}{Isabel Kloumann}, \bibinfo{person}{Artem Korenev}, \bibinfo{person}{Punit~Singh Koura}, \bibinfo{person}{Marie-Anne Lachaux}, \bibinfo{person}{Thibaut Lavril}, \bibinfo{person}{Jenya Lee}, \bibinfo{person}{Diana Liskovich}, \bibinfo{person}{Yinghai Lu}, \bibinfo{person}{Yuning Mao}, \bibinfo{person}{Xavier Martinet}, \bibinfo{person}{Todor Mihaylov}, \bibinfo{person}{Pushkar Mishra}, \bibinfo{person}{Igor Molybog}, \bibinfo{person}{Yixin Nie}, \bibinfo{person}{Andrew Poulton}, \bibinfo{person}{Jeremy Reizenstein}, \bibinfo{person}{Rashi Rungta}, \bibinfo{person}{Kalyan Saladi}, \bibinfo{person}{Alan Schelten}, \bibinfo{person}{Ruan Silva}, \bibinfo{person}{Eric~Michael Smith}, \bibinfo{person}{Ranjan Subramanian}, \bibinfo{person}{Xiaoqing~Ellen Tan}, \bibinfo{person}{Binh Tang}, \bibinfo{person}{Ross Taylor}, \bibinfo{person}{Adina Williams}, \bibinfo{person}{Jian~Xiang Kuan}, \bibinfo{person}{Puxin Xu}, \bibinfo{person}{Zheng Yan}, \bibinfo{person}{Iliyan Zarov}, \bibinfo{person}{Yuchen
  Zhang}, \bibinfo{person}{Angela Fan}, \bibinfo{person}{Melanie Kambadur}, \bibinfo{person}{Sharan Narang}, \bibinfo{person}{Aurelien Rodriguez}, \bibinfo{person}{Robert Stojnic}, \bibinfo{person}{Sergey Edunov}, {and} \bibinfo{person}{Thomas Scialom}.} \bibinfo{year}{2023}\natexlab{}.
\newblock \showarticletitle{Llama 2: Open Foundation and Fine-Tuned Chat Models}.
\newblock  (\bibinfo{year}{2023}).
\newblock
\showeprint[arxiv]{2307.09288}~[cs.CL]


\bibitem[Wang et~al\mbox{.}(2022)]%
        {sc}
\bibfield{author}{\bibinfo{person}{Xuezhi Wang}, \bibinfo{person}{Jason Wei}, \bibinfo{person}{Dale Schuurmans}, \bibinfo{person}{Quoc Le}, \bibinfo{person}{Ed Chi}, \bibinfo{person}{Sharan Narang}, \bibinfo{person}{Aakanksha Chowdhery}, {and} \bibinfo{person}{Denny Zhou}.} \bibinfo{year}{2022}\natexlab{}.
\newblock \showarticletitle{Self-Consistency Improves Chain of Thought Reasoning in Language Models}.
\newblock \bibinfo{journal}{\emph{arXiv:2203.11171 [cs.CL]}} (\bibinfo{year}{2022}).
\newblock


\bibitem[Workshop et~al\mbox{.}(2023)]%
        {67-workshop2023bloom}
\bibfield{author}{\bibinfo{person}{BigScience Workshop}, \bibinfo{person}{Teven~Le Scao}, \bibinfo{person}{Angela Fan}, \bibinfo{person}{Christopher Akiki}, \bibinfo{person}{Ellie Pavlick}, \bibinfo{person}{Suzana Ilić}, \bibinfo{person}{Daniel Hesslow}, \bibinfo{person}{Roman Castagné}, \bibinfo{person}{Alexandra~Sasha Luccioni}, \bibinfo{person}{François Yvon}, \bibinfo{person}{Matthias Gallé}, \bibinfo{person}{Jonathan Tow}, {and} \bibinfo{person}{Alexander~M. Rush}.} \bibinfo{year}{2023}\natexlab{}.
\newblock \bibinfo{title}{BLOOM: A 176B-Parameter Open-Access Multilingual Language Model}.
\newblock
\newblock
\showeprint[arxiv]{2211.05100}~[cs.CL]


\bibitem[Xu et~al\mbox{.}(2024)]%
        {unilog}
\bibfield{author}{\bibinfo{person}{Junjielong Xu}, \bibinfo{person}{Ziang Cui}, \bibinfo{person}{Yuan Zhao}, \bibinfo{person}{Xu Zhang}, \bibinfo{person}{Shilin He}, \bibinfo{person}{Pinjia He}, \bibinfo{person}{Liqun Li}, \bibinfo{person}{Yu Kang}, \bibinfo{person}{Qingwei Lin}, \bibinfo{person}{Yingnong Dang}, {et~al\mbox{.}}} \bibinfo{year}{2024}\natexlab{}.
\newblock \showarticletitle{UniLog: Automatic Logging via LLM and In-Context Learning}. In \bibinfo{booktitle}{\emph{Proceedings of the 46th IEEE/ACM International Conference on Software Engineering}}. \bibinfo{pages}{1--12}.
\newblock


\bibitem[Xu et~al\mbox{.}(2023)]%
        {xu2023prompting}
\bibfield{author}{\bibinfo{person}{Junjielong Xu}, \bibinfo{person}{Ruichun Yang}, \bibinfo{person}{Yintong Huo}, \bibinfo{person}{Chengyu Zhang}, {and} \bibinfo{person}{Pinjia He}.} \bibinfo{year}{2023}\natexlab{}.
\newblock \showarticletitle{Prompting for Automatic Log Template Extraction}.
\newblock \bibinfo{journal}{\emph{arXiv preprint arXiv:2307.09950}} (\bibinfo{year}{2023}).
\newblock


\bibitem[Zeng.(2023)]%
        {mmcu}
\bibfield{author}{\bibinfo{person}{Hui Zeng.}} \bibinfo{year}{2023}\natexlab{}.
\newblock \showarticletitle{Measuring Massive Multitask Chinese Understanding.}
\newblock \bibinfo{journal}{\emph{arXiv e-prints (2023)}} (\bibinfo{year}{2023}).
\newblock


\bibitem[Zeng et~al\mbox{.}(2023)]%
        {cgeval}
\bibfield{author}{\bibinfo{person}{Hui Zeng}, \bibinfo{person}{Jingyuan Xue}, \bibinfo{person}{Meng Hao}, \bibinfo{person}{Chen Sun}, \bibinfo{person}{Bin Ning}, {and} \bibinfo{person}{Na Zhang}.} \bibinfo{year}{2023}\natexlab{}.
\newblock \showarticletitle{Evaluating the Generation Capabilities of Large Chinese Language Models}.
\newblock \bibinfo{journal}{\emph{arXiv e-prints}} (\bibinfo{year}{2023}).
\newblock


\bibitem[Zhang et~al\mbox{.}(2023)]%
        {fineval}
\bibfield{author}{\bibinfo{person}{Liwen Zhang}, \bibinfo{person}{Weige Cai}, \bibinfo{person}{Zhaowei Liu}, \bibinfo{person}{Zhi Yang}, \bibinfo{person}{Wei Dai}, \bibinfo{person}{Yujie Liao}, \bibinfo{person}{Qianru Qin}, \bibinfo{person}{Yifei Li}, \bibinfo{person}{Xingyu Liu}, \bibinfo{person}{Zhiqiang Liu}, {and} \bibinfo{person}{et al.}} \bibinfo{year}{2023}\natexlab{}.
\newblock \showarticletitle{FinEval: A Chinese Financial Domain Knowledge Evaluation Benchmark for Large Language Models}.
\newblock \bibinfo{journal}{\emph{arXiv e-prints}} (\bibinfo{year}{2023}).
\newblock


\bibitem[Zhang et~al\mbox{.}(2022)]%
        {90-zhang2022opt}
\bibfield{author}{\bibinfo{person}{Susan Zhang}, \bibinfo{person}{Stephen Roller}, \bibinfo{person}{Naman Goyal}, \bibinfo{person}{Mikel Artetxe}, \bibinfo{person}{Moya Chen}, \bibinfo{person}{Shuohui Chen}, \bibinfo{person}{Christopher Dewan}, \bibinfo{person}{Mona Diab}, \bibinfo{person}{Xian Li}, \bibinfo{person}{Xi~Victoria Lin}, \bibinfo{person}{Todor Mihaylov}, \bibinfo{person}{Myle Ott}, \bibinfo{person}{Sam Shleifer}, \bibinfo{person}{Kurt Shuster}, \bibinfo{person}{Daniel Simig}, \bibinfo{person}{Punit~Singh Koura}, \bibinfo{person}{Anjali Sridhar}, \bibinfo{person}{Tianlu Wang}, {and} \bibinfo{person}{Luke Zettlemoyer}.} \bibinfo{year}{2022}\natexlab{}.
\newblock \showarticletitle{OPT: Open Pre-trained Transformer Language Models}.
\newblock  (\bibinfo{year}{2022}).
\newblock
\showeprint[arxiv]{2205.01068}~[cs.CL]


\bibitem[Zhang et~al\mbox{.}(2024)]%
        {zhang2024eclipse}
\bibfield{author}{\bibinfo{person}{Wei Zhang}, \bibinfo{person}{Xianfu Cheng}, \bibinfo{person}{Yi Zhang}, \bibinfo{person}{Jian Yang}, \bibinfo{person}{Hongcheng Guo}, \bibinfo{person}{Zhoujun Li}, \bibinfo{person}{Xiaolin Yin}, \bibinfo{person}{Xiangyuan Guan}, \bibinfo{person}{Xu Shi}, \bibinfo{person}{Liangfan Zheng}, {et~al\mbox{.}}} \bibinfo{year}{2024}\natexlab{}.
\newblock \showarticletitle{ECLIPSE: Semantic Entropy-LCS for Cross-Lingual Industrial Log Parsing}.
\newblock \bibinfo{journal}{\emph{arXiv preprint arXiv:2405.13548}} (\bibinfo{year}{2024}).
\newblock


\bibitem[Zhang et~al\mbox{.}(2019)]%
        {logrobust}
\bibfield{author}{\bibinfo{person}{Xu Zhang}, \bibinfo{person}{Yong Xu}, \bibinfo{person}{Qingwei Lin}, \bibinfo{person}{Bo Qiao}, \bibinfo{person}{Hongyu Zhang}, \bibinfo{person}{Yingnong Dang}, \bibinfo{person}{Chunyu Xie}, \bibinfo{person}{Xinsheng Yang}, \bibinfo{person}{Qian Cheng}, \bibinfo{person}{Ze Li}, \bibinfo{person}{Junjie Chen}, \bibinfo{person}{Xiaoting He}, \bibinfo{person}{Randolph Yao}, \bibinfo{person}{Jian-Guang Lou}, \bibinfo{person}{Murali Chintalapati}, \bibinfo{person}{Furao Shen}, {and} \bibinfo{person}{Dongmei Zhang}.} \bibinfo{year}{2019}\natexlab{}.
\newblock \showarticletitle{Robust log-based anomaly detection on unstable log data}. In \bibinfo{booktitle}{\emph{Proceedings of the 2019 27th ACM Joint Meeting on European Software Engineering Conference and Symposium on the Foundations of Software Engineering}} (Tallinn, Estonia) \emph{(\bibinfo{series}{ESEC/FSE 2019})}. \bibinfo{publisher}{Association for Computing Machinery}, \bibinfo{address}{New York, NY, USA}, \bibinfo{pages}{807--817}.
\newblock
\showISBNx{9781450355728}
\urldef\tempurl%
\url{https://doi.org/10.1145/3338906.3338931}
\showDOI{\tempurl}


\bibitem[Zhang et~al\mbox{.}(2021)]%
        {93-ZHANG2021216}
\bibfield{author}{\bibinfo{person}{Zhengyan Zhang}, \bibinfo{person}{Yuxian Gu}, \bibinfo{person}{Xu Han}, \bibinfo{person}{Shengqi Chen}, \bibinfo{person}{Chaojun Xiao}, \bibinfo{person}{Zhenbo Sun}, \bibinfo{person}{Yuan Yao}, \bibinfo{person}{Fanchao Qi}, \bibinfo{person}{Jian Guan}, \bibinfo{person}{Pei Ke}, \bibinfo{person}{Yanzheng Cai}, \bibinfo{person}{Guoyang Zeng}, \bibinfo{person}{Zhixing Tan}, \bibinfo{person}{Zhiyuan Liu}, \bibinfo{person}{Minlie Huang}, \bibinfo{person}{Wentao Han}, \bibinfo{person}{Yang Liu}, \bibinfo{person}{Xiaoyan Zhu}, {and} \bibinfo{person}{Maosong Sun}.} \bibinfo{year}{2021}\natexlab{}.
\newblock \showarticletitle{CPM-2: Large-scale cost-effective pre-trained language models}.
\newblock \bibinfo{journal}{\emph{AI Open}}  \bibinfo{volume}{2} (\bibinfo{year}{2021}), \bibinfo{pages}{216--224}.
\newblock
\showISSN{2666-6510}
\urldef\tempurl%
\url{https://doi.org/10.1016/j.aiopen.2021.12.003}
\showDOI{\tempurl}


\bibitem[Zhu et~al\mbox{.}(2019)]%
        {logpai}
\bibfield{author}{\bibinfo{person}{J. Zhu}, \bibinfo{person}{S. He}, \bibinfo{person}{J. Liu}, \bibinfo{person}{P. He}, \bibinfo{person}{Q. Xie}, \bibinfo{person}{Z. Zheng}, {and} \bibinfo{person}{M.~R. Lyu}.} \bibinfo{year}{2019}\natexlab{}.
\newblock \showarticletitle{Tools and benchmarks for automated log parsing}.
\newblock \bibinfo{journal}{\emph{2019 IEEE/ACM 41st International Conference on Software Engineering: Software Engineering in Practice (ICSE-SEIP)}} (\bibinfo{year}{2019}), \bibinfo{pages}{121--130}.
\newblock


\end{thebibliography}

\end{CJK}
\end{document}